\def\year{2022}\relax
\documentclass[letterpaper]{article} % DO NOT CHANGE THIS
\usepackage{aaai22}  % DO NOT CHANGE THIS
\usepackage{times}  % DO NOT CHANGE THIS
\usepackage{helvet}  % DO NOT CHANGE THIS
\usepackage{courier}  % DO NOT CHANGE THIS
\usepackage[hyphens]{url}  % DO NOT CHANGE THIS
\usepackage{graphicx} % DO NOT CHANGE THIS
\usepackage{natbib}  % DO NOT CHANGE THIS AND DO NOT ADD ANY OPTIONS TO IT
\usepackage{caption} % DO NOT CHANGE THIS AND DO NOT ADD ANY OPTIONS TO IT
\DeclareCaptionStyle{ruled}{labelfont=normalfont,labelsep=colon,strut=off} % DO NOT CHANGE THIS
\usepackage{amsmath}
\usepackage{multirow}
\usepackage{booktabs}
\usepackage{amssymb}
\usepackage{csquotes}
\usepackage[switch]{lineno}
\usepackage{enumitem}
\usepackage{subcaption}
 
\newcommand\ie{\emph{i.e.}}

\usepackage[normalem]{ulem}
\usepackage{todonotes} % for comments
\usepackage{tabularx}
\usepackage{algorithm}

\usepackage{algorithmic}
\usepackage{newfloat}
\usepackage{listings}
\floatstyle{ruled}
\newfloat{listing}{tb}{lst}{}
\floatname{listing}{Listing}
\title{Transmission-Guided Bayesian Generative Model for Smoke Segmentation}
\author{
    Siyuan Yan,
  Jing Zhang,
  Nick Barnes\\
}
\affiliations{
    The Australian National University\\
\{siyuan.yan, jing.zhang, nick.barnes\}@anu.edu.au 
}
\usepackage{bibentry}
\begin{document}
\maketitle

\begin{abstract}
Smoke segmentation is essential to precisely localize wildfire so that it can be extinguished in an early phase. Although deep neural networks have achieved promising results on image segmentation tasks, they are prone to be overconfident for smoke segmentation due to its non-rigid shape and transparent appearance. This is caused by both knowledge level uncertainty due to limited training data for accurate smoke segmentation and labeling level uncertainty representing the difficulty in labeling ground-truth. To effectively model the two types of uncertainty, we introduce a Bayesian generative model to simultaneously estimate the posterior distribution of model parameters and its predictions.
% the prediction is prone to be overconfident and unreliable. This is because the uncertainty is naturally inherent in data(aleatoric uncertainty) and model(epistemic uncertainty). This problem is especially serious on the smoke segmentation task due to the smoke's translucent property and the high risk of false alarm. Thus, accurate uncertainty estimation is the key to integrate the smoke segmentation model into real-world applications. 
% In this paper, we propose a Bayesian latent variable framework that can not only produce accurate smoke segmentation results but also model both epistemic uncertainty and aleatoric uncertainty. Also, 
 Further, smoke images suffer from low contrast and ambiguity, inspired by physics-based image dehazing methods, we design a transmission-guided local coherence loss to guide the network to learn pair-wise relationships based on pixel distance and the transmission feature.
%  {\bf Note deletion here- check ok}
 %NB Not sure about this - its smoke in our case, calling it degradation is for those familiar with the fog removal type literature, here it seems irrelevant
 %especially on image degradation regions. 
 To promote the development of this field, we also contribute a high-quality smoke segmentation dataset, SMOKE5K, consisting of 1,400 real and 4,000 synthetic images with pixel-wise annotation.
%  for new benchmarking. 
 Experimental results on benchmark testing datasets illustrate that our model achieves both accurate predictions and reliable uncertainty maps representing model ignorance about its prediction. Our code and dataset are publicly available at: \url{https://github.com/redlessme/Transmission-BVM}.
% In addition to demonstrating our model can achieve state-of-the-art performance on smoke segmentation benchmark datasets, we also illustrate how our model can estimate epistemic uncertainty and aleatoric uncertainty for better decision-making.

\end{abstract}

\section{Introduction}
\label{sec:intro}

Smoke segmentation (wildfire smoke in this paper)
% \footnote{In this paper, we focus on wildfire smoke.} 
aims to localize the scope of the smoke area, and is usually defined as a binary segmentation task. Existing solutions \citep{smoke1,smoke3} mainly focus on effective feature aggregation with larger receptive fields for multi-scale prediction.
% : 1) feature fusion, 2) large receptive fields, and 3) multi-scale prediction.
We argue that, different from conventional binary segmentation tasks, \ie~salient object detection \citep{BASNet_Sal,ucnet++,jing2021learning_nips,jing2021rgbd_iccv}, background detection \cite{background}, and transparent object detection \cite{tod1}, smoke segmentation is unique from at least two perspectives: 1) the shape of smoke is non-rigid; thus structure-preserving solutions \citep{BASNet_Sal,ucnet++} adopted in existing binary segmentation models may not be effective; 2) smoke can be transparent, making it similar to
% segmentation related to that of 
transparent object detection. However, differently, models for transparent objects \citep{tod1,tod3} focus on rigid objects, which makes smoke segmentation a more challenging task compared with existing binary segmentation or transparent object detection tasks.

To model smoke's non-rigid shape and transparent
% \NB{appearance? rather than appearance throughout?}
appearance, we introduce a Bayesian generative model to estimate the posterior distribution of both model parameters
% (caused by the non-rigid shape and transparent appearance of smoke \NB{This is repeated from the start of the sentence}) 
and its predictions.
% (caused by the difficulty of labeling due to the non-rigid shape and transparent appearance of smoke \NB{again}). 
Specifically, we introduce a Bayesian variational auto-encoder \citep{bnn_lvm1} to produce stochastic predictions, making it possible to capture uncertainty.
Uncertainty \citep{uncertainty0,uncertainty2} is defined as model ignorance about its prediction. \cite{uncertainty0} introduces two types of uncertainty, epistemic (model) uncertainty and aleatoric (data) uncertainty. The former is caused by ignorance about the task, which can be reduced with more training data. The latter is caused by labeling accuracy or ambiguity inherent in the labeling task, which usually cannot been reduced.
%NB which usually depends on the ability of the annotator,
% \NB{, 
% or
% } 
% and does not reduce with more samples.

\begin{figure}[!t]
\setlength{\abovecaptionskip}{0.cm}
\setlength{\belowcaptionskip}{-0.cm}
   \begin{center}
   \centering
   \begin{tabular}{ c@{} c@{ } c@{ } c@{ } c@{} c@{}}
   {\includegraphics[width=0.150\linewidth]{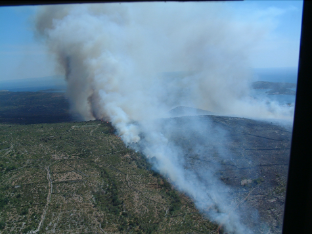}}&
   {\includegraphics[width=0.15\linewidth]{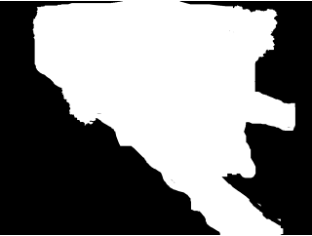}}&
      {\includegraphics[width=0.15\linewidth]{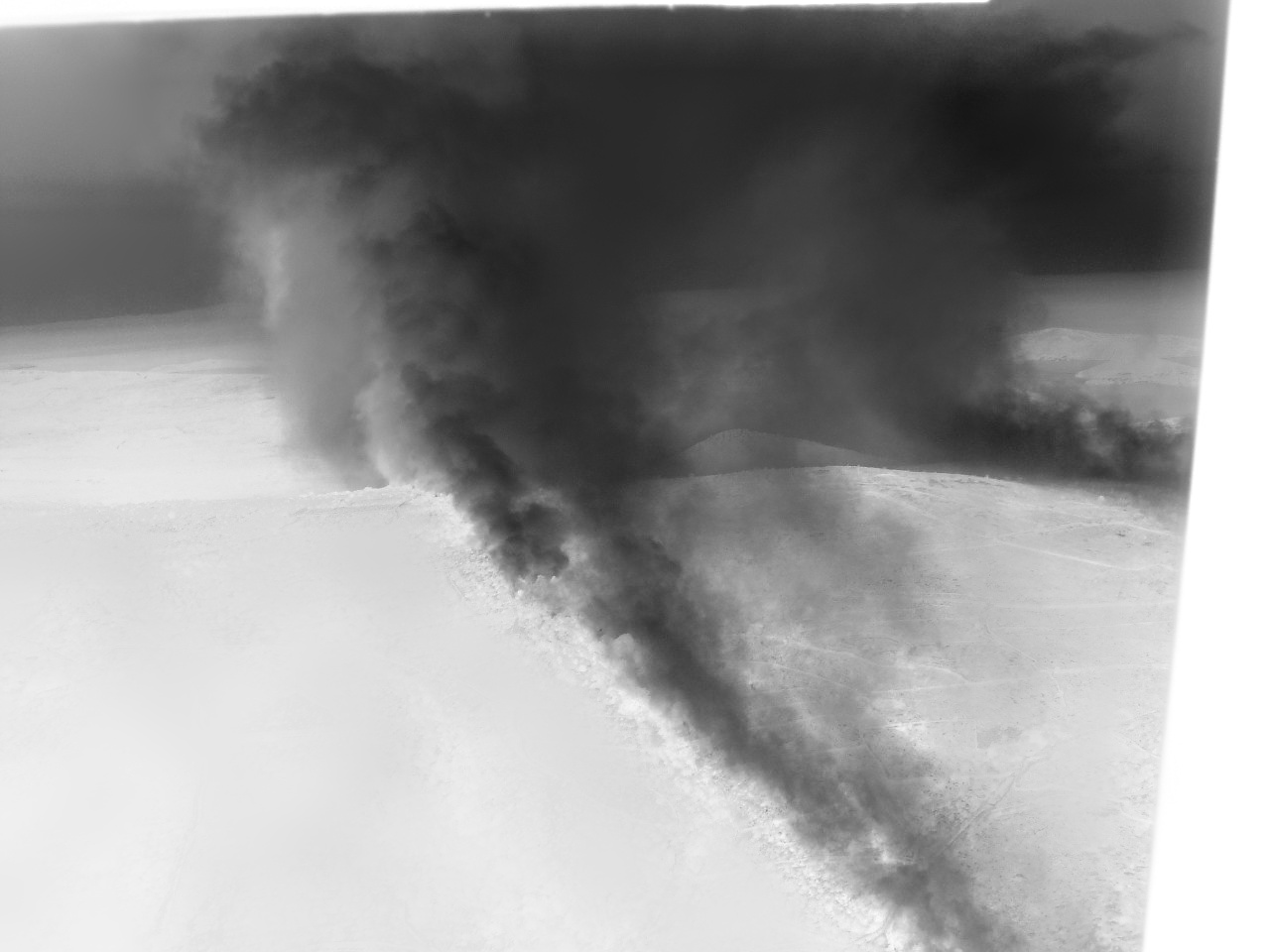}}&
   {\includegraphics[width=0.15\linewidth]{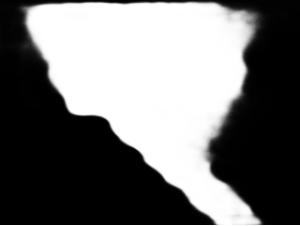}}&
    {\includegraphics[width=0.15\linewidth]{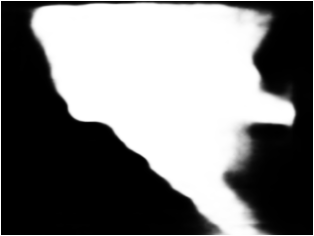}}& {\includegraphics[width=0.15\linewidth]{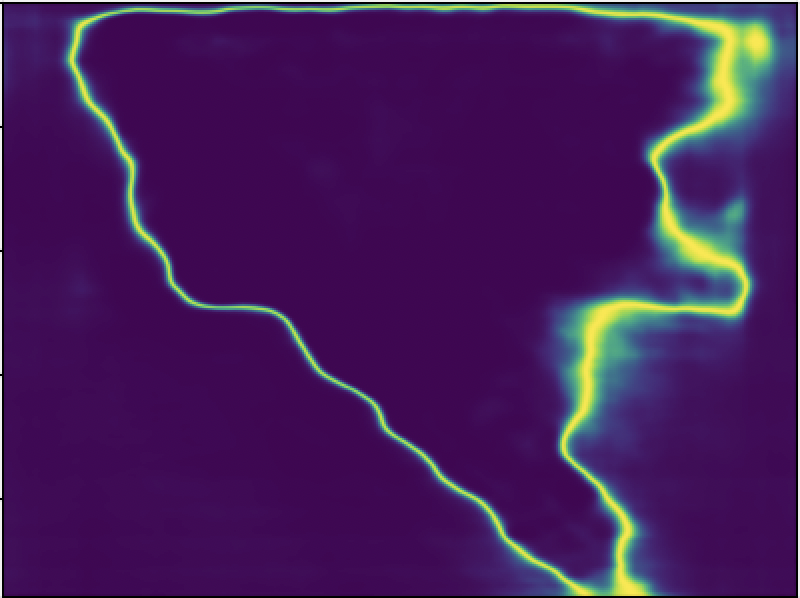}}\\
   {\includegraphics[width=0.15\linewidth]{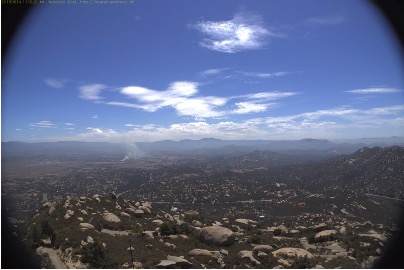}}&
   {\includegraphics[width=0.15\linewidth]{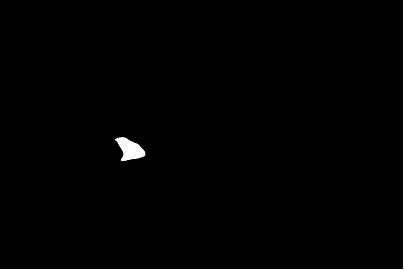}}&
    {\includegraphics[width=0.15\linewidth]{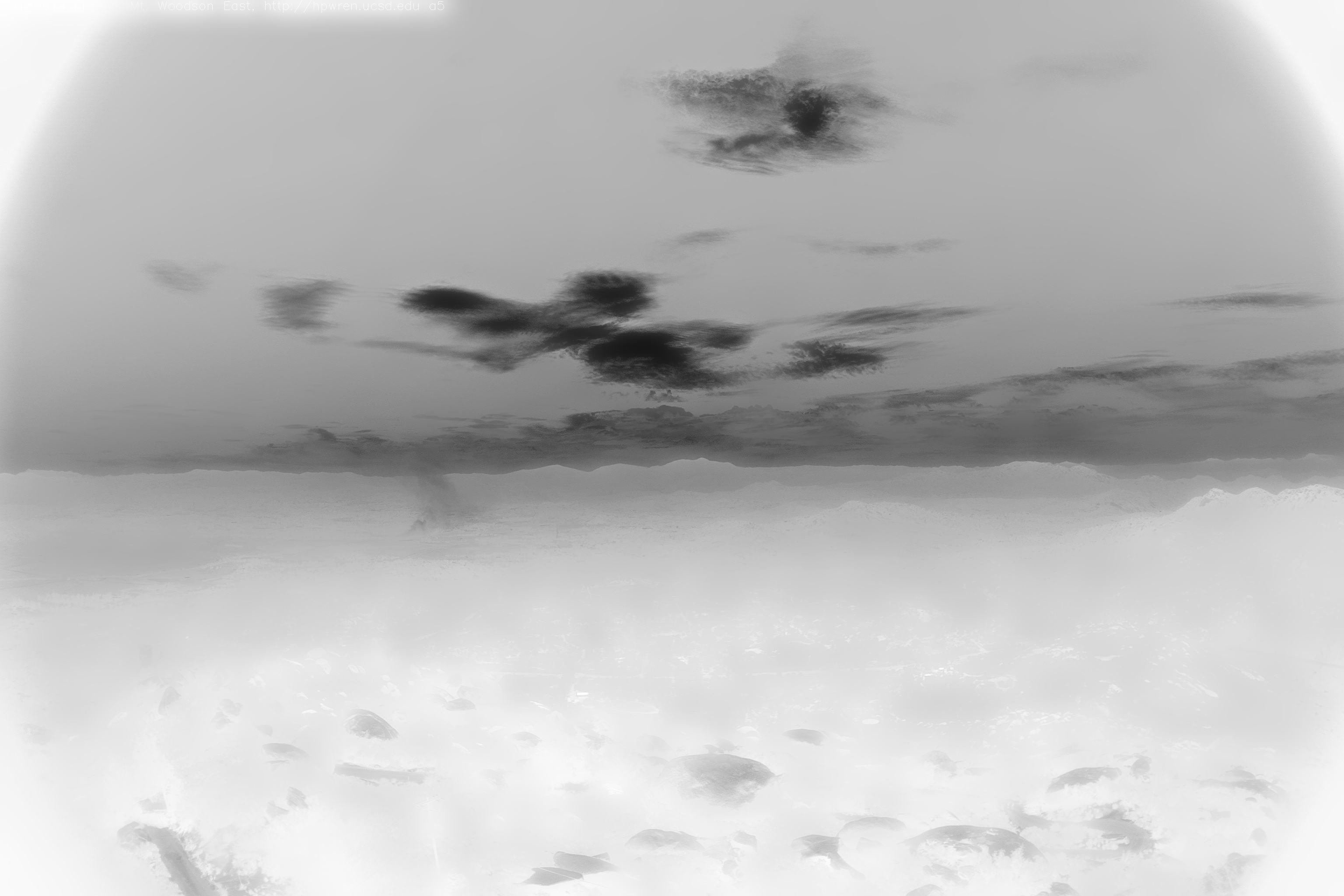}}&
   {\includegraphics[width=0.15\linewidth]{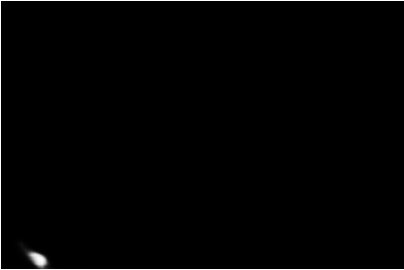}}&
  {\includegraphics[width=0.15\linewidth]{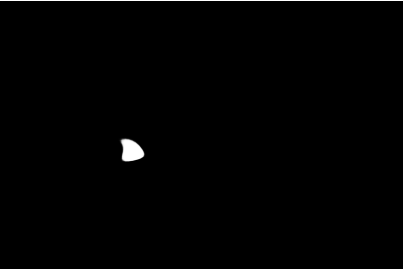}}&
   {\includegraphics[width=0.15\linewidth]{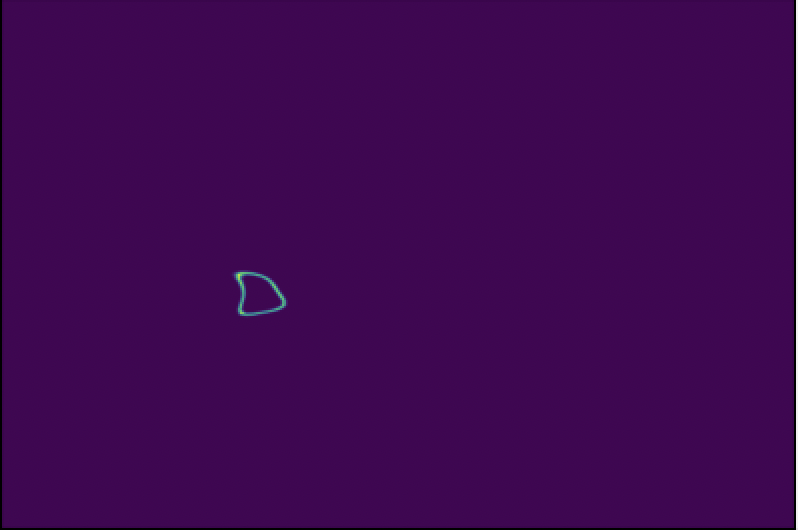}}\\
   \footnotesize{ Image} & \footnotesize{ GT} &  \footnotesize{ Trans.}&\footnotesize{w/o T}&\footnotesize{w/ T}&\footnotesize{Uncer.}\\
   \end{tabular}
   \end{center}
\caption{\normalsize Transmission loss is effective for segmenting smoke. From left to right: input images, ground-truth, transmission map, prediction without transmission loss, prediction with transmission loss, and predicted uncertainty. Row 1: modeling transmission features 
%NB relationship 
using our loss is helpful to distinguish the ambiguous boundary. Row 2: our transmission loss is effective with small smoke regions that have low contrast due to 
%it 
being semi-transparent.
}
\label{fig1_}
\end{figure}

For smoke segmentation, our limited knowledge about the smoke model leads to epistemic uncertainty, while the difficulty of labeling leads to aleatoric uncertainty. We intend to model both uncertainties within a Bayesian variational auto-encoder to achieve both accurate smoke segmentation and an estimation of a
% \NB{estimation of a} 
reliable uncertainty map representing model ignorance about its prediction given the training dataset and learned model parameters.
Further, smoke images suffer from quality degradation, \ie~low contrast due to smoke's translucent appearance,
% \NB{and/or cloud} 
causing difficulty for classification.
% \NB{for detection}\sout{ to detect}. 
In the image enhancement field, \citep{transmission,he} employ transmission that represents the portion of light that reaches the camera to model the quality degradation degree of images.
% \NB{of pixels in}.
In the
% \NB{the} 
smoke segmentation literature,
% \NB{literature}\sout{field},
\citep{tran2,d_smoke2} exploit transmission to
% \sout{well} 
enhance the difference between smoke and background. Inspired by them, we propose a transmission-guided local coherence loss to model pair-wise similarity based on pixel distance and transmission to better differentiate smoke from background.
% Further, w
We also employ reverse transmission as the weight of the loss to encourage the model to focus more on degraded regions.
% \NB{Further,} 
% to tackle the above mentioned \enquote{quality degradation} problem,
% % like low contrast \NB{What is the quality degradation problem in the smoke segmentation context?}, 
% we employ the reversed transmission \cite{transmission} value as the weight of the loss to encourage the model to focus more on degraded regions.

% Further, we note that no % Moreover, smoke segmentation is currently not well studied due to the lack of a 
We also find there exists no
high-quality, large-scale smoke segmentation dataset.
% exists. 
In Table \ref{tab1}, we compare benchmark smoke datasets in terms of size, image type, and annotation. The current largest real smoke dataset contains only 143 images.
% , which is easy to overfit. 
Although there is a large-scale synthetic image dataset SYN70K \cite{data1}, we find that the dataset is inadequate due to many redundant images.
% and the diversity of images are also limited. 
% We will perform experiments to verify this in the section ``SMOKE5K Dataset". 
Furthermore, we notice that the
% we also find that current 
widely used test sets for smoke segmentation are all synthetic. Due to
% there is a 
the huge
% leading to a huge 
domain gap between the synthetic data and the real data, we claim that the model performs good on the synthetic data may not work well in real life.
% making it less effective to use in using models in real life.
% for which there exists a huge domain gap with real smoke images. 
To promote the development of this field, we provide the first high-quality, large-scale smoke dataset with 5,000 training images and 400 testing images, where the training set consists of 1,400 real life images and 4,000 synthetic images with pixel-wise annotations, and the test images are all real life images. We also provide mask annotation and scribble annotation for fully supervised and weakly supervised smoke segmentation tasks. 
 
We summarize our contributions as: 1) we introduce the first Bayesian generative model for smoke segmentation to produce both accurate predictions and reliable uncertainty maps
% representing model ignorance over its predictions
given the non-rigid shape and transparent appearance of smoke; 
2) we contribute a novel transmission loss to differentiate smoke from its background and encourage the model to emphasize the
% pay more attention to 
quality degraded regions. 3) we contribute a large-scale smoke segmentation dataset, containing 5,400 images with both pixel-wise mask and scribble annotations for effective model learning and evaluation.

\section{Related Work} 
\noindent\textbf{Smoke Segmentation:}
Existing methods rely on
% use 
deep convolutional neural networks (CNNs), and mainly focus on: 1) multi-scale prediction \citep{fcn} as the size of the smoke varies in images; 2) high-low level feature aggregation for fusion \citep{refinenet,ccl}; and 3) enlarging the receptive field for effective context modeling \citep{deeplabv1,aspp}.
% With the rapid development of deep learning in recent years, Convolutional Neural Network (CNN) based methods have been explored on the smoke segmentation task. Current existing methods mainly focus on multi-scale prediction, feature fusion, and increasing the receptive field. 
Among them, \citep{smoke1} designs a saliency detection network to highlight the most informative smoke with different levels of feature fusion. \cite{smoke3} introduces a gated recurrent network with classification assistance resulting in further refinement. 
\cite{ATT} proposes an Attention U-net to fuse coarse and fine layers for multi-scale prediction. \cite{smoke5} relies on ASPP \cite{aspp} to achieve large receptive fields. Different from current smoke segmentation methods that utilize generic semantic segmentation architectures and strategies, we tackle smoke segmentation based on smoke's unique attributes, introducing uncertainty and transmission loss to model the non-rigid shape and translucent appearance. 

\noindent\textbf{Uncertainty Estimation:}
% \textbf{Uncertainty Estimation}
%NB Uncertainty represents model ignorance about it's prediction.
\cite{uncertainty2} defines two types of uncertainty: aleatoric and epistemic.
% The former 
% % There are two kinds of uncertainty to model. Aleatoric (data) uncertainty 
% accounts for the noise inherent in the data, while the latter
% % Epistemic (model) uncertainty 
% represents the uncertainty in the model parameters,
% % and convergence, 
% which can be explained away with more training data.
% reduced when training data increases.
% The two kinds of uncertainty can also be combined to predict, called predictive(total) uncertainty. \\
Epistemic uncertainty is usually modeled by replacing model weights using a parametric distribution. Most solutions for epistemic uncertainty estimation are based on Bayesian neural networks \citep{bnn2}. As the posterior inference of model parameters is intractable,
% but BNN suffers from a huge computation burden due to the intractability of the posterior. To tackle this, 
some works propose to use variational approximations to approximate the posterior inference, \ie~
% Bayesian inference, like 
Monte Carlo dropout \cite{drop}, weight decay \cite{weight}, early stopping \cite{early-stop}, ensemble models \cite{ensemble}, and M-head models \cite{m-head}.
Aleatoric uncertainty is usually obtained by generating a distribution over the model's prediction
% logit 
space \citep{uncertainty5}. \cite{prob_unet} proposes a UNet \cite{unet} with a conditional variational auto-encoder \cite{cvae} to capture labelling ambiguity.
% this kind of uncertainty. 
\cite{face} uses stochastic embedding to capture the data uncertainty for classification.
% Also, some researchers \citep{decomp1,decomp2,decomp3,DEUP} decompose epistemic uncertainty and aleatoric uncertainty from predictive uncertainty. 
\cite{decomp1} defines the combination of the two types of uncertainty as predictive uncertainty to capture both epistemic uncertainty and aleatoric uncertainty simultaneously.

\begin{table}[t]
% \footnotesize
    \small
    \centering
    \renewcommand\tabularxcolumn[1]{m{#1}}
 	\newcolumntype{e}{>{\hsize=0.04\textwidth}X}
 	\newcolumntype{T}{>{\hsize=0.125\textwidth}X}

    \begin{tabularx}{\textwidth} {T|cc|cc}
        \cline{1-5}
		Dataset   & Size & Type  &Scribble &Mask\\
		\cline{1-5}
	
		DST\citeyearpar{data3} & 740 & Synthetic  &   &\checkmark \\
	    
		SSD\citeyearpar{data5}& 143 & Real& & \checkmark\\
			
		SYN70K\citeyearpar{data1}& 73632 & Synthetic & &\checkmark\\ 
        \cline{1-5}
		 \textbf{Ours} & 5400 & Mixed  & \checkmark&\checkmark\\ 
		\cline{1-5}
    \end{tabularx}
    \caption{ \normalsize Benchmark
    % Comparison with public 
    smoke segmentation datasets.}
    \label{tab1}
\end{table}

% \subsection{Uncertainty Quantification}
\noindent\textbf{Medium Transmission:}
Medium transmission represents the portion of the light that reaches the camera, which is usually estimated using the dark channel prior algorithm \cite{he}. In image dehazing and enhancement \citep{he,underwater1}, transmission has been applied to emphasize degraded regions of the image. In smoke segmentation, \citep{tran2} exploit transmission directly as a feature to detect smoke by thresholding the transmission map. Within the deep learning methods, \cite{tran3} performs smoke image classification by directly feeding the transmission map as input into a CNN to enhance the difference between smoke and background. To the best of our knowledge, although transmission has been explored in image enhancement, no previous deep learning based methods have investigated it for segmentation tasks. Different from existing solutions, we introduce a novel loss to model transmission similarity of pixels and encourage the model to focus on the degraded regions.

\noindent\textbf{Smoke Segmentation Dataset:}
% The smoke segmentation task is not well studied due to the lack of high quality, large-scale dataset.
% Currently, there are only a few public datasets, and their quality is not satisfactory. 
Currently, the largest smoke dataset is SYN70K \cite{data1}, a synthetic dataset consisting of a limited number of smoke objects in different scenes. Also, \cite{data4} contributes a wildfire dataset with 2,192 real images captured by HPWREN cameras, but it does not contain mask annotation. To promote the field, we introduce the first high-quality, large-scale smoke segmentation dataset.

\section{Our Method}
\begin{figure}[!t]
   \begin{center}
   {\includegraphics[width=0.99\linewidth]{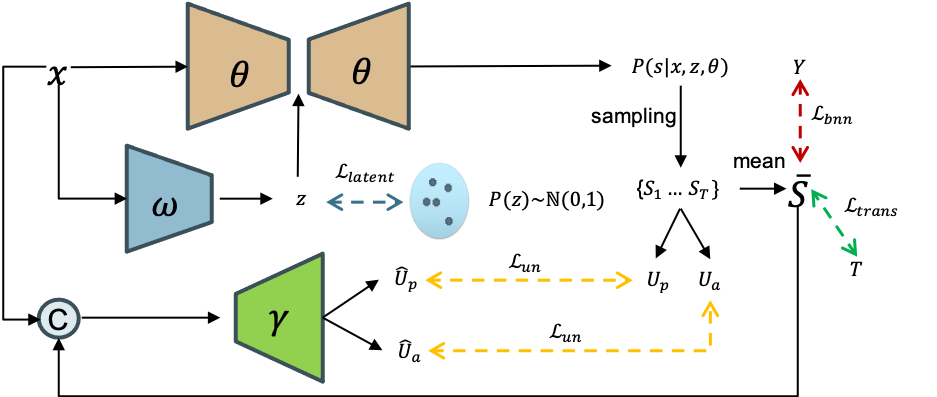}}
   \end{center}
%   \vspace{-10mm}

\caption{\normalsize
Overview of our Bayesian latent variable model for smoke segmentation, where \textcircled{c} is the concatenation operation. To simplify the structure, we do not include the uncertainty calibration entropy loss here. The main modules included in our framework: a Bayesian Neural Network $P_\theta(y|x,z)$ to generate stochastic predictions, an inference model $P_\omega(z|x)$ to obtain the latent variable $z$, and the uncertainty estimation network $P_\gamma(x,P_\theta(y|x,z))$ to approximate the sampling based uncertainty for sampling-free uncertainty estimation during testing.
% BNN denotes the Bayesian neural network. LVN denotes latent variable network. UEN denotes the uncertainty estimation network.Y: ground truth; T: transmission map; $U_{p}$: predictive uncertainty map; $U_{a}$: aleatoric uncertainty map; $\mathcal{L}_{bnn}$: structure-aware loss; $\mathcal{L}_{latent}$: KL divergence; $\mathcal{L}_{trans}$: transmission loss; $\mathcal{L}_{un}$: uncertainty consistency loss. 
% During inference, our network predicts the result and the corresponding uncertainty maps by inference part inside the red rectangle without any sampling operation.
}
   \label{fig:overview}
\end{figure}

% In this section, we introduce the whole architecture of our Bayesian latent variable model (BNN-LVN), which is shown in Figure. 
% Let us first define o
Our training dataset is $\mathcal{D}=\lbrace (x_{i},y_{i})\rbrace_{i=1}^{N} $, where $x_{i}$ is an input image, $y_{i}$ is the corresponding ground truth, and N is the size of the training dataset.
Our proposed model consists of three sub-networks (see Fig.~\ref{fig:overview}): 1) An inference network $P_\omega(z|x)$ with parameter set $\omega$
% latent variable network 
that encodes input $x_{i}$ into a low dimensional latent variable $z_{i}$, where the prior of $z_{i}$ is defined as a standard normal distribution $\mathcal N(0,1)$.
% where $z_{i}$ tends to be close to the Gaussian distribution $\mathcal N(0,1)$.
The latent variable $z_{i}$ is used to capture the inherent noise in the input image that leads to the difficulty of labeling: 
% in the input $x_{i}$ and the uncertainty caused by the noise is known as 
aleatoric uncertainty. 2) A Bayesian network $P_\theta(y|x,z)$ maps input $x_{i}$ and generated latent variable $z_{i}$ to stochastic prediction $s_{i}=P_\theta(y|x_i,z_i)$, where model parameters $\theta$ are assumed to follow some specific distribution $P(\theta)$.
% by Monte-Carlo dropout. 
% During training, the network collects weight samples, the uncertainty captured by latent variable z and model weights are known as predictive uncertainty
% \NB{What is this sentence - a list of what the network does - not clear from grammar}. 
3) An uncertainty estimation network $g_\gamma$ that takes image pairs $(x_{i},s_{i})$ as input to approximate the sampling-based uncertainty produced by our Bayesian latent variable network, achieving
% which is used during testing to achieve 
sampling-free uncertainty estimation at test time, where $\gamma$ is the corresponding parameter set.
% network with latent variables during training. 
% The uncertainty captured by the uncertainty estimation network is called total uncertainty.
Following \cite{decomp1}, we use $g_\gamma$ to regress the
% the uncertainty captured by $g_\gamma$ as 
total uncertainty and the aleatoric uncertainty, where the former is defined as the sum of
% the former includes both 
aleatoric uncertainty and epistemic uncertainty.

% \NB{why slashslash?}\\
% In the following subsections, we will introduce each sub-network and our physical \NB{Physics inspired, or physically-inspired? physical inspired is not really a thing} model inspired, transmission guided loss.
\subsection{Bayesian Latent Variable Model}

\noindent\textbf{Latent Variable Model:} 
 Smoke images are often corrupted by senor noise, and ambiguous due to smoke's translucient appearance. Conventional deterministic networks based on the biased dataset
 can easily have over-confident predictions, leading to both high false-positive and false-negative rates. We argue that early detection of wildfires is critical to community safety, and a well-calibrated model is desirable.
%  which is aware of its prediction is desirable.
%  , which is
%  high rates of false alarms and miss detection, the lack of subsequent diagnosis based on uncertainty mapping limits real-world application.
%  \NB{Further, early detection of wildfires is critical to community safety.} Thus, it is essential to incorporate data uncertainty into the neural network. 
 
Let us assume the noise-aware model prediction as:
%  \begin{equation}
%       y_{i}=f_{w}(x_{i},z_{i})+\epsilon_{i}\sigma_{i}
%  \end{equation}
 \begin{equation}
      y_{i}=f_{\theta}(x_{i},z_{i})+\epsilon_{i},
 \end{equation}
 where $f_{\theta}(x_{i},z_{i})$ is the prediction of a neural network, $z_{i}$ is a latent variable, $\epsilon_{i}$ is additive noise drawn from a normal distribution $\mathcal{N}(0,\sigma^2\mathbb{I})$, with $\sigma$ as standard deviation.
%  , and $\sigma_{i}$ is an input-dependent noise parameter. 
 Given input $x_{i}$ and latent variable $z_{i}$, network output will be corrupted by additive noise $\epsilon_{i}$, which we assume is conditioned on the input image.
 It will lead the network to force $\sigma$ to be small to reduce the influence of the noise in the prediction. Then, the latent variable $z_{i}$ can capture unobserved stochastic features that can affect the network's prediction.
 
 Following conventional practice of variational auto-encoders, we train the latent variable by maximizing the expected variational lower bound (ELBO) as:
 \begin{equation}
    \label{vea_lower_bound}
\begin{aligned}
    L(\theta,\omega;x) &= \mathbb{E}_{z\sim{P_\omega(z|x)}}[-\log(P_\theta(y|x,z))]\\&+D_{KL}(P_\omega(z|x)||P(z)),
\end{aligned}
\end{equation}
where $P_{\omega}(z|X)$ is the posterior distribution of the latent variable $z$ or the inference model, and $P(z)$ is its prior distribution, which we defined as a standard Gaussian distribution $\mathcal{N}(0,1)$. $D_{KL}(P_\omega(z|x)||P(z))$ is the Kullback-Leibler divergence regularizer. We design the inference model $P_{\omega}(z|X)$ with five convolutional layers that maps input $x$ into a low dimensional latent variable $z$ to capture the inherent ambiguities of human labeling. Specifically, the output of the inference model is $\mu^{post}$ and $\sigma^{post}$, leading to
% , with 
$P_\omega(z|x) = \mathcal{N}(\mu^{post},\sigma^{post})$ where $\mu^{post}$ and $\sigma^{post}$ represent the mean and the standard deviation of the posterior distribution of $z$, and $z$ is then obtained with the reparameteration trick: $z=\mu^{post}+\sigma^{post}*\epsilon$, where $\epsilon\sim\mathcal{N}(0,1)$.

% To train the network, we use a Kullback-Leibler divergence regularizer to constrain $P_{w}(z|X)$ to be close to a standard Gaussian distribution $\mathcal{N}(0,1)$. The regularization term is defined as:\\
%  \begin{equation}
%   \mathcal{L}_{latent}= KL[(P_{w}(z|x)||P(z))]
% \end{equation}
% \noindent
% where $P(z)$ is a normally distributed prior $\mathcal{N}(0,1)$.

\noindent\textbf{Bayesian Neural Network with Latent Variable:}
%\subsubsection{Bayesian Neural Network with Latent Variable}
% Epistemic uncertainty is usually increased when the test images are out of the training distribution. 
Knowing when the model cannot be trusted is important in safety-critical systems such as wildfire detection. Bayesian neural networks (BNN) can model epistemic uncertainty by replacing the neural network’s weights with distributions. However, they suffer from 
%a 
huge computational burden due to the intractability of the posterior inference $P(\theta|D)$. Fortunately, BNNs can be approximated via variational inference with Monte Carlo (MC) dropout \cite{drop} by using Monte Carlo integration \cite{drop} to %\sout{as an}
approximate
%\sout{ion of}
the intractable marginal likelihood $P(y|x)$:
% and then epistemic uncertainty can be captured by calculating the entropy or variance of T times MC sampling:
\begin{equation}
    P(y|x,D) \approx \frac{1}{B} \sum_{b=1}^{B}P(y|x,\theta_{b}),
\end{equation}
where $B$ indicates the times of sampling.

As $z$ can model the aleatoric uncertainty, and $P(\theta|D)$ can model epistemic uncertainty, with a Bayesian latent variable model, we aim to model both epistemic and aleatoric uncertainty, leading to predictive uncertainty estimation.
% , we incorporate a BNN and an LVN into a unified framework.
We design our BNN based latent variable model (BNN-LVM) with a ResNet50 backbone \cite{resnet} as encoder to obtain feature maps $\{f_k\}_{k=1}^4$ from different stages of the backbone. The latent variable $z$ is introduced to the network via two steps. Firstly, we sample $z$ from $P_\omega(z|x)$, and tile it to achieve the same spatial size as $f_{4}$. Secondly, we concatenate the tiled $z$ with $f_{4}$, and feed it to
a $3\times3$ convolutional layer to obtain $f_{4}'$ of the same size as $f_{4}$, which will serve as the new backbone feature.
% Then, we feed four features maps and the latent variable $z$ sampled from $P_\omega(z|x)$ into the decoder by concatenating the tiled $z$ (to achieve the same spatial size as $f_{4}$) with $f_{4}$. Then we feed the concatenated feature to a $3\times3$ convolutional layer to obtain $f_{4}'$ of the same size as $f_{4}$. 
% each of two inputs channel-wise. 
To capture the epistemic uncertainty, the produced new features $\{\{f_k\}_{k=1}^3,f_{4}'\}$ are fed into four different $3 \times 3$ convolutional layer with a dropout rate of 0.3 to get the new backbone features $\{s_k\}_{k=1}^4$. Subsequently, we use four residual channel attention modules \cite{rcab} after each $s_k$
% feature maps $\{s_k\}_{i=1}^4$ 
to obtain discriminative feature representations.
% and fuse them together. 
Finally, we use a DenseASPP \cite{aspp} module to aggregate higher-lower level features with enlarged receptive field and get the final prediction of our BNN-LVM as $s=P_\theta(y|x,z)$.
% where $z$ is the latent variable produced by the LVN, $\theta$ is the weight samples produced by MC-dropout.

% \Jing{I'm here!}
To train the BNN, we adopt the weighted structure-aware loss \cite{sloss} as:
% , which is in the form: 
\begin{equation}
\label{bnn_loss}
    \mathcal{L}_{bnn}(s,y)=w*\mathcal{L}_{ce}(s,y)+\mathcal{L}_{iou}(s,y),
\end{equation}
\noindent
where $s$ and $y$ are the prediction and the ground truth respectively, $w$ is the edge-aware weight defined as $1+5*|H_{AP}(y)-y|$ where $H_{AP}$ is the average pooling function, $\mathcal{L}_{ce}$ and $\mathcal{L}_{iou}$ denote the cross-entropy loss and the IOU loss. $\mathcal{L}_{bnn}(s,y)$ is then the expected negative log likelihood term in Eq.~\ref{vea_lower_bound}.

\noindent\textbf{Uncertainty Quantification:}
Uncertainty quantification \cite{jiawei2022modeling,li2021uncertainty,ucnet++} can provide
% people 
more information about model prediction, leading to better decision making. For smoke segmentation, quantification of the uncertainty map can potentially reduce the false-alarm rate and guide people to resolve ambiguities. 

Given the model prediction $P_\theta(y|x,z)$ from our BNN module,
% BNN with latent variables is defined as  $P(y|x,z,\theta)$, and it does not only depend on input $x$, but also latent variable $z$ and the model's weight samples $\theta$. Thus, 
the predictive (total) uncertainty can be measured as entropy of the mean prediction,
% by the entropy of the expected distribution $P(y|x,D)$, 
which can be formulated as $\mathcal{H}[\mathbb{E}_{P(\theta|D),z\sim P_{\omega}(z|x)}[P_\theta(y|x,z)]]$ where $\mathcal{H}$ denotes the entropy and $\mathbb{E}$ denotes the expectation. The corresponding expectation can be approximated by Monte Carlo (MC) integration, leading to mean prediction as:
% of the model weights.
% % So, our predictive uncertainty map can be calculated as the entropy of the mean prediction. Given 
% Let's define the mean prediction as:
\begin{equation}
\label{mean_prediction}
     p_\mu=\frac{1}{B}P_{\theta_{b}}(y|x,z),
\end{equation}
where $B$ is the iterations of
% % number of repetitions of 
MC sampling, and $\theta_b$ is the parameter set of the $b^{th}$ iteration of sampling.
The predictive uncertainty is then defined as:
% the entropy of $p_\mu$, which is
$U_{p}=\mathcal{H}[p_{\mu}]$. The aleatoric uncertainty is the average entropy for a fixed set of model weights, which can be formulated as: $U_{a}=\mathbb{E}_{P(\theta|D),z\sim P_\omega(z|x)}[\mathcal{H}[P_\theta(y|x,z)]]$. The epistemic uncertainty is then defined as the gap between predictive uncertainty and aleatoric uncertainty: $U_e = U_p-U_a$.

% \begin{equation}
%      U_{p}= -\sum_{c \in {0,1}} \frac{1}{T}P(y=c|x,\theta_{t})\log(\frac{1}{T}P(y=c|x,\theta_{t}))
%      \label{eq7}
% \end{equation}
% \begin{equation}
%      U_{p}= - \frac{1}{T}P(y|x,\theta_{t})\log(\frac{1}{T}P(y=c|x,\theta_{t}))
%      \label{eq7}
% \end{equation}

% Modified by \cite{ep}, we measure the aleatoric uncertainty map by:
% \begin{equation}
%     U_{a} \approx -\sum_{c \in {0,1}}P(y=c|x,\theta_{t}^{*})\log(\frac{1}{T}P(y=c|x,\theta_{t}^{*}))
%     \label{eq8}
% \end{equation}

% where $\theta^{*}$ denotes the weights that lead to the smallest average entropy of $U_{a}$, this can force $U_{a}\leq U_{p}$ in the most pixels.

% Epistemic uncertainty can be assessed by calculating the mutual information between model prediction $y$ and model weights $\theta$:

% \begin{equation}
% \begin{aligned}
%     \mathcal{MI}[y,\theta|x,D]=\mathcal{H}[\mathbb{E}_{p(\theta|D)}[p(y|x,\theta)]] \\
%     -\mathbb{E}_{p(\theta|D)}[\mathcal{H}[p(y|x,\theta)]]
%     \label{mi}
% \end{aligned}
% \end{equation}
% where $\mathcal{MI}$ is the mutual information.

% We can see the epistemic uncertainty is actually quantified by calculating the difference between predictive uncertainty and aleatoric uncertainty. Thus, we can calculate the epistemic uncertainty map by
% \begin{equation}
%     U_{e}=U_{p}-U_{a}
%     \label{eq9}
% \end{equation}

% where we set all negative pixels of $U_{e}$ to be 0, benefit from Equation \ref{eq8}, we empirically find the portion of negative pixel is very small and will not affect the final results.
\noindent\textbf{Uncertainty Estimation Network:}
Conventional solutions for uncertainty estimation involve
% , especially for epistemic uncertainty, requires 
multiple iterations of sampling \cite{jiawei2022modeling,ucnet} during testing, limiting the
% samples during the inference stage, limiting their 
real-time application. As real-time performance is important for wildfire detection, we move the sampling process from the inference to the training stage. During training, we sample our BNN-LVN multiple times to obtain sampling-based predictive uncertainty and aleatoric uncertainty. Then, we use a designed uncertainty estimation network to approximate the calculated predictive uncertainty and aleatoric uncertainty. Specifically, our network produces two types of uncertainty with a widely used M-head structure \cite{m-head}, where we use a shared encoder with five convolutional layers and two independent decoders with three convolutional layers to regress each type of uncertainty.
% different predictions. 
The uncertainty estimation network $g_\gamma$
% $P(\hat{U_{a}},\hat{U_{p}}|x,y)$ 
uses batch normalisation and LeakyReLU activation for all layers except the last layers of the two decoders. It takes the concatenation of image $x$ and its prediction $s$ as input to approximate the $U_{a}$ and $U_{b}$ by uncertainty consistency loss:
\begin{equation}
\label{uncertainty_consistency}
    \mathcal{L}_{un}=\frac{1}{2}(\mathcal{L}_{mse}(U_{p},\hat{U_{p}})+\mathcal{L}_{mse}(U_{a},\hat{U_{a}})),
\end{equation}
where $\mathcal{L}_{mse}$ is $L2$ distance, $\hat{U_{p}}$ and $\hat{U_{a}}$ are the approximated uncertainty maps with the proposed uncertainty estimation network $g_\gamma(x,s)$.

\subsection{Transmission Guided Local Coherence Loss}
With the loss function in Eq.~\ref{vea_lower_bound} and the uncertainty estimation related loss in Eq.~\ref{uncertainty_consistency}, we can already train our Bayesian latent variable model for smoke segmentation.
% Many existing losses \cite{w1,w2} preserve the structure information of objects based on the intensity or gradient of the given image. 
However, we find that model with the smoke segmentation related loss in Eq.~\ref{bnn_loss} fails to predict distinct boundaries at the ambiguous smoke edges. Inspired by image dehazing, we propose a transmission-guided local coherence loss, which exploits transmission as an essential feature to discriminate the smoke boundary. Our loss is based on the assumption that smoke objects tend to have a different transmission from most regions of the background. It can enforce similar predictions for pixels with similar transmission and close distance. Further, outdoor smoke images often suffer from quality degradation, \ie~low contrast. \cite{transmission} shows that image quality degradation regions usually have lower transmission values. Inspired by \cite{transmission}, we use the reversed transmission value as the weight of our loss to encourage the model to focus on the degraded regions, which is
% Reversed transmission guided loss is then
defined as:
% Our loss is in the form:
\begin{equation}
\label{transmission_loss}
    \mathcal{L}_{trans}= \sum_{m}\sum_{n \in K_{m}}(1-T(m))W(m,n)D(m,n),
\end{equation}
where $K_{m}$ is a $k \times k$ kernel centered at pixel $m$, $D(m,n)$ is the $L1$ distance between the prediction of pixels $m$ and $n$, and $W(m,n)$ is a modified bilateral kernel in the form:
\begin{equation}
    W(m,n)=\frac{1}{w} \exp(-\frac{||P(m)-P(n)||^2}{2\sigma_{P}^2}-\frac{||T(m)-T(n)||^2}{2\sigma^2_{T}}),   
\end{equation}
where $\frac{1}{w}$ is the normalized weight term, $P(m), P(n)$, and  $T(m),T(n)$ represent the spatial information and transmission information of the pixels $m$ and $n$ respectively, and $\sigma_{P}^2, \sigma_{T}^2$ are the kernel bandwidth.

To estimate medium transmission, we calculate transmission information of each pixel of the intensity image by:
\begin{equation}
    T(m)=1-\min \limits_{c}(\min\limits_{n \in K(m)}(\frac{I^c(n)}{A^c})),
\label{d2}
\end{equation}
where $A$ is global atmospheric light, $I$ is the intensity image, $c$ is the color channel, $K(m)$ is a local patch of size $k\times k$ centered at pixel $m$ and then $\min \limits_{c}(\min\limits_{n
\in K(m)}(\frac{I^c(n)}{A^c})$ is the normalized haze map defined at \cite{he}. We can see the estimated medium transmission is based on the global atmospheric light $A$. To get $A$, we follow \cite{he} and obtain it by picking the top 0.1\% brightest pixels in the dark channel of the intensity image. 

% The detail of the dark channel algorithm is introduced in the supplementary material.

\subsection{Objective Function}
So far, we have $\mathcal{L}(\theta,\omega;x)$ in Eq.~\ref{vea_lower_bound} for the Bayesian latent variable model, $\mathcal{L}_{un}$ in Eq.~\ref{uncertainty_consistency} for the uncertainty estimation module, and $\mathcal{L}_{trans}$ in Eq.~\ref{transmission_loss} to force the network focus on the image degradation regions.
% the , and $\mathcal{L}_{trans}$. 
We further adapt entropy loss as a regularizer to encourage the prediction to be binary. Entropy loss is defined as the entropy of model prediction, which is minimized when the model produces binary predictions. However, we find that directly applying entropy loss to our task leads to
% often makes the prediction
overconfident predictions,
% in practice, 
causing extra false positives. To reduce the \enquote{binarization regularizer} for less confident regions, we add a temperature scaling term \cite{on_calibration} to the original entropy loss, which uses learned total uncertainty as the temperature so that the model can produce softened predictions on high uncertainty pixels. Note that the temperature in our method is learned by the uncertainty estimation network rather than being a user-defined temperature as used in traditional temperature scaling methods.
% , which is less flexible. 
Our uncertainty-based temperature scaling can also be used to calibrate our model. 
%\NB{what you mean by free lunch here is not clear} 
A well-calibrated model cannot only reduce the gap between prediction confidence and model accuracy but also achieve better performance. 

Our uncertainty calibrated entropy loss is defined as:
\begin{equation}
    \mathcal{L}_{c}=-\sum_{c \in {0,1}} \sigma(s/U_{p})\log\sigma(s/U_{p}),
    \label{eq10}
\end{equation}
\noindent
where $\sigma$ is the sigmoid function, $s$ is the network prediction, and $U_{p}$ is the total uncertainty. 

Finally, our total loss for BNN-LVM is defined as:
% \begin{equation}
%     \mathcal{L}_{gen}=\mathcal{L}_{bnn}+\lambda_{1}\mathcal{L}_{latent}+\lambda_{2}\mathcal{L}_{trans}+\lambda_{3}\mathcal{L}_{c},
% \end{equation}
\begin{equation}
\label{bnn_lvm}
    \mathcal{L}_{gen}=\mathcal{L}(\theta,\omega;x)+\lambda_{1}\mathcal{L}_{trans}+\lambda_{2}\mathcal{L}_{c},
\end{equation}
and empirically
% where 
we set
% $\lambda_{1}=10$, 
$\lambda_{1}=0.3$ and
$\lambda_{2}=0.01$. With Eq.~\ref{bnn_lvm} we can update our Bayesian latent variable model, and the uncertainty estimation model $g_\gamma$ is updated with loss function $\mathcal{L}_{un}$ in Eq.~\ref{uncertainty_consistency}.

\begin{table*}[t!]
  \centering
  \small
%   \scriptsize
%   \renewcommand{\arraystretch}{1.0}
  \renewcommand{\tabcolsep}{1.4mm}

  \begin{tabular}{lr|cccccccccccccc|c}
  \hline

    \textit{Dataset}&  & SMD    & LRN &
  Deeplab v1  & HG-Net8 & LKM  & RefineNet& PSPNet & CCL & DFN & DSS & W-Net &CGRNet & Ours\\
  &&\citeyearpar{smd}&\citeyearpar{lrn}&\citeyearpar{deeplabv1}&\citeyearpar{hgnet}&\citeyearpar{lkm}&\citeyearpar{refinenet}&\citeyearpar{pspnet}&\citeyearpar{ccl}&\citeyearpar{dfn}             &\citeyearpar{data1}&\citeyearpar{wave}&\citeyearpar{smoke3}\\

  \hline

    \multirow{1}{*}{\textit{DS01}}
     &     & .3209&.3069&.2981&.3187&.2658&.2486&.2366&.2349&.2269&.2745&.2688&.2138& \textbf{.1210}  \\
%   \hline
    \multirow{1}{*}{\textit{DS02}}
     && .3379&.3078&.3030&.3301&.2799&.2590&.2480&.2498&.2411&.2894&.2548&.2280 & \textbf{.1362}  \\
%   \hline 
   \multirow{1}{*}{\textit{DS03}}
     && .3255&.3041&.3010&.3215&.2748&.2515&.2430&.2429&.2332&.2861&.2596&.2212 & \textbf{.1291}  \\
   \hline

  \end{tabular}
    \caption{ \normalsize Performance ($mMse\downarrow$) on three synthetic datasets.}\label{tab:BenchmarkResults}
  \label{tab2}
\end{table*}

\section{SMOKE5K Dataset}
As shown in Table \ref{tab1},
the
% the size of the 
existing largest real smoke segmentation dataset contains
% is 
only 143 images. Although \citeauthor{data1} create a large synthetic dataset for benchmarking, their dataset suffers from redundancy,
% mainly has two weaknesses: (1) The dataset is too large to train, and many images are redundant. (2) The diversity of their dataset is limited 
as they crop and copy a limited number of simulated smoke to different scenes. To promote the development of the smoke segmentation, we create a high-quality smoke segmentation dataset, namely \textit{SMOKE5K}, that consists of both synthetic images and real images for new benchmarking.
% We will demonstrate the quality of our dataset is better than the current largest benchmark dataset SYN70K by two experiments. More analysis of our dataset is included in the supplementary material.

\subsection{Dataset Construction}
We create our dataset based on an open wildfire smoke dataset \cite{data4}, the current largest synthetic smoke dataset \cite{data1} and a few images from the internet. Due to the large redundancy of the
% we think most images in the 
synthetic dataset,
% are redundant, 
we only choose 4,000 images from the synthetic dataset and discard images that contain duplicated smoke foreground. For real datasets, we choose 1,360 good quality smoke images from the \cite{data4} dataset, and discard
% while we remove images that are duplicated or have 
smoke images of poor quality, and we also choose 40 smoke images from the internet to further increase the diversity of the test set. As the real smoke datasets do not have pixel-wise annotations, we then annotate the dataset with both binary mask and scribble annotation.
% for both fully supervised and weakly supervised smoke segmentation tasks. 
Finally, we end up with 5,400 images, where
% We extract 
400 real smoke images from it is used for testing, which is defined as \enquote{Total}, and the remaining 5,000 images are for training. To better explore the difficulty of different attributes in the smoke segmentation models, we further extract 100 images from the 400 test images, denoted as \enquote{Difficult}, which mainly contains transparent smoke and smoke with more diverse shapes.

\subsection{Dataset Comparison}

\begin{figure}[t]
    \centering
    \setlength{\abovecaptionskip}{0.cm}
\setlength{\belowcaptionskip}{-0.cm}
    \begin{tabular}{c@{ }c@{ }}
    % \subfigure{
    % \begin{minipage}[t]{0.48\linewidth}
    % \centering

    \includegraphics[width=0.45\linewidth]{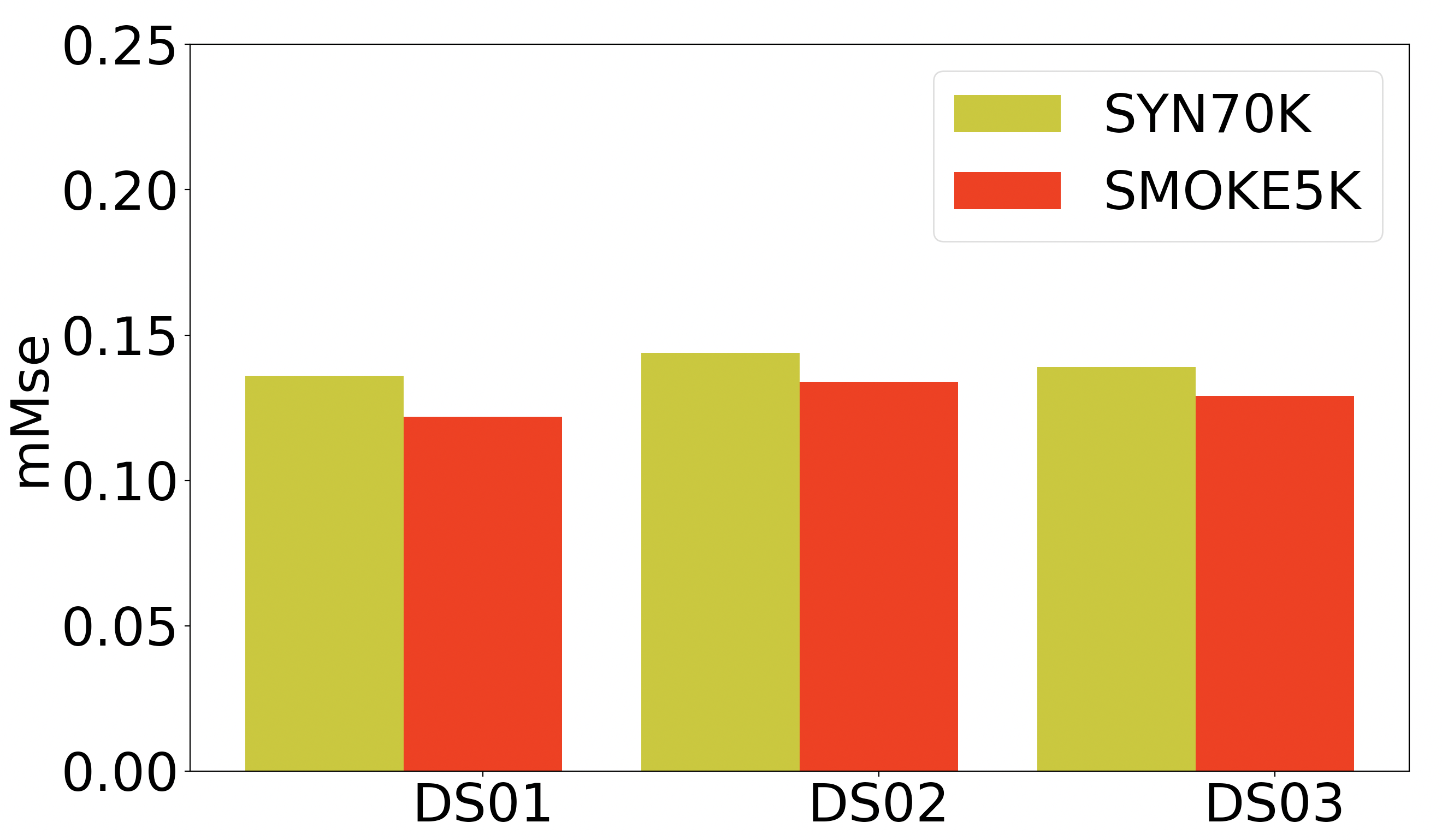}&
    % \end{minipage}
    % }
    % \subfigure{
    % \begin{minipage}[t]{0.48\linewidth}
    % \centering

    \includegraphics[width=0.45\linewidth]{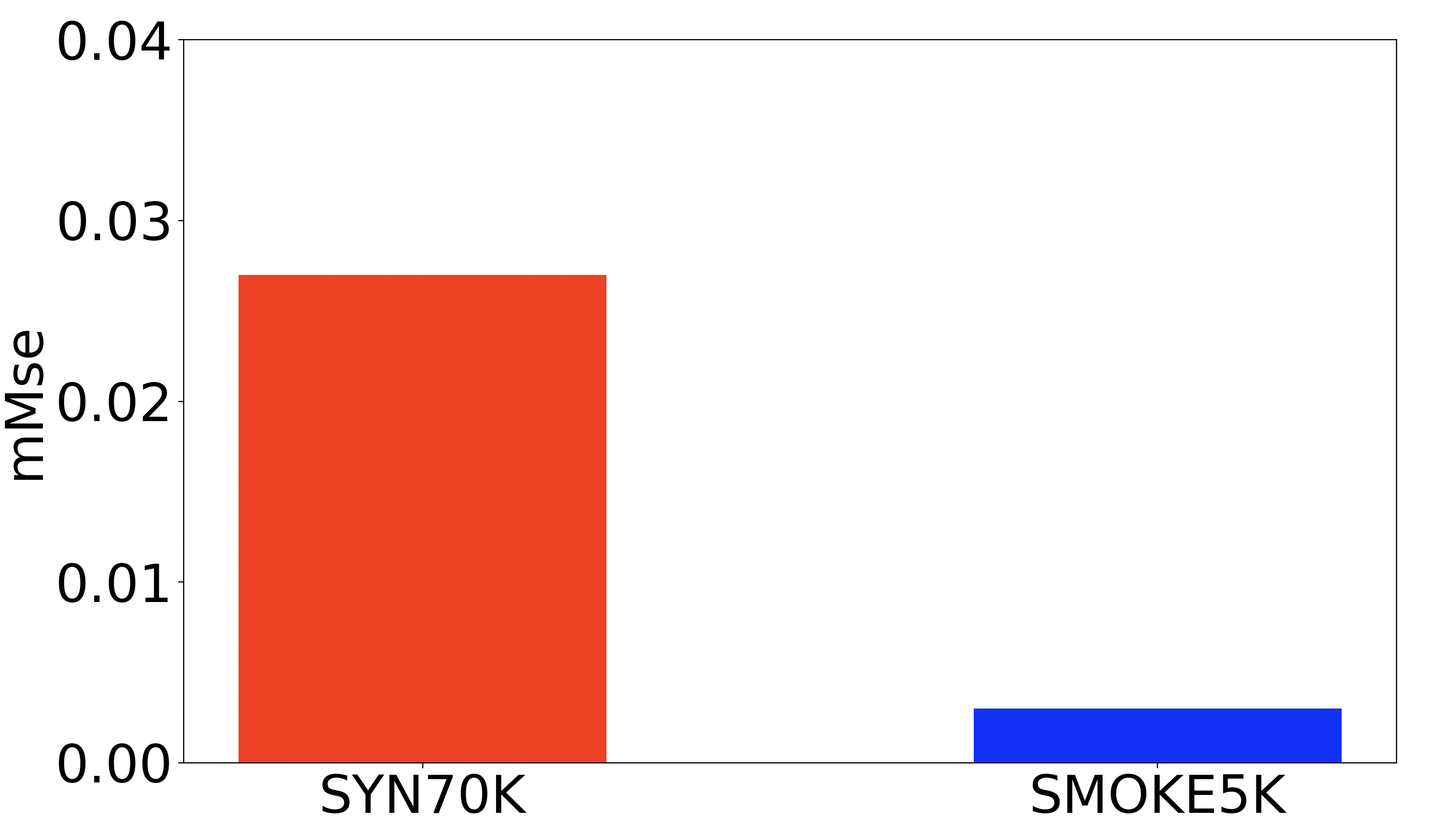}\\
    \footnotesize{a}&\footnotesize{b}\\
 
    \end{tabular}
    % \end{minipage}
    % }
    \caption{\normalsize (a) Comparing the model trained using SYN70K and the model trained using SMOKE5K on the synthetic test set. (b) Comparing the model trained using SYN70K and the model trained using SMOKE5K on our real smoke dataset.}
    \label{fig_exp}
\end{figure}

We have carried out two experiments to verify the superiority of our dataset. Firstly, we train the same segmentation model on both SYN70K and our dataset SMOKE5K. It takes about three days to train on SYN70K and about 12 hours to train on our dataset. Then, we evaluate the two models on three synthetic datasets (DS01, DS02, DS03) from \cite{data1}, and the results are shown in Fig. \ref{fig_exp} (a). We observe that although SYN70k is at least 14 times larger than our dataset, the model trained with our dataset still produces the lower mean square error (MSE) on all three synthetic test sets.
% demonstrating that many images in SYN70K are redundant. 
Further, we evaluate the two models on our real image test set, and the result is shown in Fig.~\ref{fig_exp} (b), which further explains the effectiveness of our new training dataset.
% We can see the model trained using our dataset shows much better performance, verifying that our dataset is more diverse.

% \begin{figure}[h!]
%     \centering
%     \begin{tabular}{
%         >{\centering\arraybackslash}m{0.2\linewidth}
%         >{\centering\arraybackslash}m{0.2\linewidth}
%         >{\centering\arraybackslash}m{0.2\linewidth}
%         >{\centering\arraybackslash}m{0.2\linewidth}
%     }
%     \includegraphics[width=\linewidth,height=1.2cm]{Image/f1.jpg}
%     &
%     \includegraphics[width=\linewidth,height=1.2cm]{Image/f2.jpg}
    
%     &
%     \includegraphics[width=\linewidth,height=1.2cm]{Image/f3.jpg}
%     &
%     \includegraphics[width=\linewidth,height=1.2cm]{Image/f4.jpg}
%     \\
%     \includegraphics[width=\linewidth,height=1.2cm]{Image/f1.png}
%     &
%     \includegraphics[width=\linewidth,height=1.2cm]{Image/f2.png}
%     &
%     \includegraphics[width=\linewidth,height=1.2cm]{Image/f3.png}
%     &
%     \includegraphics[width=\linewidth,height=1.2cm]{Image/f4.png}
%     \\

%     \end{tabular}
%     \caption{Image quality of different smoke datasets.}
%     \label{fig1}
% \end{figure}

\begin{table}[t!]

  \centering
  \small

%   \scriptsize
  \renewcommand{\arraystretch}{1.0}
  \renewcommand{\tabcolsep}{1.3mm}

  \begin{tabular}{lr|ccccc|c}
  \hline
%   \toprule
%   &  &\multicolumn{14}{c|}{Fully Sup. Models}&\multicolumn{1}{c}{Weakly Sup./Unsup. Models} \\
    &\textit{Metric}&F3Net&BASNet&SCRN&ITSD&UCNet & Ours\\
  &&\citeyearpar{sloss}&\citeyearpar{BASNet_Sal}& \citeyearpar{scrnet}&\citeyearpar{ITSD}&\citeyearpar{ucnet++}
  \\

  \hline

    \multirow{2}{*}{\textit{T}}
     &$\mathcal{M}\downarrow$& .004& .005&.003&.003&.003&\textbf{.002} \\
%   \hline
    % \multirow{1}{*}{ }
%   \hline 
%   \multirow{1}{*}{}
     &$F_{\beta}\uparrow$& .670 & .733&.769 &.774&.787&\textbf{.791} \\
   \hline
     \multirow{2}{*} {\textit{D}}
     &$\mathcal{M}\downarrow$& .011& .014&.009&.010&.009&\textbf{.006} \\
%   \hline
    % \multirow{1}{*}{ }

%   \hline 
%   \multirow{1}{*}{}
     &$F_{\beta}\uparrow$& .633 & .662&.682 &.688&.707&\textbf{.741} \\
   \hline
  \end{tabular}
    \caption{\normalsize Performance on the test set of SMOKE5K.
  }

  \label{tab3}
\end{table}

\section{Experimental Results}

\subsection{Implementation Details}

\noindent\textbf{Datasets:}
% Firstly, f
For fair comparisons with existing smoke segmentation models, we first follow the same settings in \cite{data1} and train our model using SYN70K \cite{data1} and evaluate on their three public smoke segmentation benchmarks: (1) DS01; (2) DS02; (3) DS03. Then, we
provide a new benchmark
by training
% setting for smoke segmentation, which
% trains 
with our SMOKE5K dataset and tests on our real-image testing set.
% We provide a new benchmark on our new testing set with both quantitative and qualitative comparisons.

\noindent\textbf{Baselines and Evaluation Metrics:}
% Existing smoke segmentation models are mostly trained with the synthetic training dataset, and then test on the synthetic testing dataset, which is our first setting. 
We compare with existing smoke segmentation models following the first setting (training and testing on synthetic datasets) and
% then compare with those models and
% To the best of our knowledge, there is no public high-quality smoke segmentation model available yet in this field. So, we directly compare performance with previous benchmarks in the same setting \cite{smoke3}, 
report performance in Table \ref{tab2}.
%For our new dataset, a
As no code is available for existing smoke segmentation models, we select five state-of-the-art saliency detection baselines (F3Net~\cite{sloss}, BASNet~\cite{BASNet_Sal}, SCRN \cite{scrnet}, ITSD \cite{ITSD} and UCNet \cite{ucnet++}) due to the similarity between the two tasks. %and 
We re-train them on our new smoke segmentation benchmark, the SMOKE5K dataset.
%which is our new smoke segmentation benchmark. 
We show the
% and the 
results
% are provided 
in Table \ref{tab3}. 

Following previous smoke segmentation literature, we choose the Mean Square Error (Mse) ($\mathcal{M}$) as the evaluation metric. The average
% value of 
Mses (mMse) on the test set is used to measure performance.
% quantify evaluation performance.
Mse computes the per-pixel accuracy of model prediction, which depends on the size of the smoke foreground, leading to inherent small Mse for smoke image with small size of smoke.
% Although mMse is a widely used evaluation metric for smoke segmentation, it is 
% sensitive to the size of the smoke foreground, leading to relative smaller Mse for smaller smoke and larger, which will assign a small object a smaller error and a large object a larger error. 
To tackle the problem, we adopt the F-measure ($F_{\beta}$) as complementary
% complement 
to mMse for comprehensive performance evaluation, which
% to better reflect prediction quality.
% F-measure 
is a weighted combination of precision and recall values. Further, as an
% generative model for 
uncertainty estimation model,
% who is aware of uncertainty of it's prediction, 
we also adopt the reliability diagram \cite{on_calibration} and dense calibration measure (ECE) \cite{ece} to represent
% which is a function of model accuracy and its confidence, representing 
model's calibration degree.
% To evaluate whether the model is well-calibrated, we use the 
% dense calibration measure (ECE) \cite{ece} 
% as a measure of model calibration degree. Specifically,
% based on the knowledge that confidence of a well-calibrated model should be consistent with model accuracy. Along with it, we also 
% we report the dense calibration measure (ECE) \cite{ece,on_calibration} to represent
% % which is a function of model accuracy and its confidence, representing 
% model's calibration degree.

\noindent\textbf{Training Details:}
We train our framework using PyTorch with a maximum of 50 epochs.
Each image is re-scaled to $480\times 480$.
Empirically, we set the dimension of the latent space $z$ as 8.
% as 8 to achieve a trade-off between the diversity of prediction and model performance. 
The learning rates of the generator $P_\theta(y|x,z)$ and the uncertainty estimation network $g_\gamma$ are initialized to 2.5e-5 and 1.5e-5, respectively. We use the Adam optimizer and decrease the learning rate 0.8 after 40 epochs with a maximum epoch of 50. We adopt ResNet50 backbone to the encoder of $P_\theta(y|x,z)$, which is initialized with parameters trained for image classification, and the other newly added layers are initialized with the PyTorch default initialization strategy. It took about three days of training on SYN70K and 15 hours on SMOKE5K with batch size 6 using a single NVIDIA GeForce RTX 2080Ti GPU.

\subsection{Comparison with State-of-the-Art Methods}

\begin{figure}[!t]
   \begin{center}
   \begin{tabular}{ c@{ } c@{ } c@{ } c@{ }  c@{ }   }

   {\includegraphics[width=0.18\linewidth]{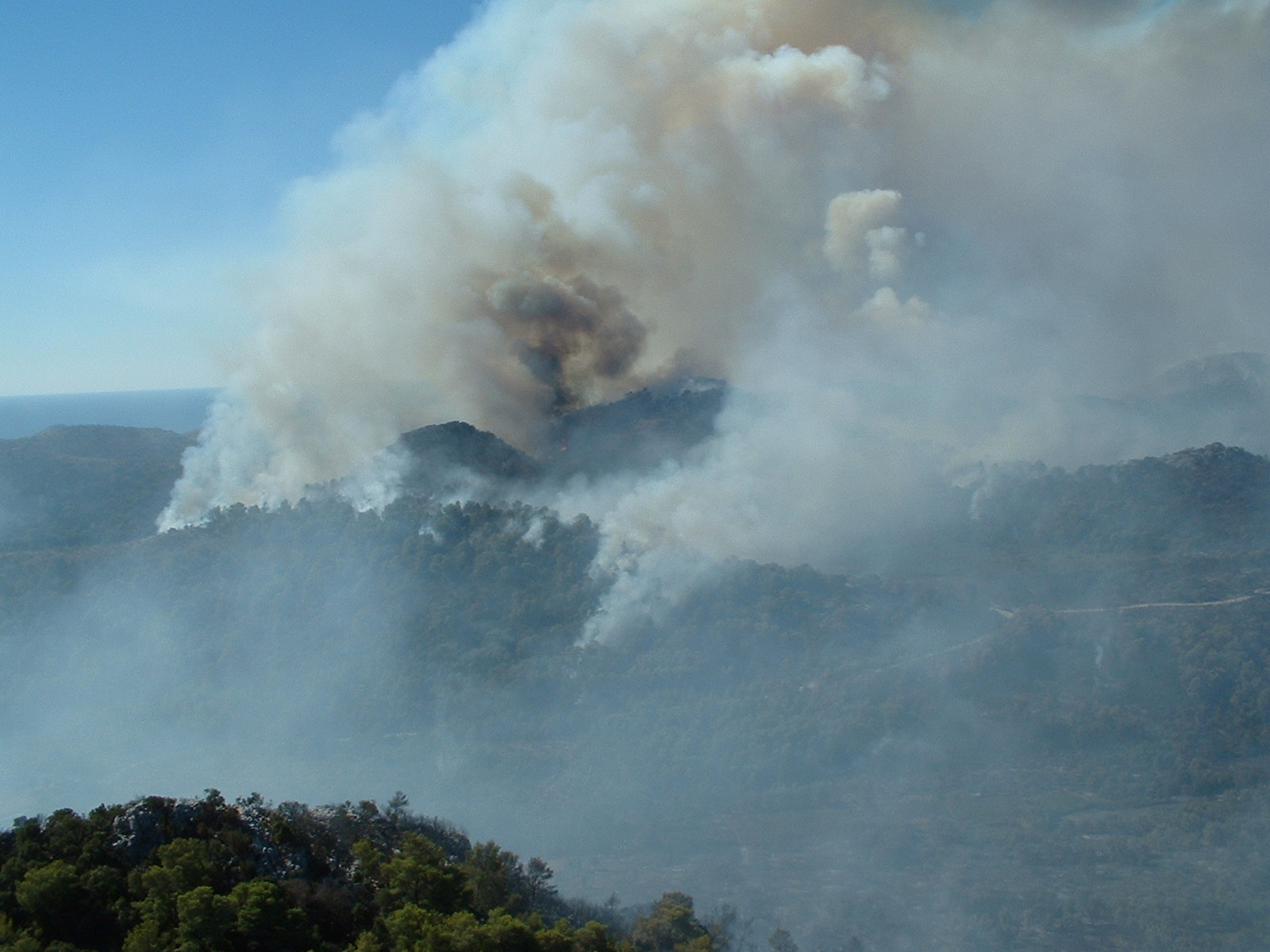}}&
   {\includegraphics[width=0.18\linewidth]{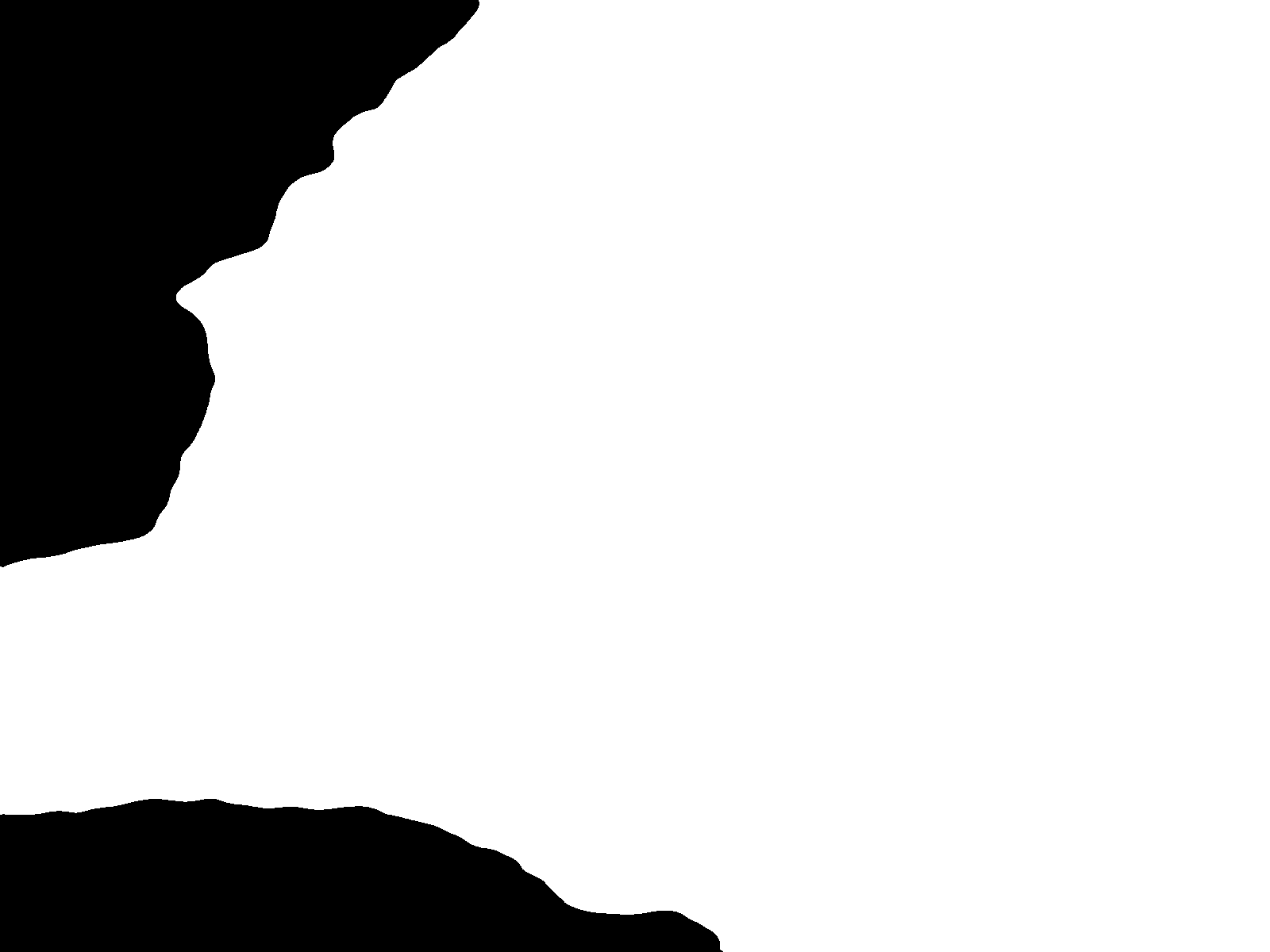}}&
   {\includegraphics[width=0.18\linewidth]{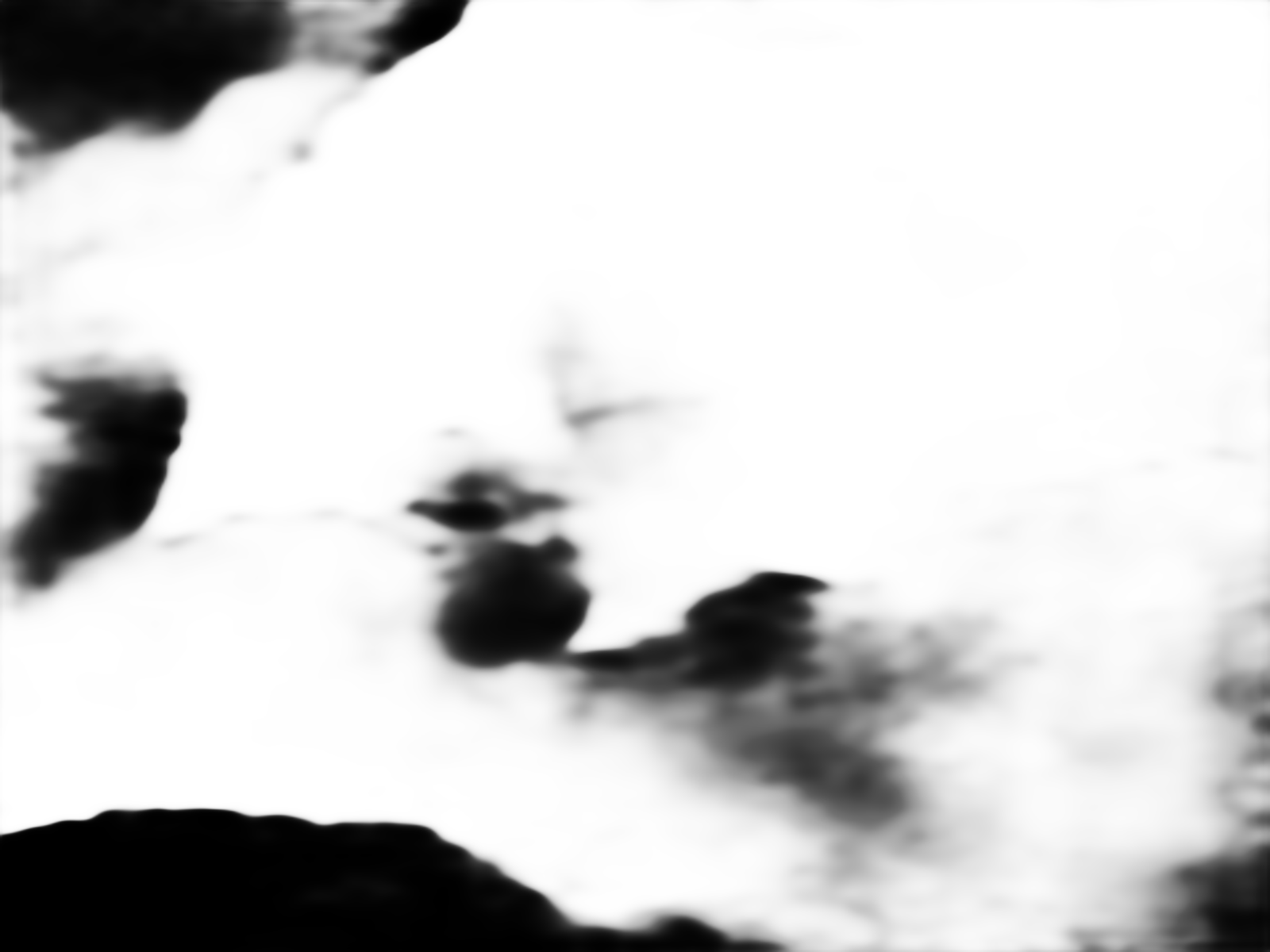}}&
   {\includegraphics[width=0.18\linewidth]{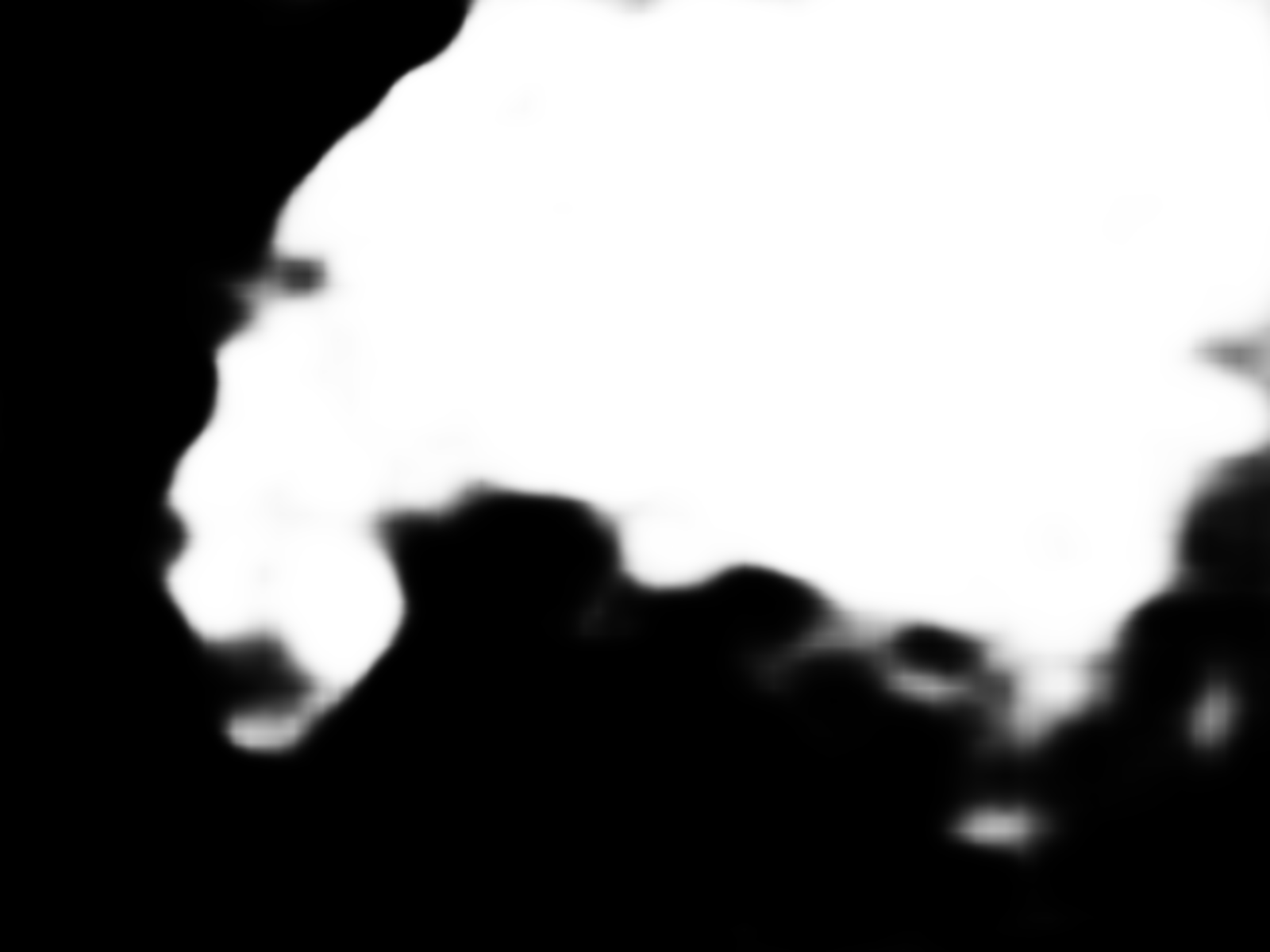}}&
   {\includegraphics[width=0.18\linewidth]{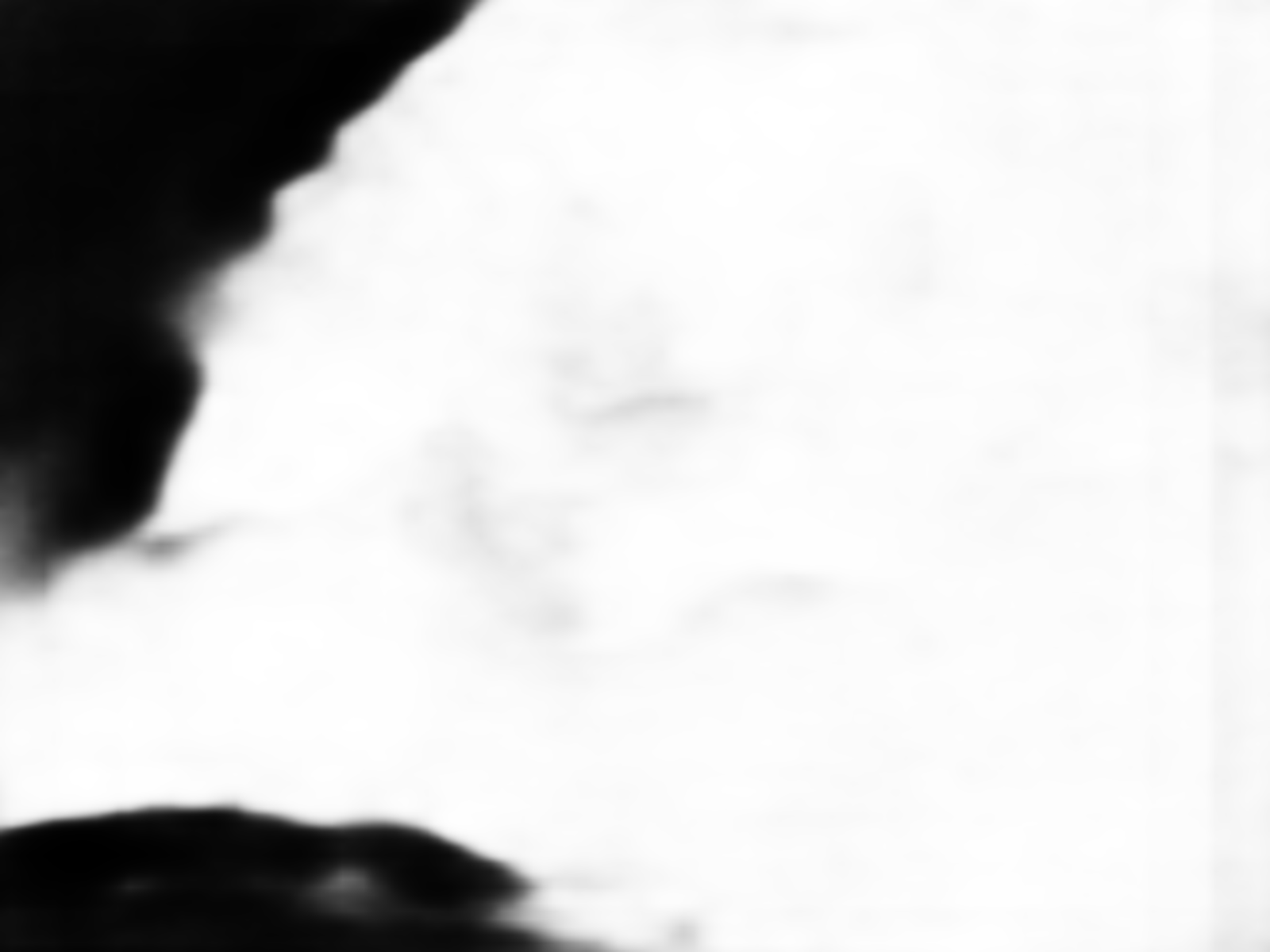}}
   \\
    {\includegraphics[width=0.18\linewidth]{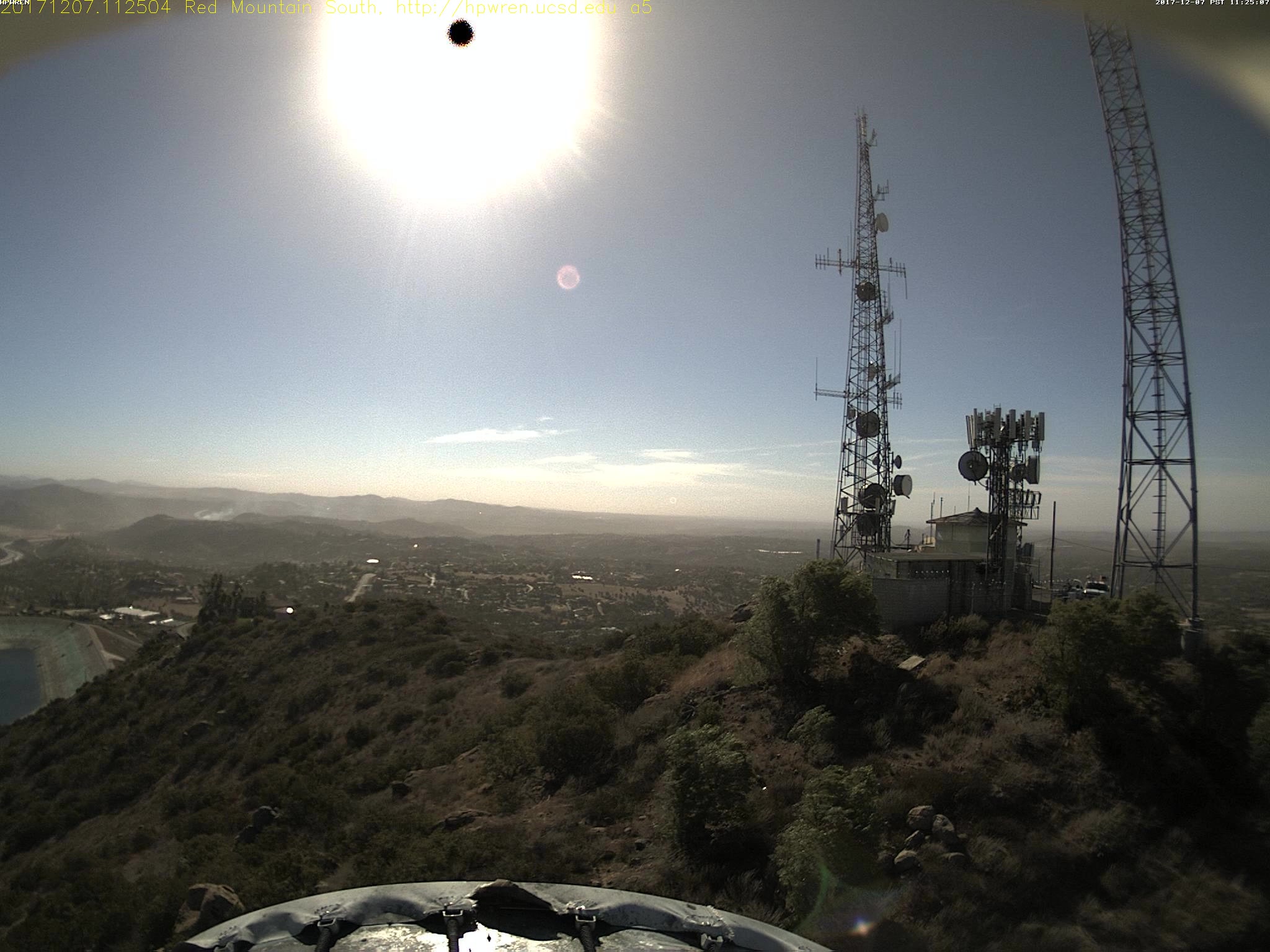}}&
   {\includegraphics[width=0.18\linewidth]{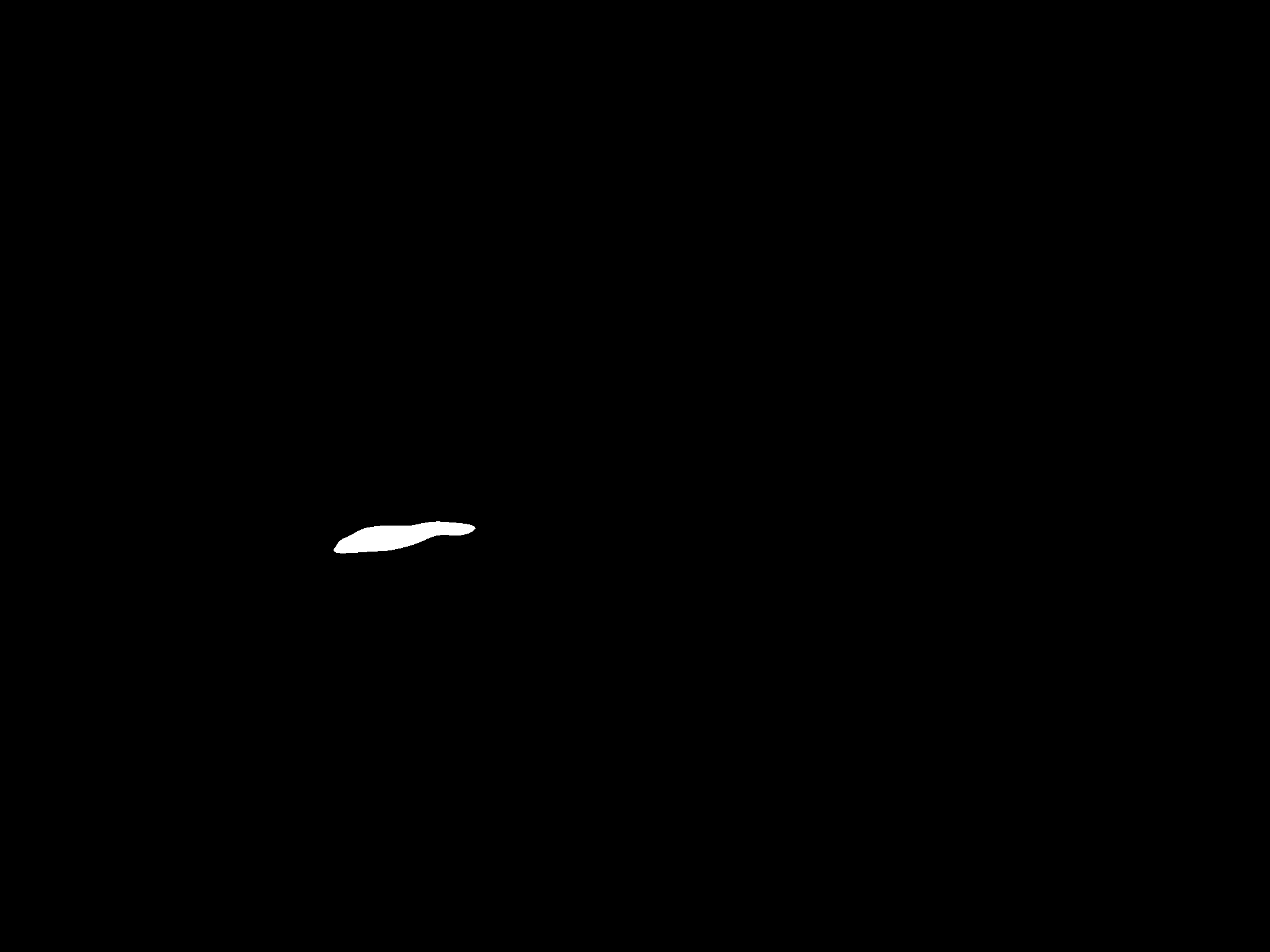}}&
   {\includegraphics[width=0.18\linewidth]{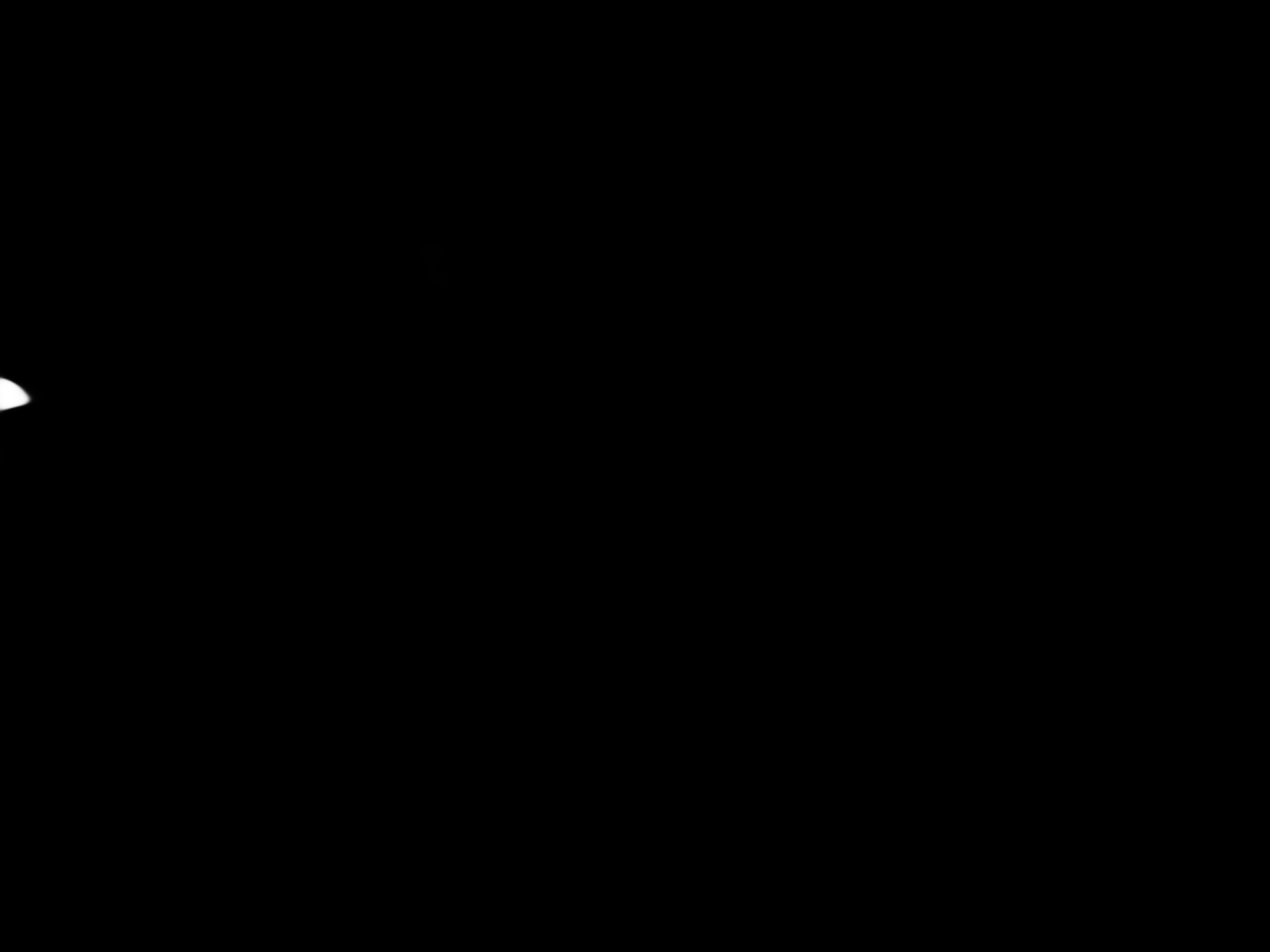}}&

   {\includegraphics[width=0.18\linewidth]{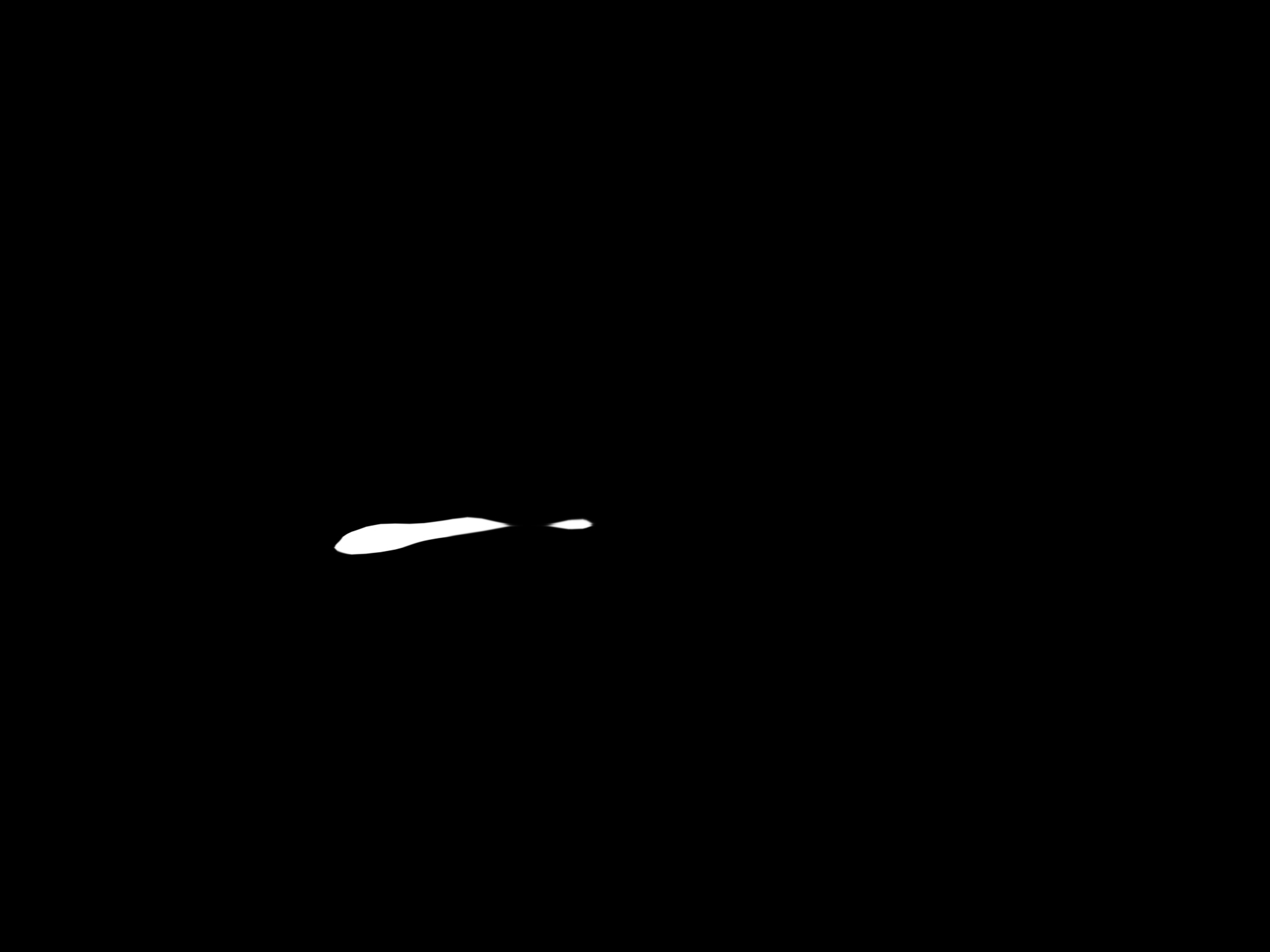}}&
   {\includegraphics[width=0.18\linewidth]{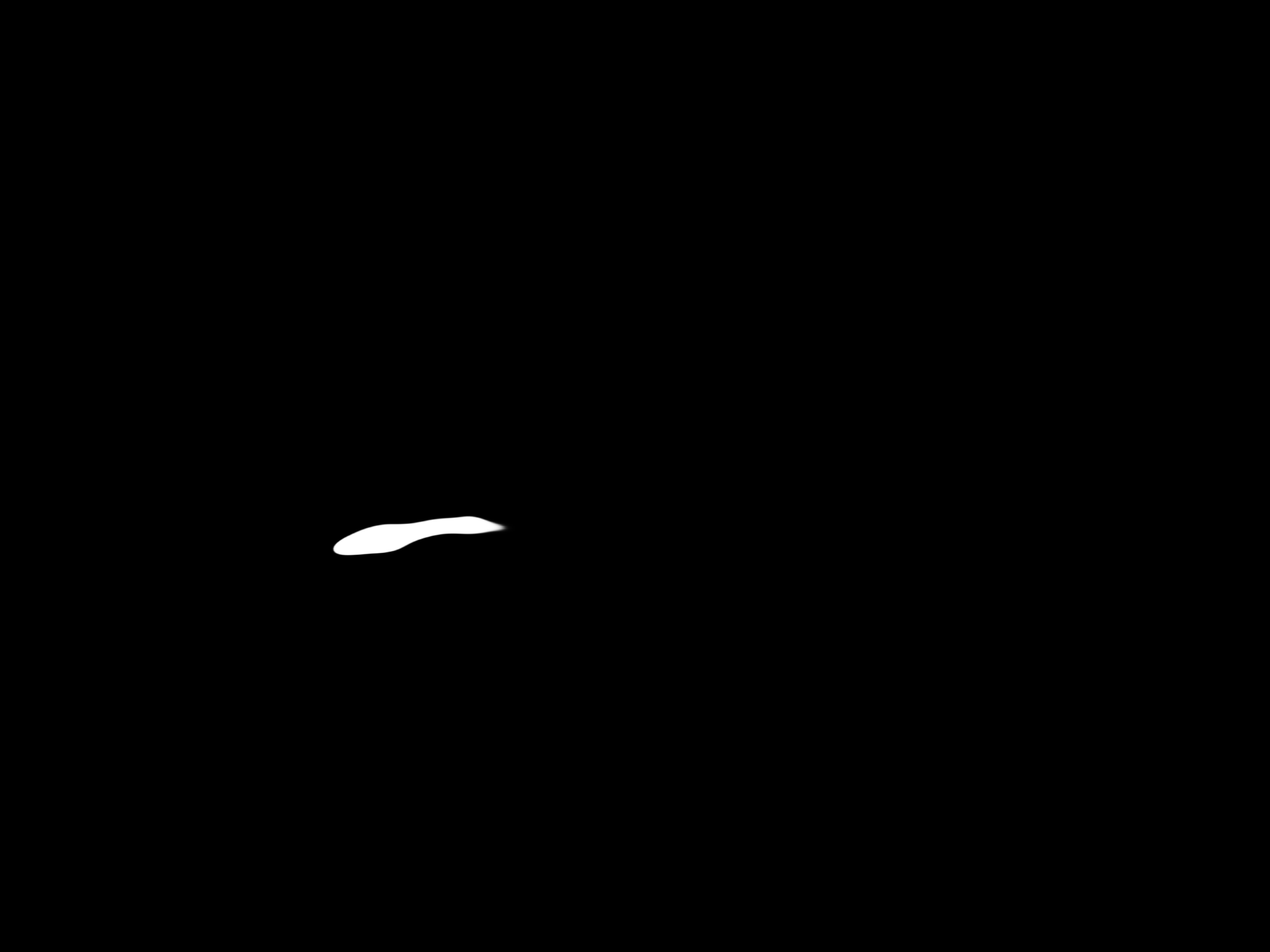}}
  \\

    {\includegraphics[width=0.18\linewidth=]{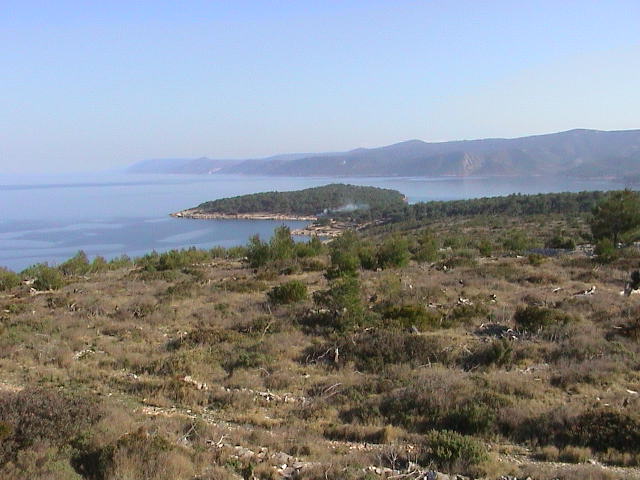}}&
   {\includegraphics[width=0.18\linewidth]{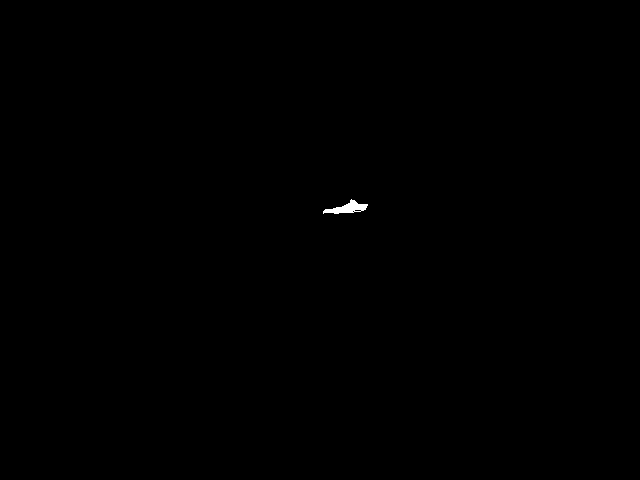}}&
   {\includegraphics[width=0.18\linewidth=]{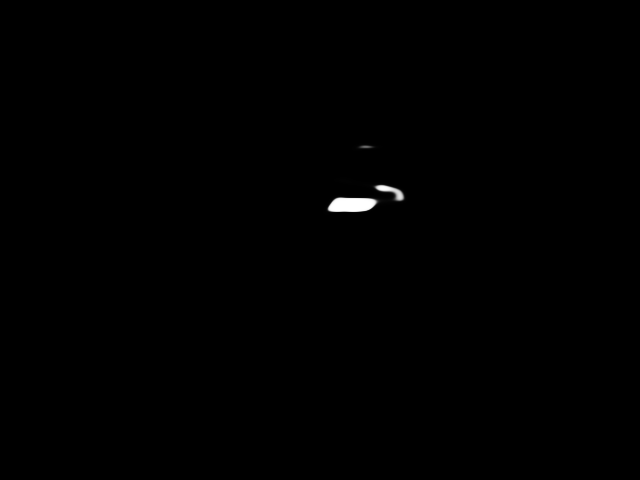}}&
   {\includegraphics[width=0.18\linewidth]{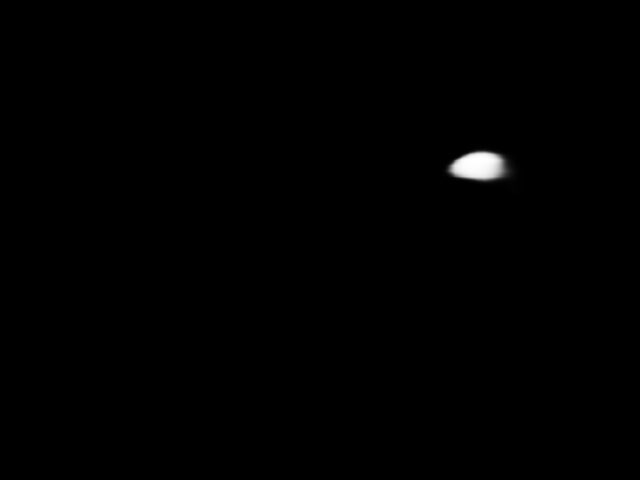}}&
   {\includegraphics[width=0.18\linewidth]{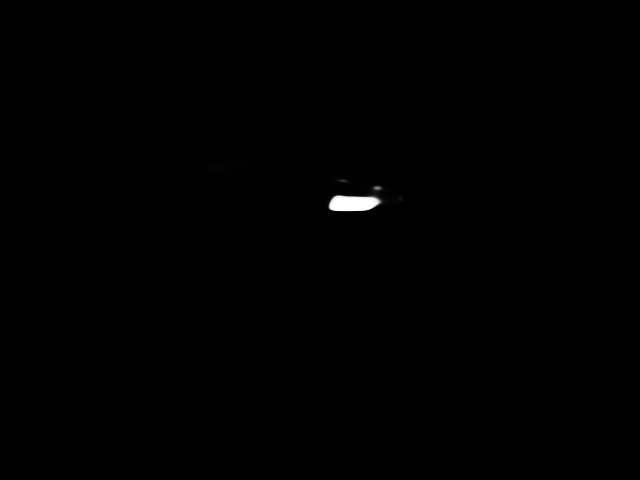}}\\
  \footnotesize{Image} & \footnotesize{GT} & \footnotesize{ITSD} & \footnotesize{UCNet} & \footnotesize{Ours} \\

  \end{tabular}
      \end{center}
    \vspace{-2mm}
      \caption{\normalsize Qualitative comparison of predictions by our model and other state-of-the-other models.}

      \label{fig:qualitative}

  \vspace{-4mm}
\end{figure}

\noindent\textbf{Quantitative Comparison:}
In Table \ref{tab2}, we compare our model with the state-of-the-art methods by training and testing on the synthetic dataset \cite{data1}, and the significant performance improvement of our model validates effectiveness of our solution.
% We can see that our model achieves a new state-of-the-art performance, improving over all previous approaches on all test sets by a large margin. Our model achieves a significant improvement for mMse, compared with the previous best smoke segmentation model CRGNet \cite{smoke3}.
% \subsubsection{Benchmarking on SMOKE5K.} 
 We also provide a new benchmark of state-of-the-art models trained and tested on our SMOKE5K dataset, shown in Table \ref{tab3}. We retrain all models on our training set with their original setting. For the ``Total (T)" test set, our model outperforms F3Net \cite{sloss}, BASNet \cite{BASNet_Sal}, SCRN \cite{scrnet}, and ITSD \cite{ITSD}, and slightly outperforms UCNet \cite{ucnet++}. For the ``Difficult (D)" test set, the consistently better performance against all state-of-the-art models
%  with a larger margin 
 demonstrates the effectiveness of our model for smoke segmentation.
%  the transparent and non-rigid shape issues.  
 
\noindent\textbf{Qualitative Comparison on SMOKE5K Dataset: }
We also visualize the predictions of our model and compared models, namely ITSD \cite{ITSD} and UCNet \cite{ucnet++} (the two best models in Table \ref{tab3}) in Fig.~\ref{fig:qualitative}.
% (More comparisons with other models are introduced in the supplementary material). 
In the first row, we find our prediction more complete and precise for the transparent part of the smoke that fills much of the image. In the second and third rows, we find that our model also performs better
% has better detection ability 
for small size of
% and transparent 
smoke.

\subsection{Uncertainty Decomposition}
\begin{figure}[!t]
   \begin{center}
   \begin{tabular}{ c@{ } c@{ } c@{ } c@{ }  c@{ }  c@{ } }

   {\includegraphics[width=0.15\linewidth]{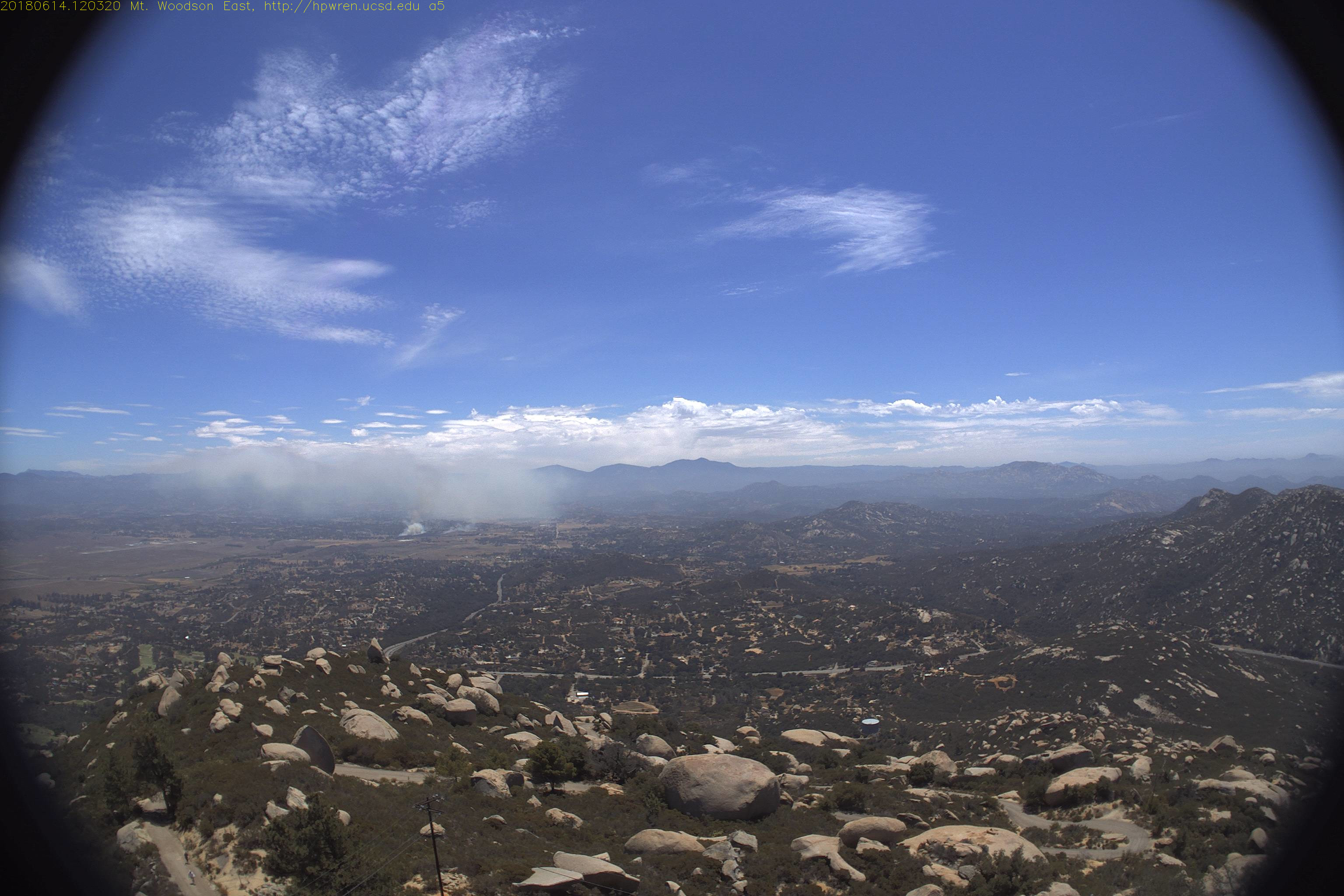}}&
   {\includegraphics[width=0.15\linewidth]{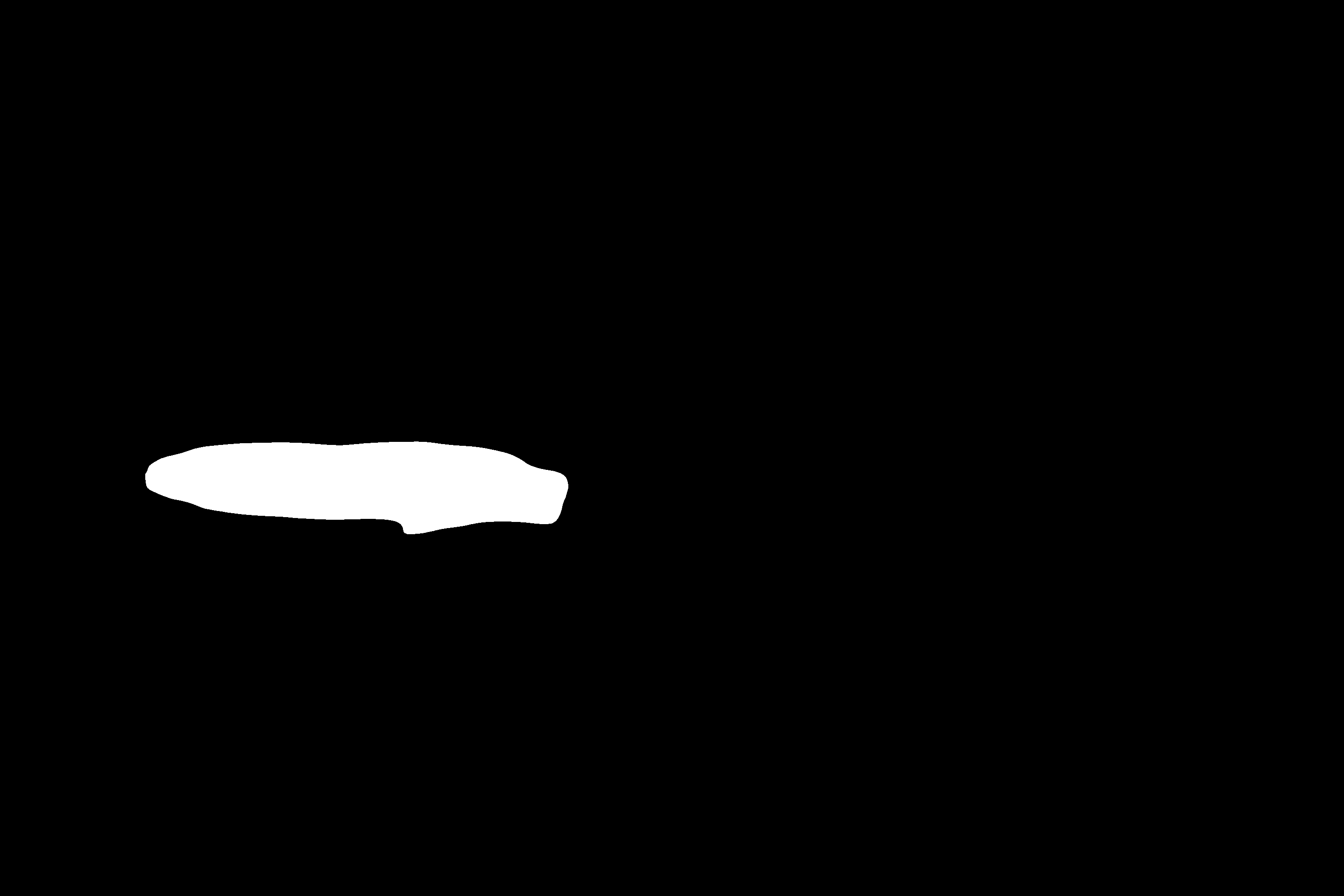}}&
   {\includegraphics[width=0.15\linewidth]{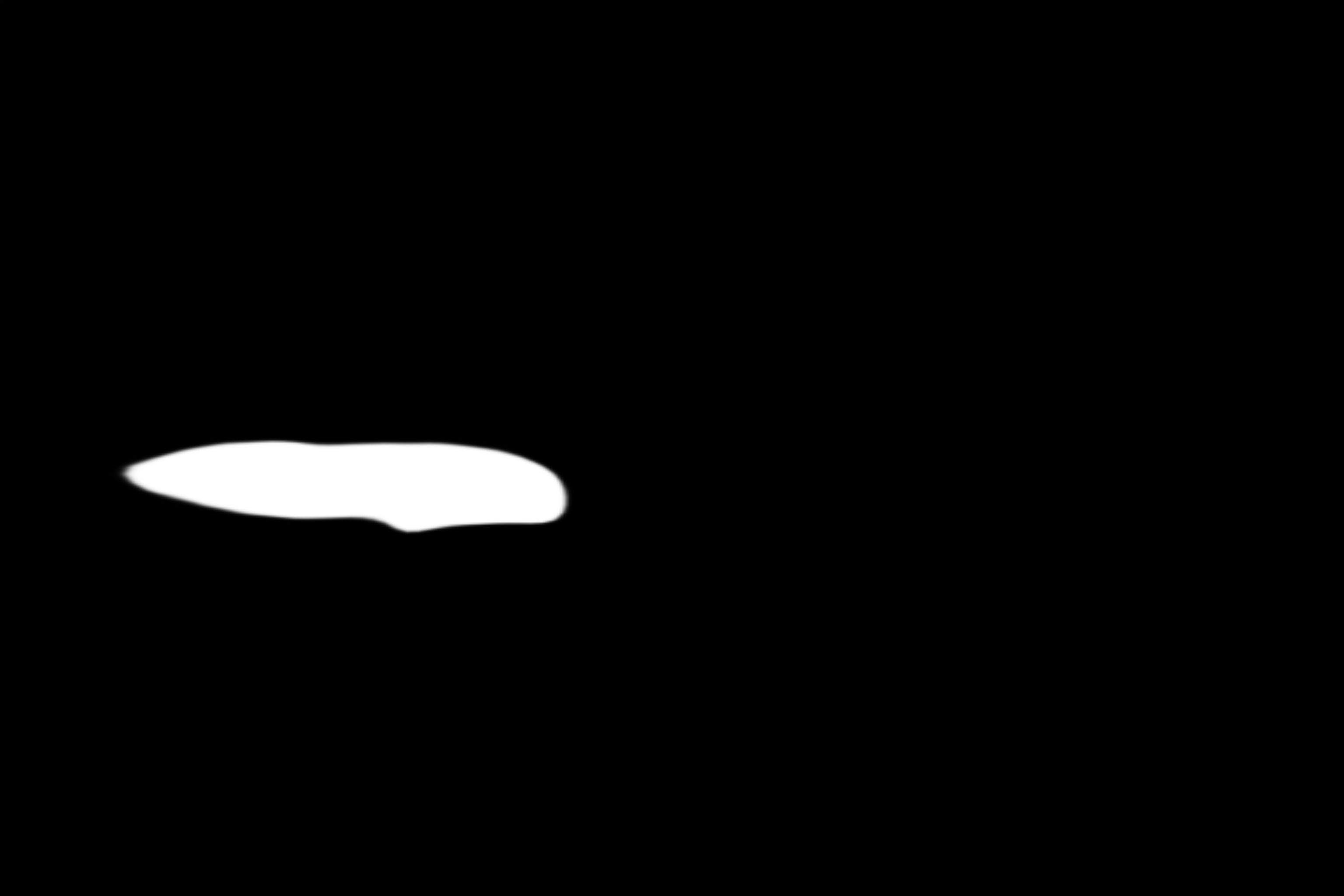}}&
   {\includegraphics[width=0.15\linewidth]{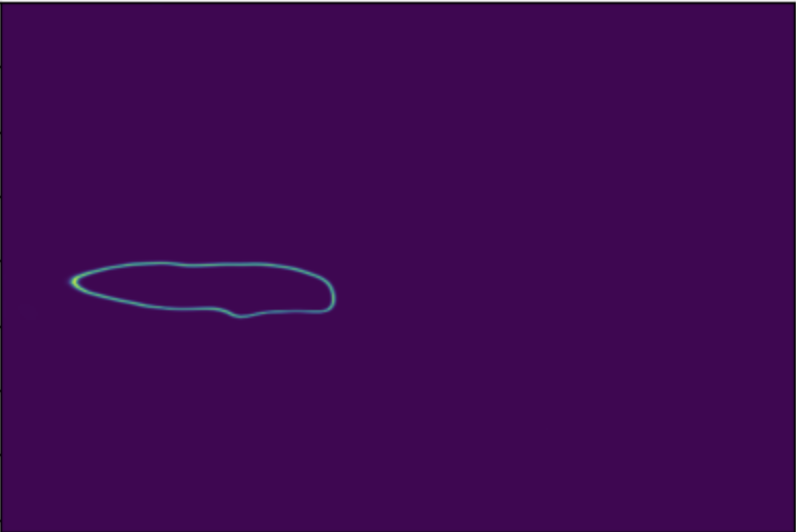}}&
   {\includegraphics[width=0.15\linewidth]{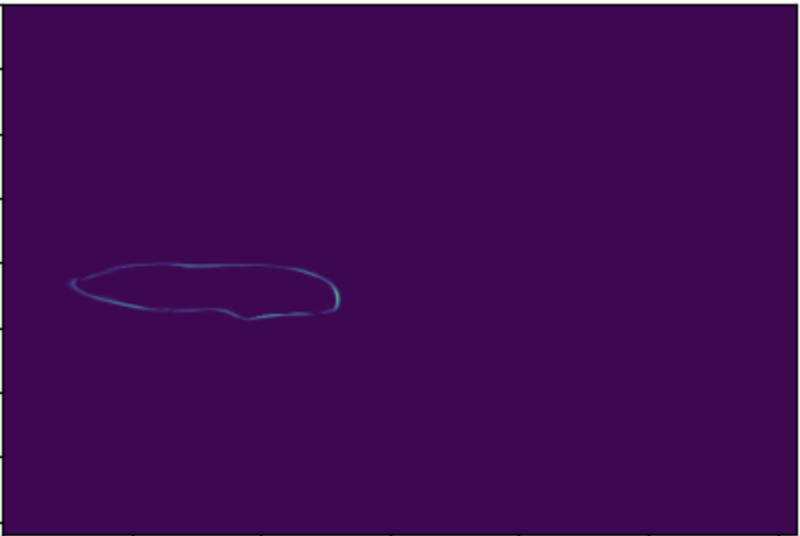}}&
   {\includegraphics[width=0.15\linewidth]{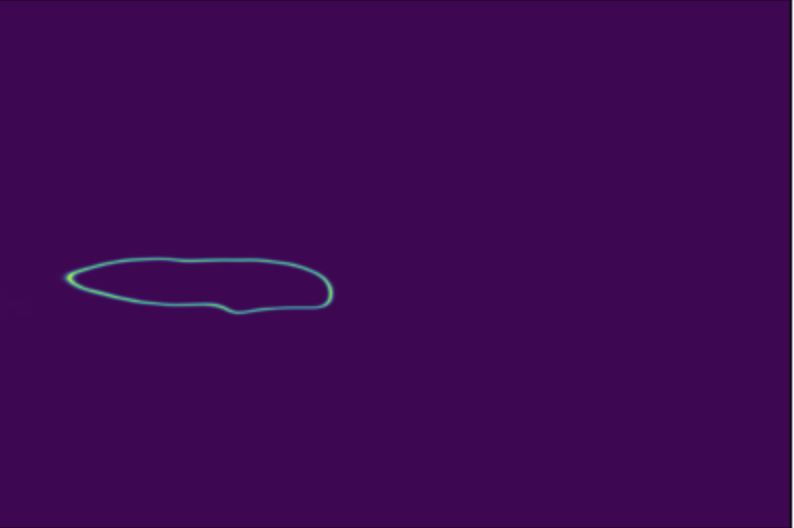}}\\
    {\includegraphics[width=0.15\linewidth]{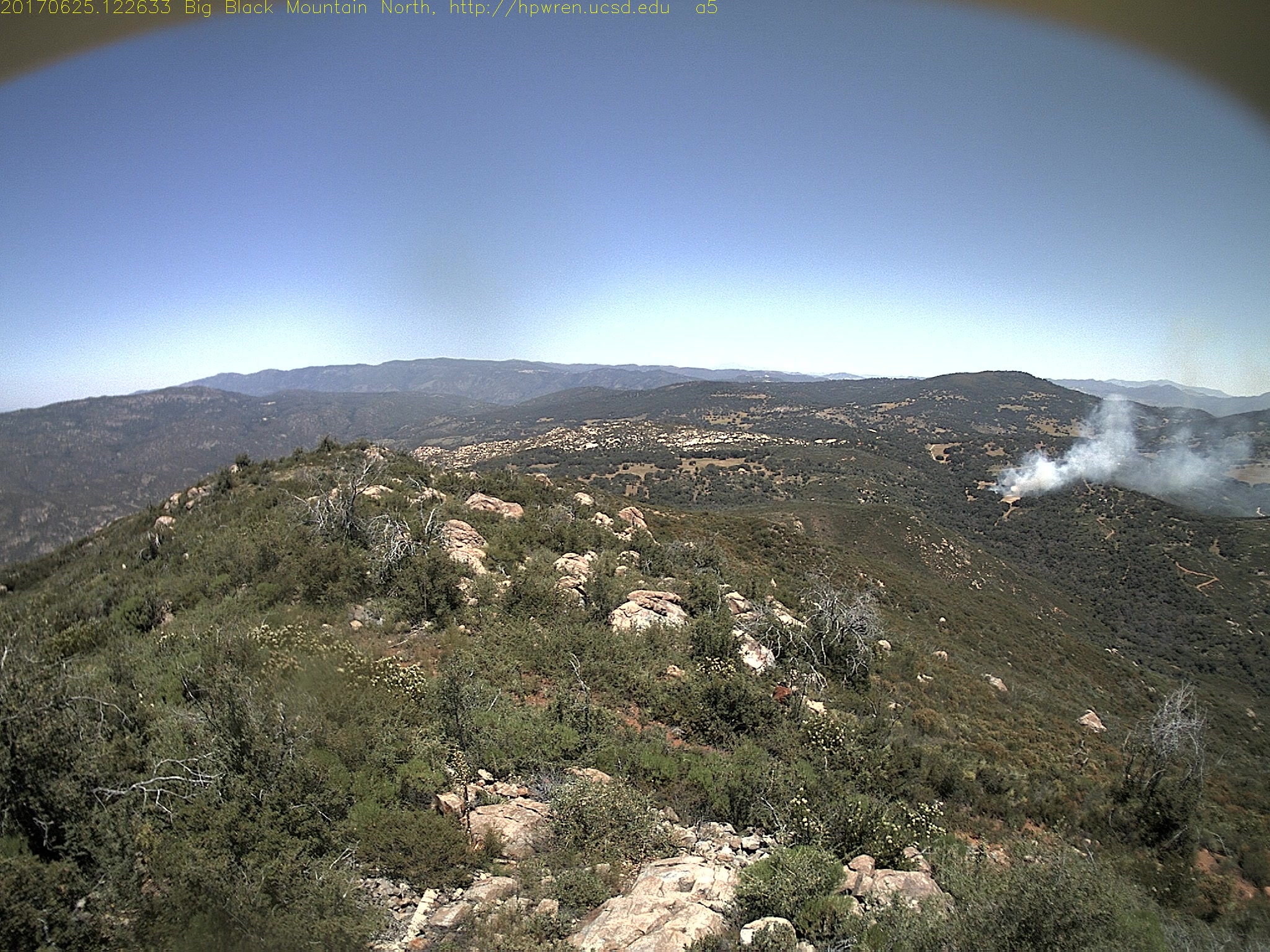}}&
   {\includegraphics[width=0.15\linewidth]{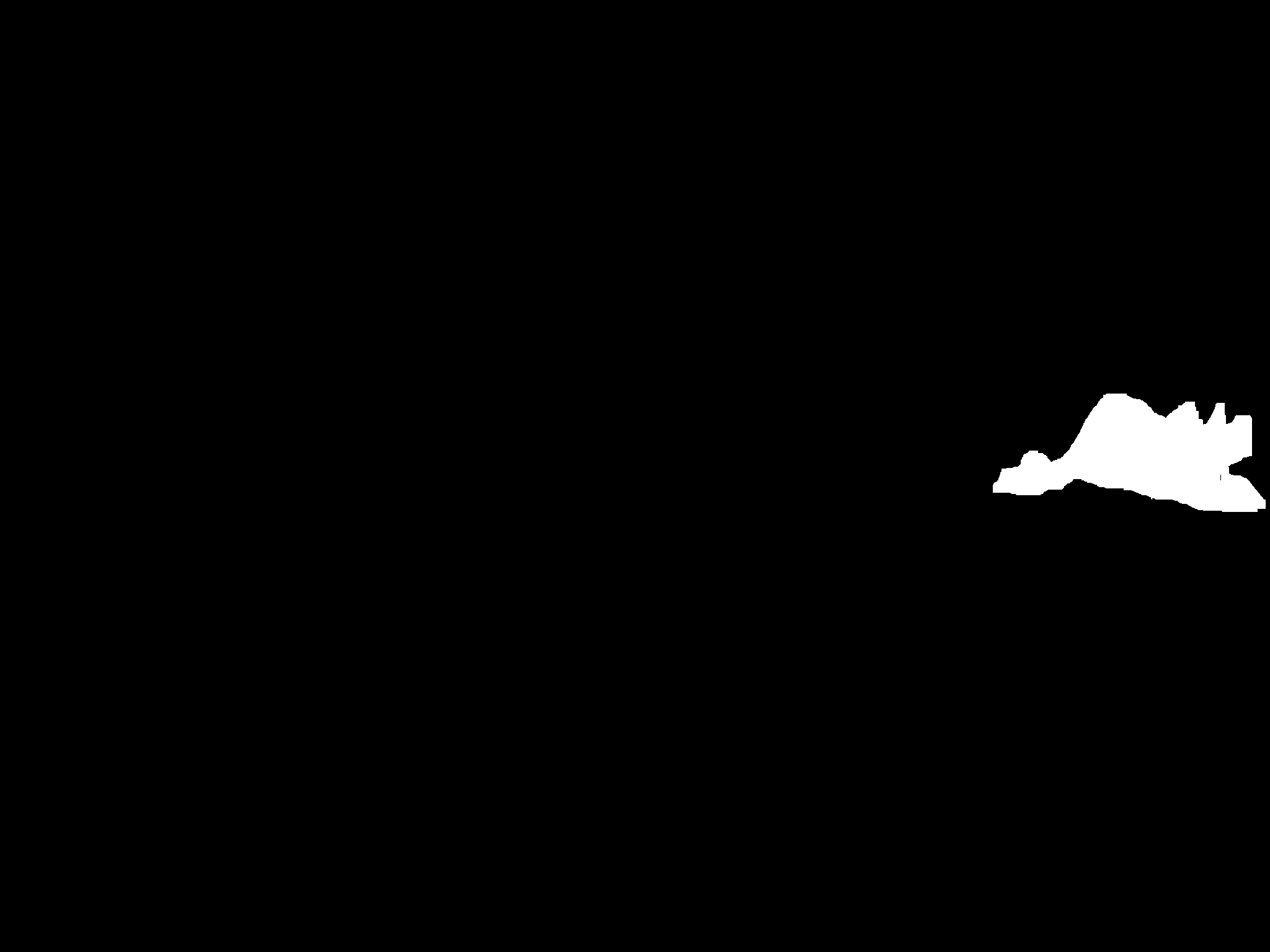}}&
   {\includegraphics[width=0.15\linewidth]{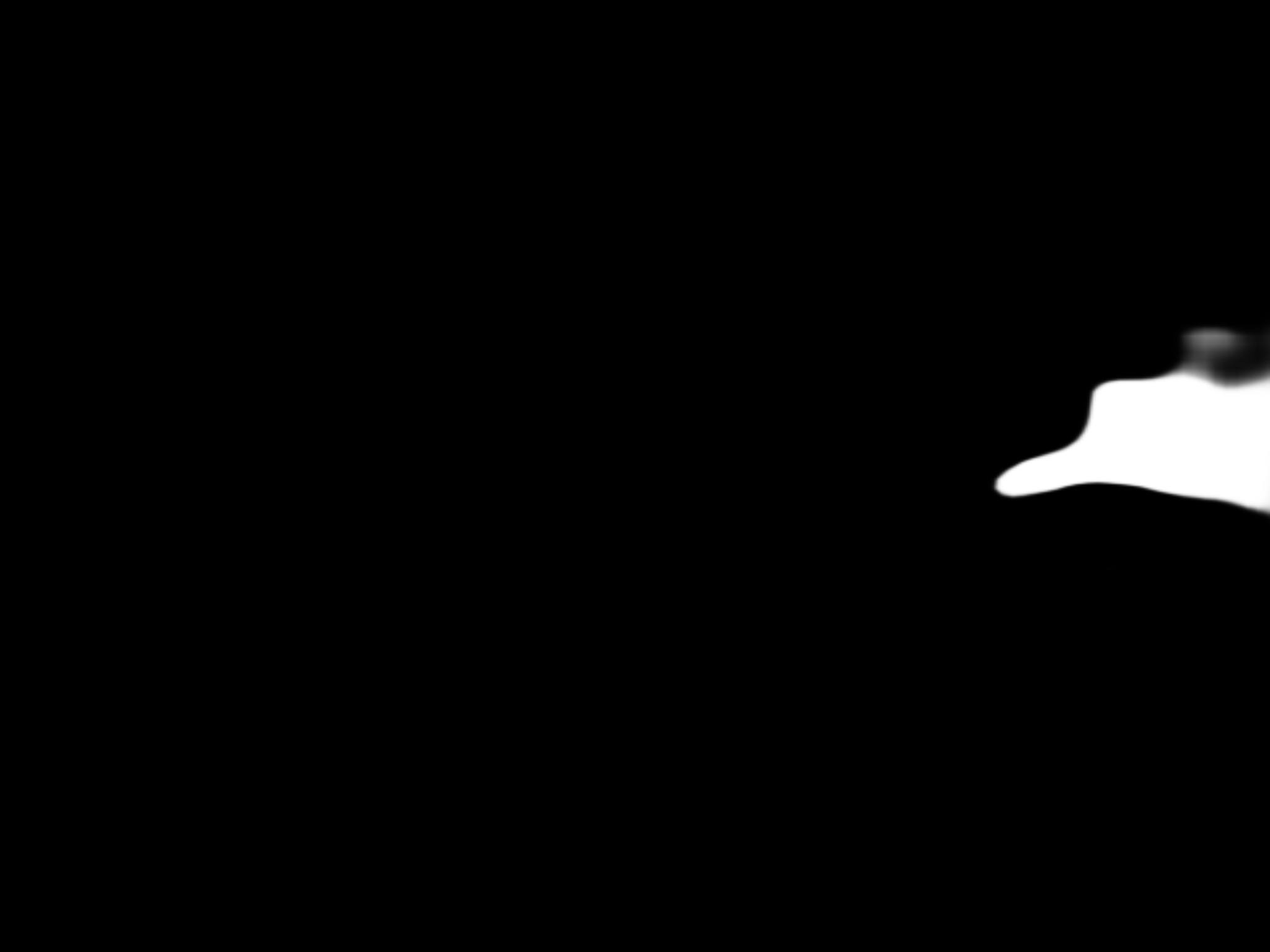}}&
   {\includegraphics[width=0.15\linewidth]{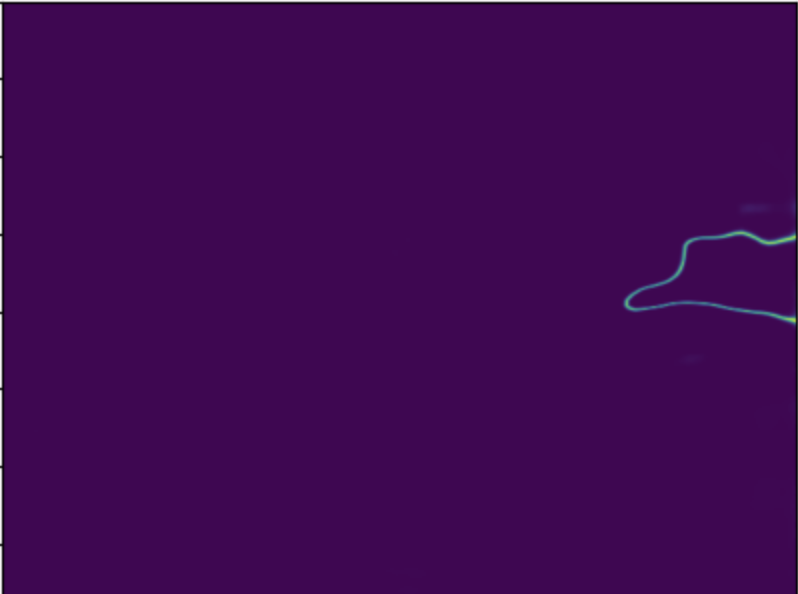}}&
   {\includegraphics[width=0.15\linewidth]{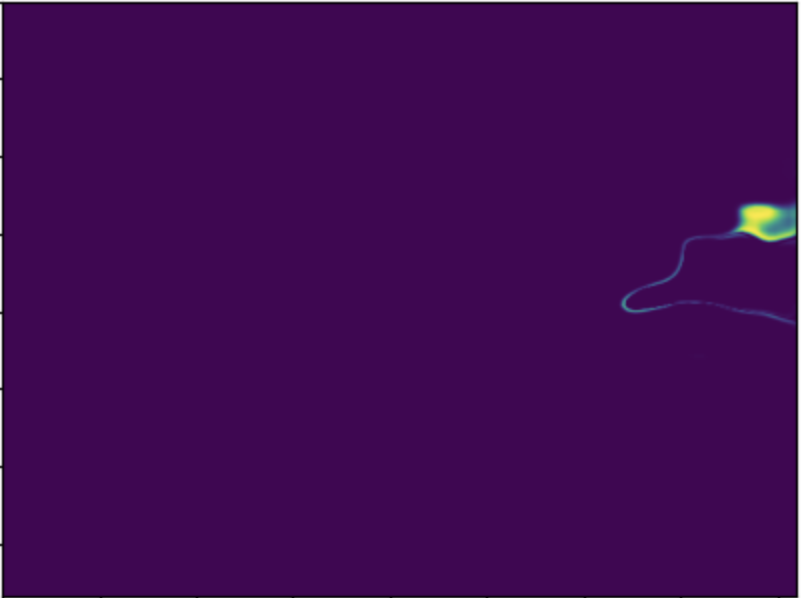}}&
   {\includegraphics[width=0.15\linewidth]{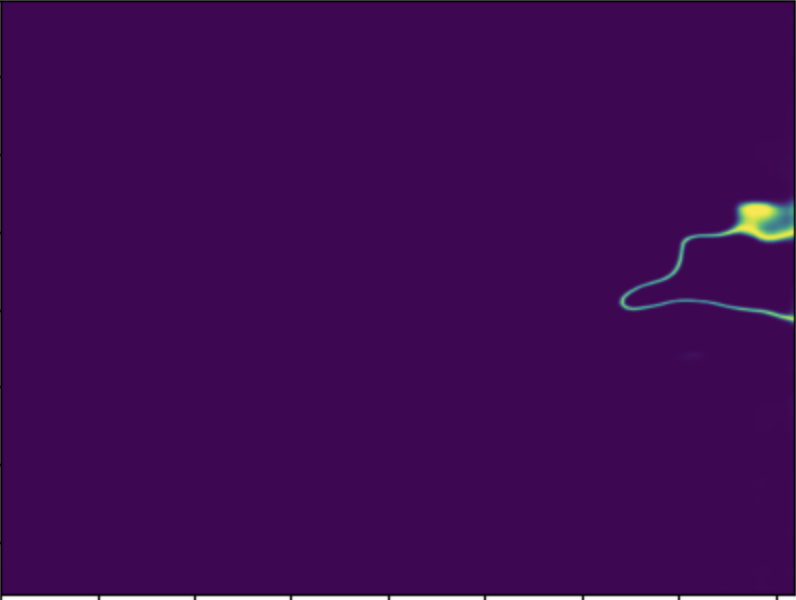}} \\

   {\includegraphics[width=0.15\linewidth]{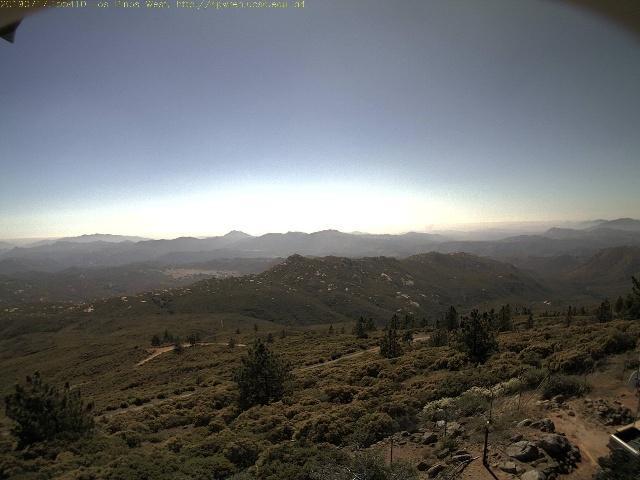}}&
   {\includegraphics[width=0.15\linewidth]{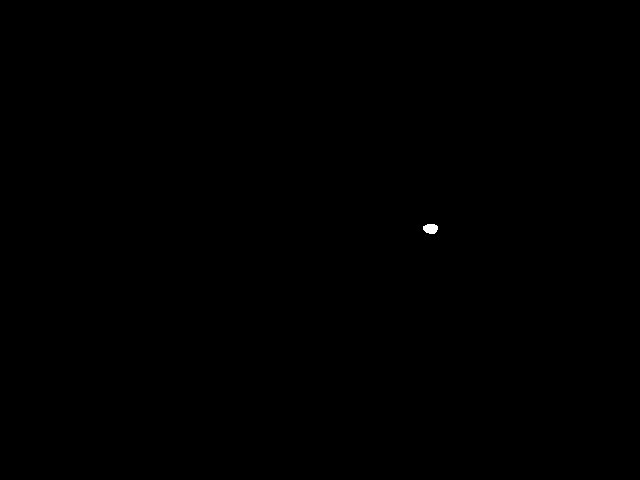}}&
   {\includegraphics[width=0.15\linewidth]{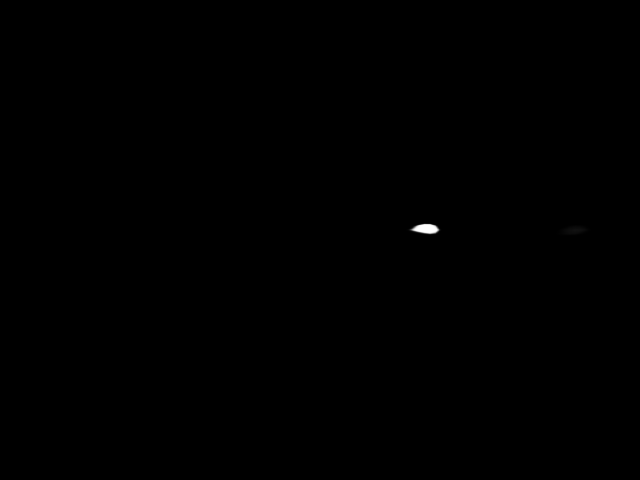}}&
   {\includegraphics[width=0.15\linewidth]{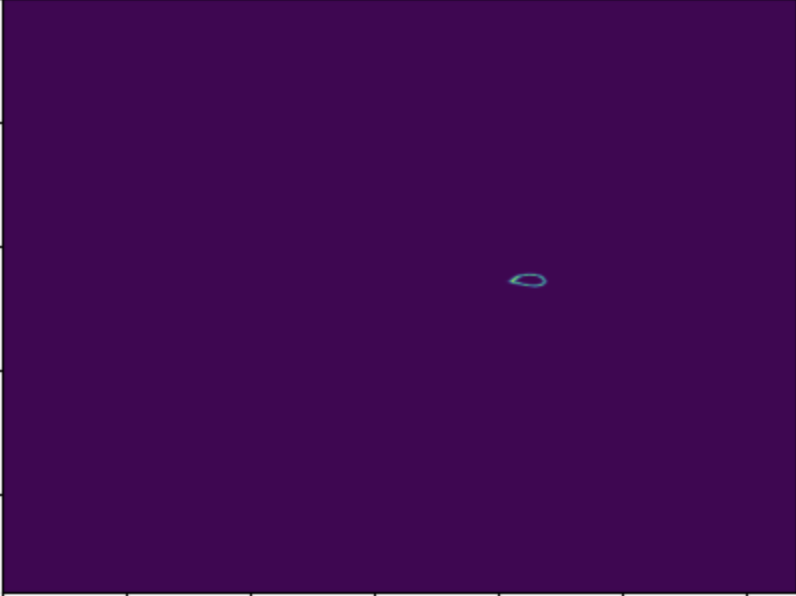}}&
   {\includegraphics[width=0.15\linewidth]{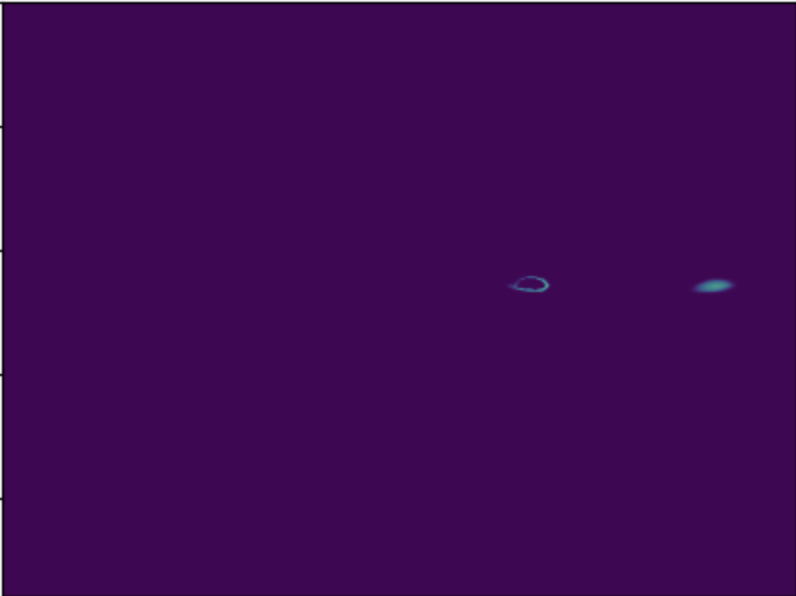}}&
   {\includegraphics[width=0.15\linewidth]{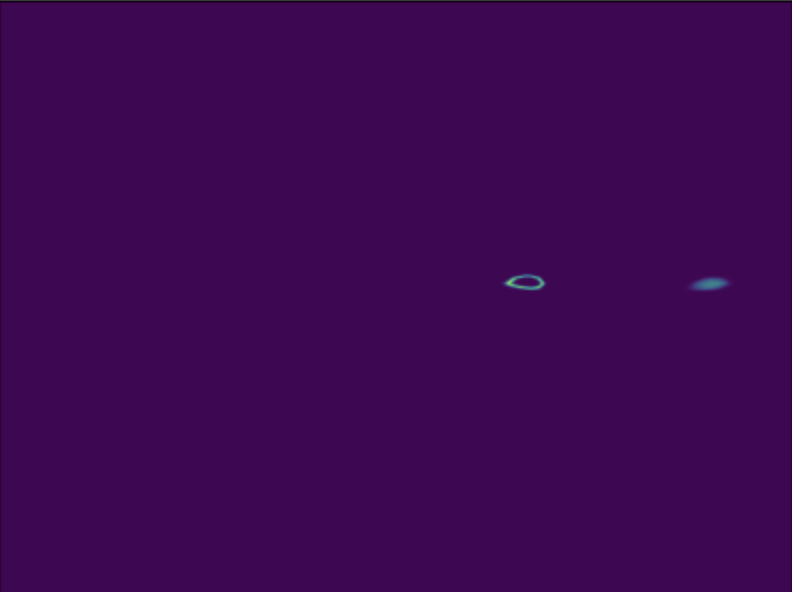}} \\

   \footnotesize{Img} & \footnotesize{GT} & \footnotesize{Pred} & \footnotesize{Aleatoric} & \footnotesize{Epistemic} & \footnotesize{Total}\\
   \end{tabular}
   
   \end{center}
 
    \vspace{-2mm}
\caption{ \normalsize Visualisation of three kinds of uncertainty captured by our model (Please zoom in to see the uncertainty map). From top to bottom row: Simple example, normal example, difficult example.}
 \vspace{-2mm}
   \label{uncertainty decomp}
%   \setlength{\abovecaptionskip}{0.cm}
% \setlength{\belowcaptionskip}{-0.cm}
%   \vspace{2mm}
\end{figure}

In Fig. \ref{uncertainty decomp}, we choose three representative images according to the difficulty to segment, denoted as simple, normal and difficult examples. We visualize our model predictions with the corresponding three types of uncertainties, namely \enquote{Aleatoric}, \enquote{Epistemic} and \enquote{Total} uncertainty. For the simple example (the first row), epistemic and aleatoric uncertainty capture similar patterns.
% , but aleatoric uncertainty is increased on the boundary. 
For the medium example (the second row), epistemic uncertainty is increased on the translucent region of the smoke, which accounts for the out-of-distribution example of our training sets. For the difficult example (the third row), epistemic uncertainty is especially increased in fog regions.
% , which shows that epistemic uncertainty increases objects with similar appearance. 
Further, we can see the total uncertainty captures both aleatoric uncertainty and epistemic uncertainty. In summary, we find that it is important to model aleatoric uncertainty for: (1) modeling ambiguity in annotations, and (2) modeling the non-rigid shape of objects. Further, it is important to model epistemic uncertainty for: (1) safety-critical applications, identifying out-of-distribution samples;
% , it can provide more information rather than only the model's predictions, guiding people to make better decision-making; 
and (2) the false-positive problems, where the epistemic uncertainty is usually increased in false-positive regions. The total uncertainty in general can
% the sum of aleatoric and epistemic uncertainty,
provide more information about model prediction, leading to explainable and well-calibrated models.
% informative results for uncertainty, which can guide people to know when to trust the model's prediction. Also, it is also can be used to improve model performance by uncertainty calibration.
 %\NB{This sentence is broken in grammar, and I'm not sure what you are trying to say}.

\subsection{Ablation Study}
We perform the following experiments to analyze each component of our model. All experiments are evaluated on the test set of SMOKE5K. To evaluate the effectiveness of our method, we set our baseline by removing the transmission loss and the uncertainty calibration entropy loss, which is shown as \enquote{1} in Table \ref{tab4}. Note that it means our baseline does not exploit estimated uncertainty. Then, we combine the transmission loss and uncertainty calibration entropy loss gradually, denoted in \enquote{2} and \enquote{4} in Table \ref{tab4}.
% Further, we discuss the impact of hyperparameter and inference speed in supplementary material.
% In Table \ref{tab4}, we see that the best performance is obtained by combining all components of our model, which demonstrates the importance of each component.
% Also, consistently greater improvement in the ``Difficult" test set shows that the effectiveness of our method on translucent and non-rigid shape problems.

\begin{table}[t!]
  \centering
  \small
  \renewcommand{\arraystretch}{1.0}
  \renewcommand{\tabcolsep}{2.0mm}

  \begin{tabular}{lr|c|c|c|c|ccc}
  \hline
%   \toprule
%   &  &\multicolumn{14}{c|}{Fully Sup. Models}&\multicolumn{1}{c}{Weakly Sup./Unsup. Models} \\
    &&Base.& $\mathcal{L}_{trans}$&$\mathcal{L}_{en}$&$\mathcal{L}_{c}$&$\mathcal{M}\downarrow$&$F_{\beta}\uparrow$ \\

  \hline
    
    \multirow{4}{*}  {\textit{T}}
     &1&\checkmark && & &.005&.712 \\
%   \hline
    % \multirow{1}{*}{2 }
     &2&\checkmark&\checkmark && &.004&.741 \\
%   \hline 

     &3&\checkmark& \checkmark & \checkmark&& .006&.695\\

     &4&\checkmark& \checkmark & &\checkmark& \textbf{.002}&\textbf{.791}\\
   \hline

    \multirow{4}{*} {\textit{D}}
     &1&\checkmark & & &&.014&.646 \\
%   \hline

     &2&\checkmark&\checkmark & &&.008&.714 \\
%   \hline 

     &3&\checkmark& \checkmark &\checkmark& & .007&.726\\

     &4&\checkmark& \checkmark & &\checkmark& \textbf{.006}&\textbf{.741}\\
   \hline

  \end{tabular}
    \caption{\normalsize Ablation study for transmission loss, entropy loss and uncertainty calibration mechanism.}
  \label{tab4}
\end{table}

\noindent\textbf{Impact of the Transmission Loss:}
We perform this ablation study by adding transmission loss to our baseline, shown as \enquote{2} in Table \ref{tab4}. Compared with baseline, we find transmission loss helps the model achieve a performance improvement on both \enquote{Total} and \enquote{Difficult} test sets. The greater improvement on the \enquote{Difficult} test set demonstrates our loss is useful for dealing with smoke's transparent appearance. Further, in Fig. \ref{fig1_}, it can be seen that the model with transmission loss (\enquote{w/T}) can better recognize ambiguous regions and deal with low contrast than the model without the transmission loss (\enquote{w/o T}), which demonstrates the effectiveness of the proposed transmission loss.

\begin{figure}[t]
    \centering
   
%  	\newcolumntype{T}{>{\hsize=0.125\textwidth}X}
    \begin{tabular}{ c@{ }c@{ }}
    % \subfigure{
    % \begin{minipage}[t]{0.48\linewidth}
    % \centering

    \includegraphics[width=1\linewidth]{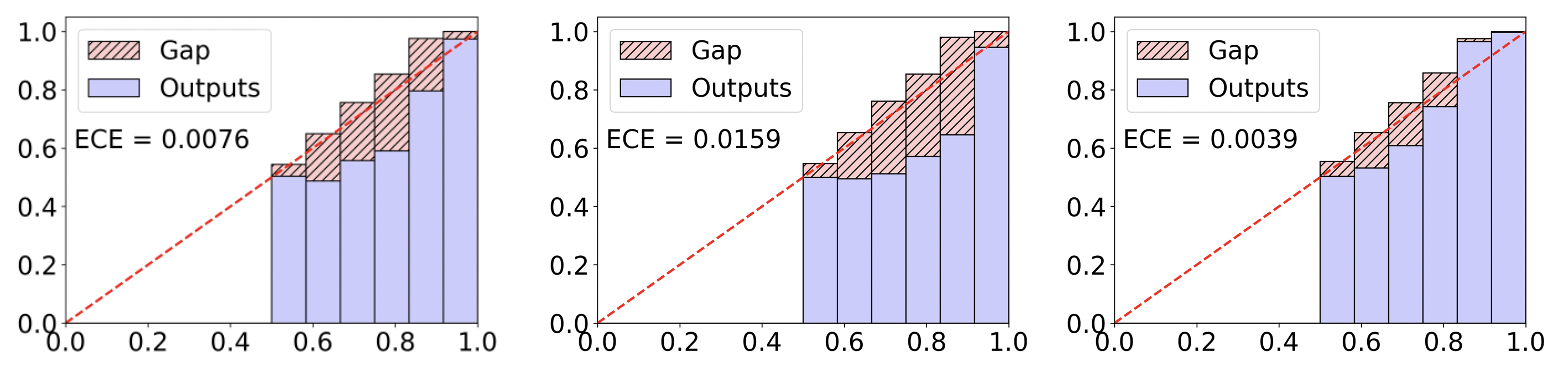}\\

    \end{tabular}
    % \end{minipage}
    % }
    \caption{\normalsize Reliability diagrams for UCNet (the $1^{st}$), our model before calibration (the $2^{nd}$), and our model after calibration (the $3^{rd}$). The horizontal axis of the reliability diagram denotes confidence, and the vertical axis denotes the accuracy.}
    % \vspace{-2mm}
    \label{fig6}
\end{figure}

\noindent\textbf{Impact of Uncertainty Calibration:}
In this ablation study, we set our baseline as our model with transmission loss (\enquote{2} in Table \ref{tab4}). Based on it, we introduce entropy loss $\mathcal{L}_{en}$ to it to validate the uncertainty calibration entropy loss, and denote this experiment as \enquote{3} in Table \ref{tab4}.
% we compare our new baseline (2), baseline with entropy loss $\mathcal{L}_{en}$, which reflects removing the temperature $U_{p}$ in the uncertainty calibration loss $L_{c}$ in Eq.\ref{eq10}. (3), and baseline with our uncertainty calibration entropy loss (4)
% % , shown in 2, 3, and 4 
% in Table \ref{tab4}. 
We can see that the performance for the \enquote{Total} test set is worse when we directly add entropy loss. The reason is that although entropy loss can encourage predictions to be binary, it will make predictions to be over-confident, and this negative influence will be significant for our \enquote{Total} test set as most smoke regions are very small. Then, we add uncertainty-based temperature scaling into entropy loss as in Eq.~\ref{eq10}.
% , which becomes our uncertainty calibration entropy loss in Equation \ref{eq10}. 
We can see that the performance in all metrics is higher than the previous two models, which clearly shows the advantage of the proposed uncertainty calibration loss $\mathcal{L}_c$ in Eq.~\ref{eq10}.
% Thus, we demonstrate the effectiveness of our loss is not only due to the advantage of entropy loss. 

Further, as an uncertainty-aware learning method, we aim to achieve well calibrated model where model accuracy is consistent with model confidence. To measure the
% to evaluate the model's 
calibration quality, we follow the calibration definition in \cite{on_calibration}, and report the reliability diagrams in Fig. \ref{fig6}.
% , we say a model is well-calibrated when the model's prediction (confidence) is equal to its accuracy. 
% We show the reliability diagrams of UCNet, 
We show the reliability diagrams of our model without calibration (setting 2 in Table \ref{tab4}) and our model after calibration (setting 4 in Table \ref{tab4}) in the second and third diagrams of Fig. \ref{fig6}, respectively. We also show reliability diagrams of UCNet \cite{ucnet++} (another uncertainty estimation method using CVAEs \cite{cvae} for saliency detection) in the first diagram of Fig. \ref{fig6}.
% UCNet \cite{ucnet++} is a state-of-the-art stochastic saliency detection model that can model the uncertainty during labeling, namely the aleatoric uncertainty. 
It can be seen that the gap between accuracy and confidence for UCNet is smaller than our model without calibration, leading to the ECE of 0.0076, but the calibration quality is still very low. With our uncertainty calibration entropy loss, the gap between confidence and accuracy for our model is significantly reduced, the dense calibration measure (ECE) is also reduced from 0.0159 to 0.0039, leading to a better calibrated model.
% which achieves better confidence estimation. Also, compared with UCNet, our model achieves better calibration quality in terms of ECE.

\section{Conclusion}

In this paper, we propose a Bayesian generative model to perform smoke segmentation and quantify the corresponding informative uncertainties. We also explore the medium transmission feature and propose a novel transmission loss to tackle smoke's ambiguous boundary and low contrast problems. Further, we release the first high-quality, large-scale smoke segmentation dataset to promote the development of this field. Experiments on all benchmark test sets demonstrate the effectiveness of the proposed method, leading to both an accurate smoke segmentation model and reliable uncertainty maps indicating model's confidence towards it's prediction.

\vspace{1mm}
%\paragraph
\noindent
\section{Acknowledgments}
This research was supported by funding from the ANU-Optus Bushfire Research Center of Excellence.

\bibliography{aaai22}

% \section{Acknowledgments}
% AAAI is especially grateful to Peter Patel Schneider for his work in implementing the original aaai.sty file, liberally using the ideas of other style hackers, including Barbara Beeton. We also acknowledge with thanks the work of George Ferguson for his guide to using the style and BibTeX files --- which has been incorporated into this document --- and Hans Guesgen, who provided several timely modifications, as well as the many others who have, from time to time, sent in suggestions on improvements to the AAAI style. We are especially grateful to Francisco Cruz, Marc Pujol-Gonzalez, and Mico Loretan for the improvements to the Bib\TeX{} and \LaTeX{} files made in 2020.

% The preparation of the \LaTeX{} and Bib\TeX{} files that implement these instructions was supported by Schlumberger Palo Alto Research, AT\&T Bell Laboratories, Morgan Kaufmann Publishers, The Live Oak Press, LLC, and AAAI Press. Bibliography style changes were added by Sunil Issar. \verb+\+pubnote was added by J. Scott Penberthy. George Ferguson added support for printing the AAAI copyright slug. Additional changes to aaai22.sty and aaai22.bst have been made by Francisco Cruz, Marc Pujol-Gonzalez, and Mico Loretan.

% \bigskip
% \noindent Thank you for reading these instructions carefully. We look forward to receiving your electronic files!

\maketitle
\section{Overview}
In this supplementary material, we provide more details about our dataset, evaluation metrics, qualitative comparisons, dark channel prior algorithm, and some discussions about model details.
\section{Evaluation Metrics}
% In this section, we provide detail about the Mean square error ($\mathcal{M}$), F-measure ($F_{\beta}$), and Dense Calibration Measure ($ECE$) \cite{ece}.
% \subsection{ Mean Square Error}
% \subsection{ F-measure}

{\bf Mean Square Error:}
% \textbf{Saliency prediction network (SPN):}
$\mathcal{M}$ is defined as the per-pixel square difference between the prediction $s$ and the ground truth $y$. The average Mse ($mMse$) is usually used to evaluate model performance on test set, which is in the form of:
\begin{equation}
    mMse=\frac{1}{N} \sum_{i=1}^{w}\sum_{j=1}^{h}(s_{ij}-y_{ij})^{2}   \label{eq1}    
\end{equation}
where N is the number of pixels in the image, $w$ and $h$ are the width and height of the image, $s$ and $y$ are the prediction and ground truth. 

{\bf F-measure:}
F-measure combines both precision and recall to capture both properties, which is calculated based on each pair of precision and recall as:

\begin{equation}
    F_{\beta}=\frac{(1+\beta^2)\cdot precision \cdot recall}{\beta^2 \cdot precision + recall}
\end{equation}
where $\beta^{2}$ is set to 0.3 to assign more weight to precision.

{\bf Dense Calibration Measure: } $\mathcal{M}$ and $F_{\beta}$ only evaluate the model accuracy but ignore the gap between model accuracy and its confidence, which cannot evaluate how the model is calibrated in a dataset. 
The dense calibration measure is an extension of \cite{on_calibration}, which is the weighted average of the difference between every bin's accuracy and confidence. For each image $x_{i}$, the prediction $y_{i}$ is grouped into M interval bins, the dense calibration error is then defined as:

\begin{equation}
    ECE^{i}=\sum_{m=1}^{M}\frac{|B_{m}|}{\sum_{m}|B_{m}|}|macc(B_{m})^{i}-conf(B_{m})^{i}|
\end{equation}
where $B_{m}$ is the pixels in the $m$-th interval bin, $macc(B_{m})^{i}$ and $conf(B_{m})^{i}$ are the average accuracy and average confidence of the $m$-th interval bin, and $\sum_{m}|B_{m}|$ is the number of pixels in the image $x_{i}$. When $ECE^{i}=0$ ($macc(B_{m})^{i}=conf(B_{m})^{i}$ ) for each image $x_{i}$ in a dataset, the model is called perfectly calibrated model in this specific dataset.

\section{Dataset }

\subsection{Dataset Description}
\begin{figure*}[t]
   \begin{center}
     \renewcommand{\arraystretch}{0.2}
   \begin{tabular}{ c@{ } c@{ } c@{ } c@{ }  c@{ }   }

   {\includegraphics[width=0.178\linewidth, height=0.119\linewidth]{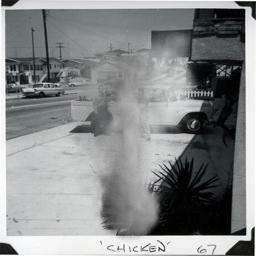}}&
   {\includegraphics[width=0.178\linewidth, height=0.119\linewidth]{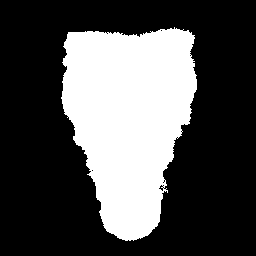}}&
   {\includegraphics[width=0.178\linewidth, height=0.119\linewidth]{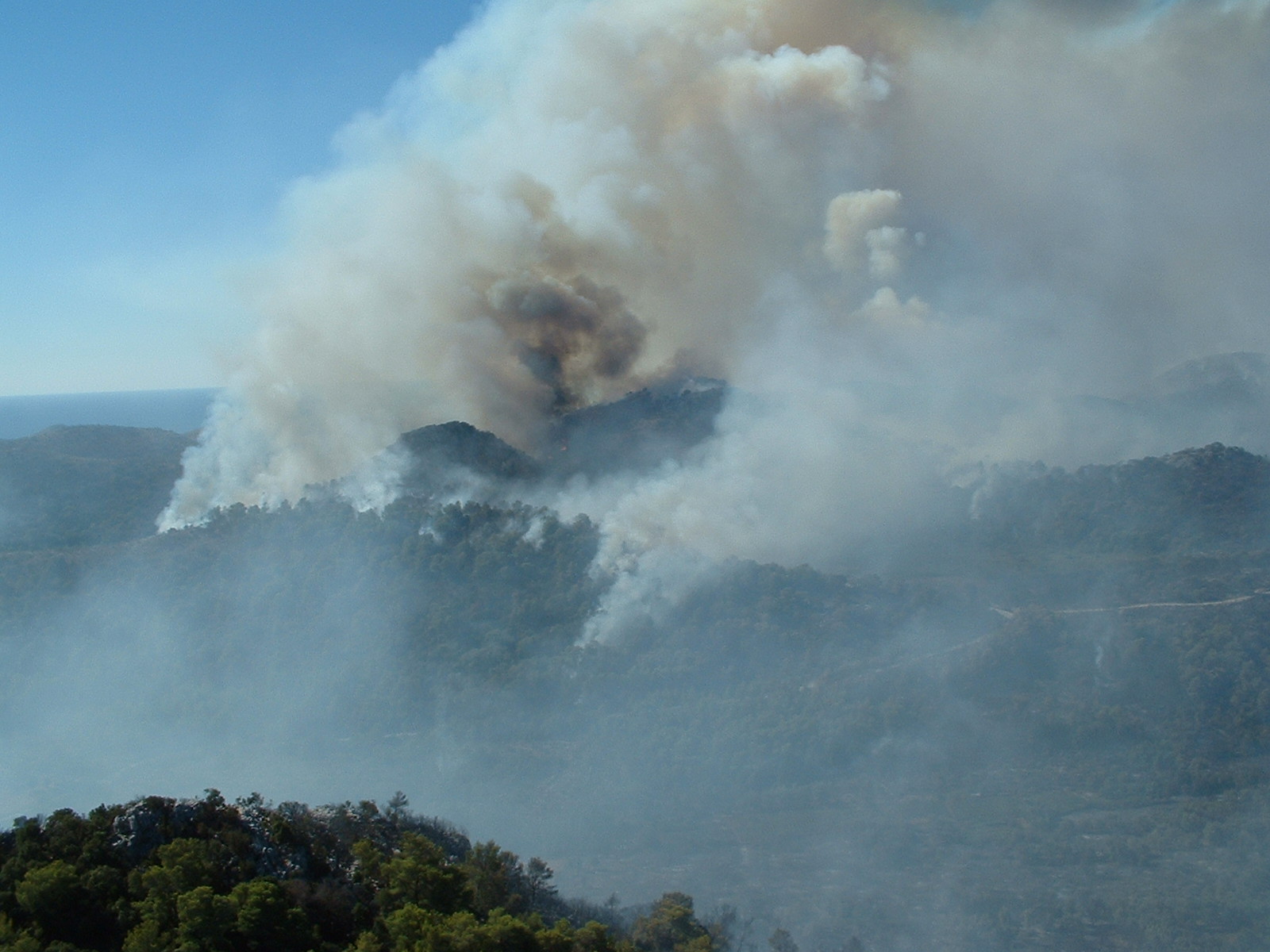}}&
   {\includegraphics[width=0.178\linewidth, height=0.119\linewidth]{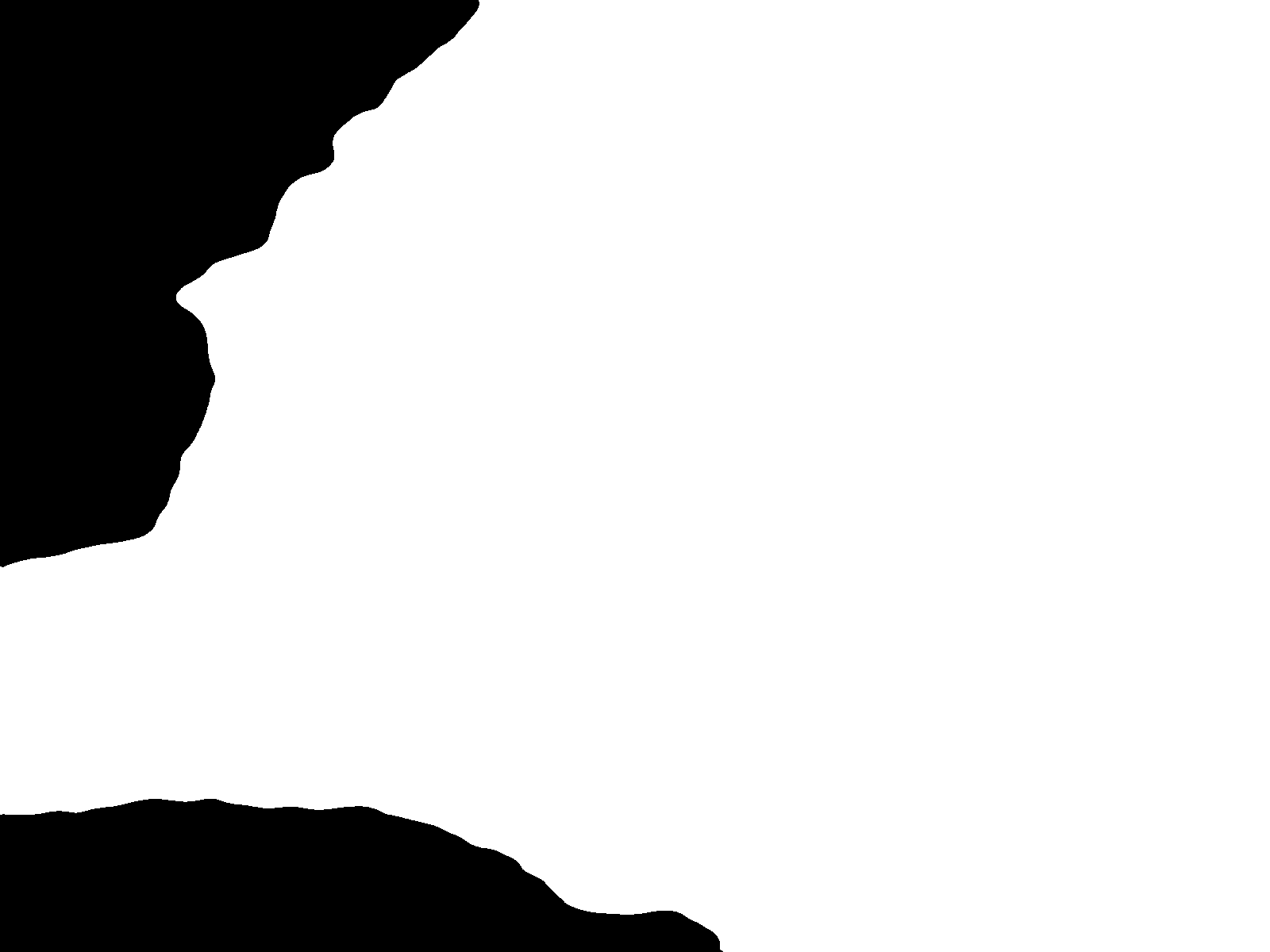}}&
   {\includegraphics[width=0.178\linewidth, height=0.119\linewidth]{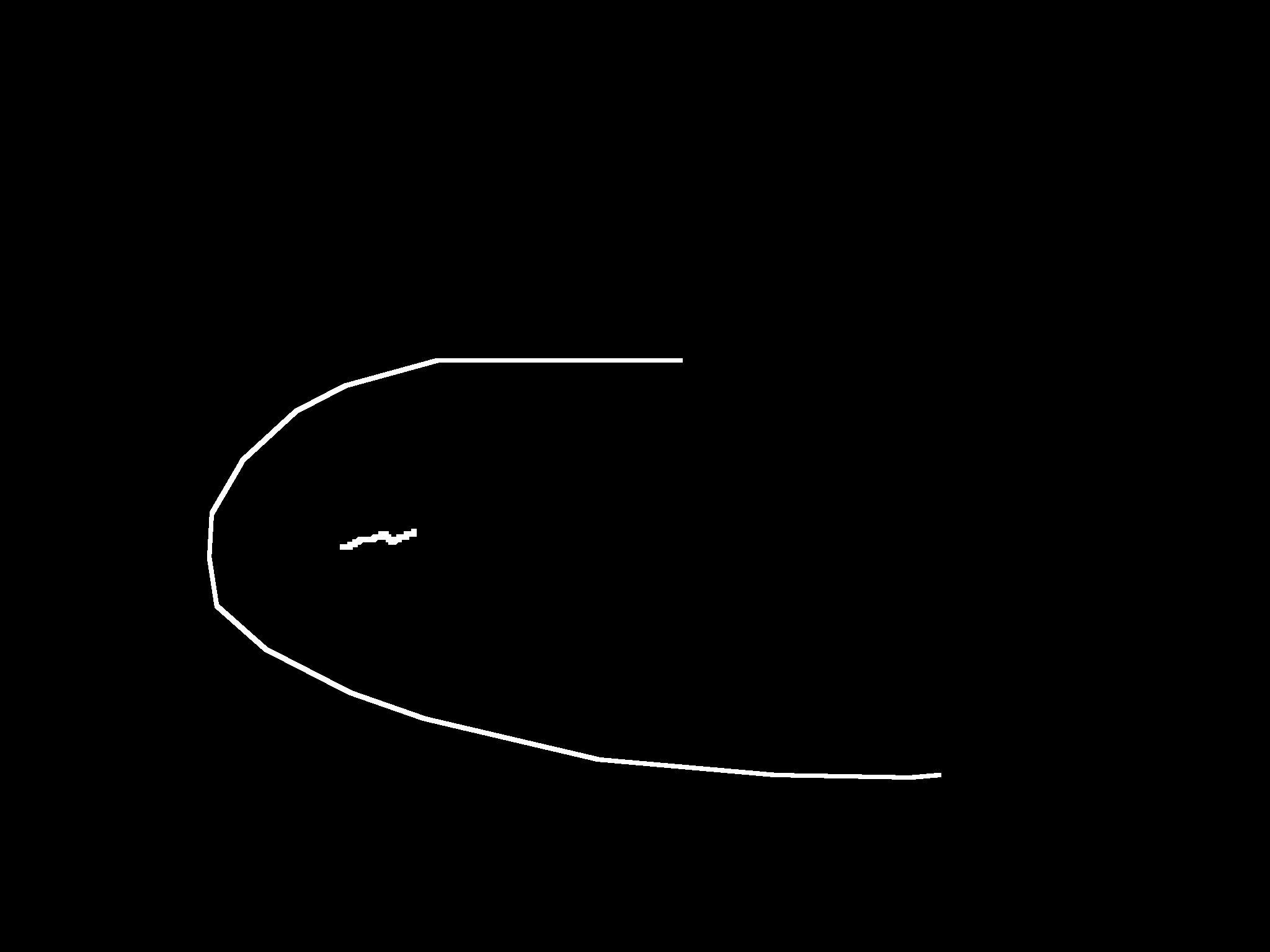}}
   
   \\
    {\includegraphics[width=0.178\linewidth, height=0.119\linewidth]{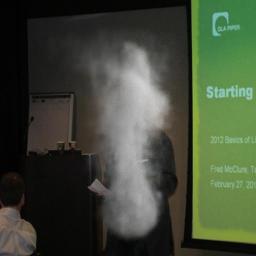}}&
   {\includegraphics[width=0.178\linewidth, height=0.119\linewidth]{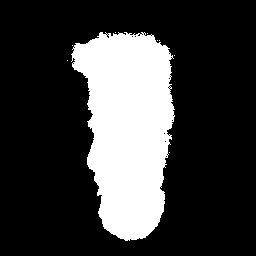}}&
   {\includegraphics[width=0.178\linewidth, height=0.119\linewidth]{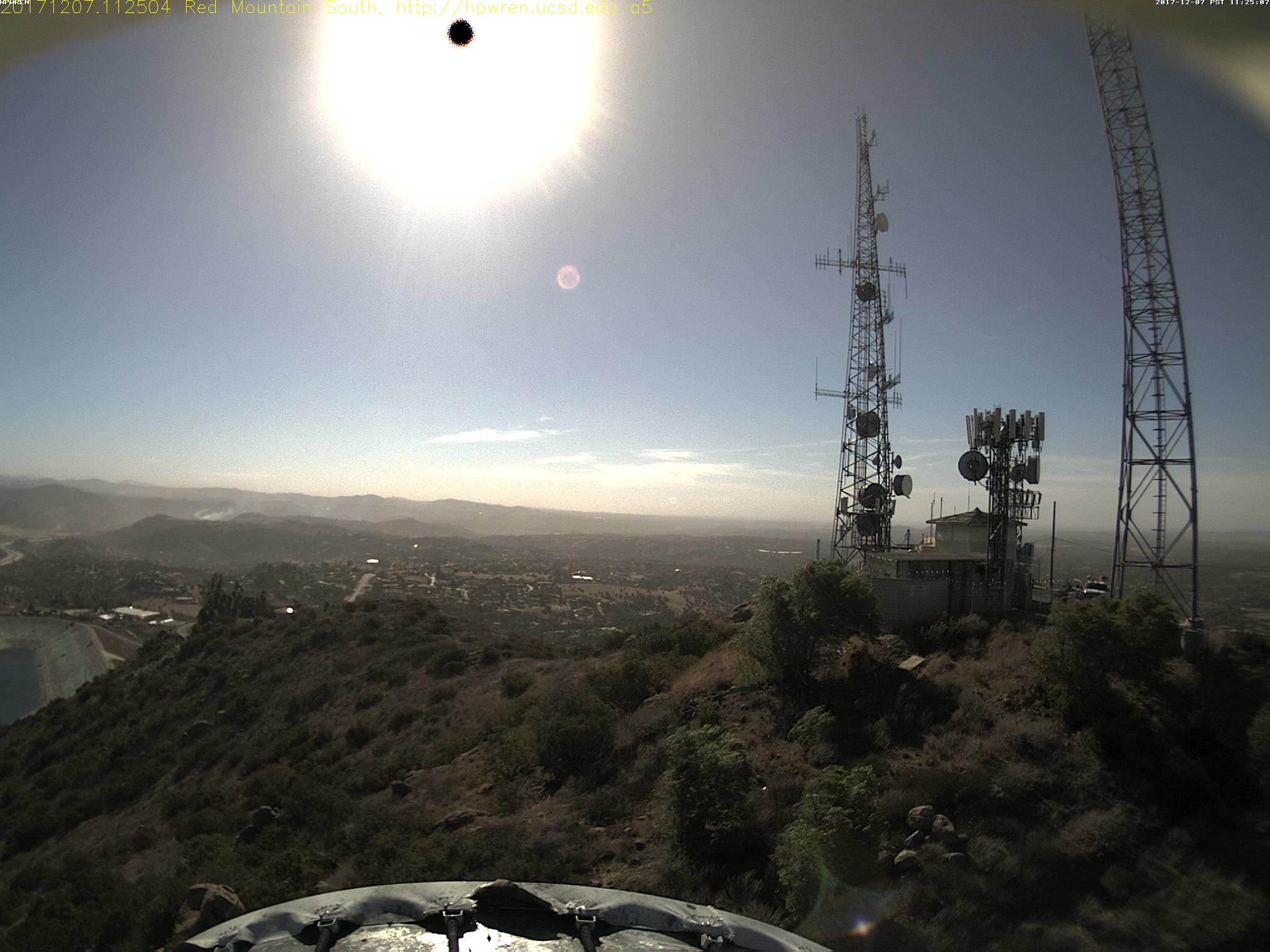}}&
   {\includegraphics[width=0.178\linewidth, height=0.119\linewidth]{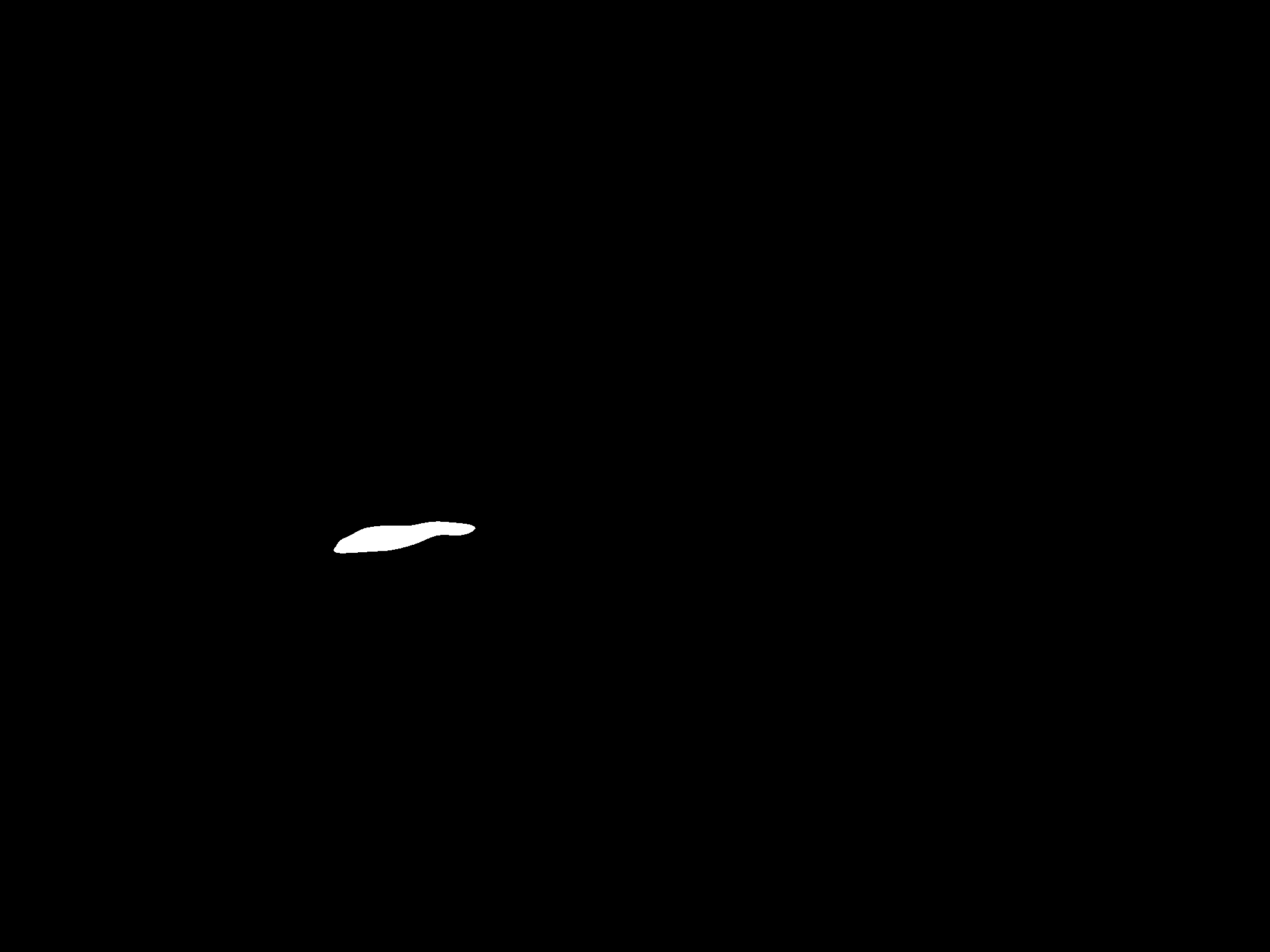}}&
   {\includegraphics[width=0.178\linewidth, height=0.119\linewidth]{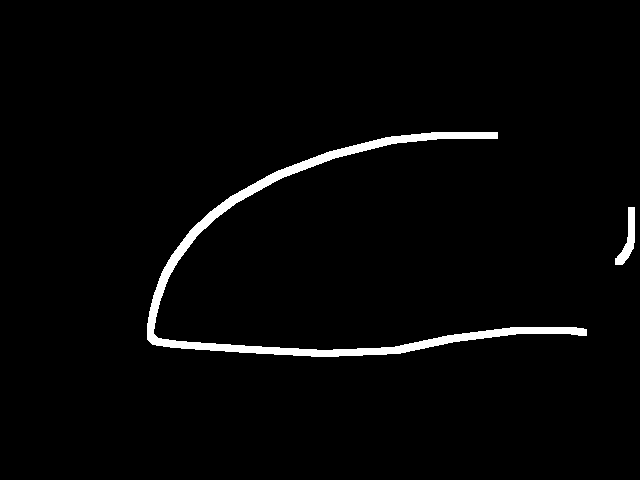}}
  \\
    {\includegraphics[width=0.178\linewidth, height=0.119\linewidth]{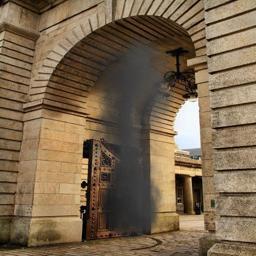}}&
   {\includegraphics[width=0.178\linewidth, height=0.119\linewidth]{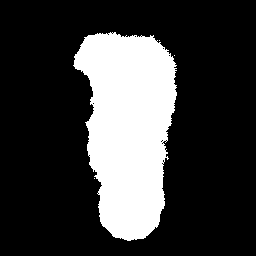}}&
   {\includegraphics[width=0.178\linewidth, height=0.119\linewidth]{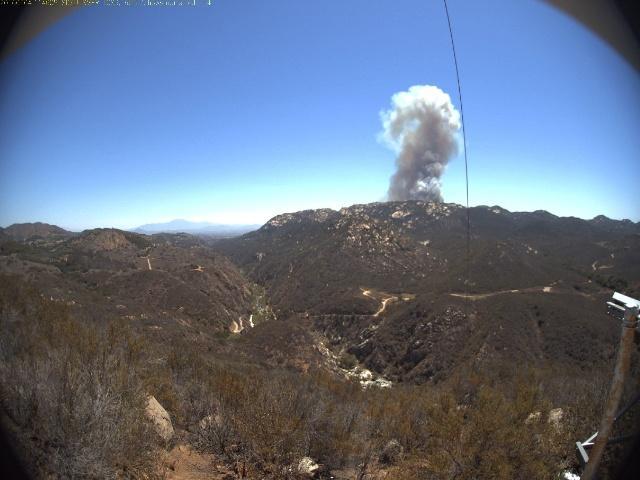}}&
   {\includegraphics[width=0.178\linewidth, height=0.119\linewidth]{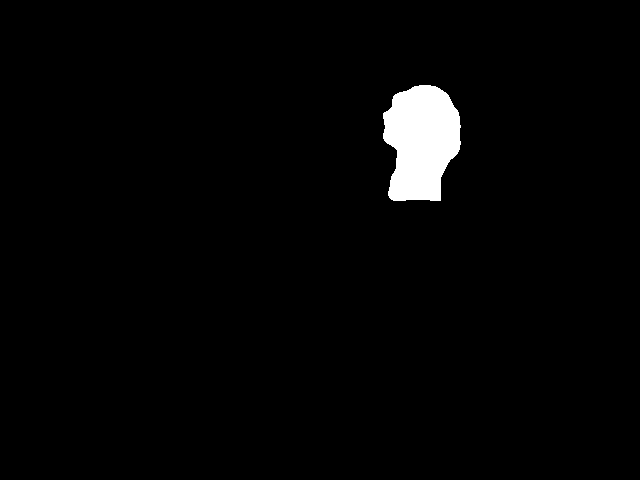}}&
   {\includegraphics[width=0.178\linewidth, height=0.119\linewidth]{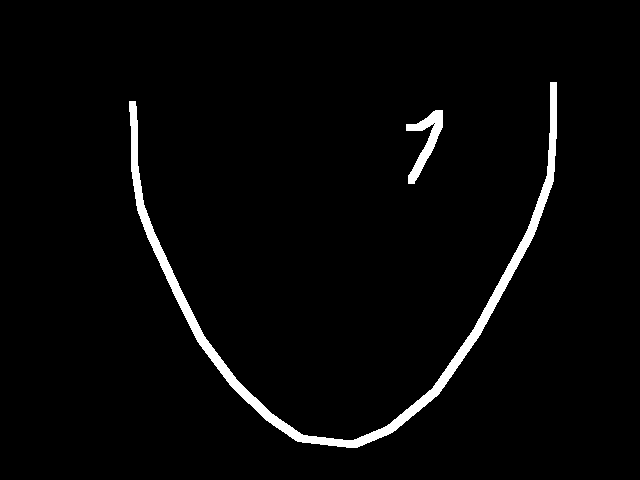}}
    \\
       {\includegraphics[width=0.178\linewidth, height=0.119\linewidth]{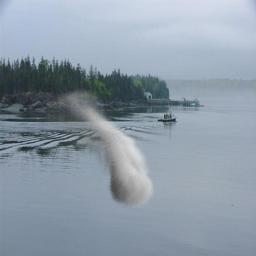}}&
   {\includegraphics[width=0.178\linewidth, height=0.119\linewidth]{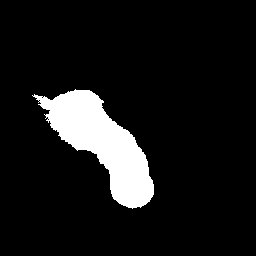}}&
   {\includegraphics[width=0.178\linewidth, height=0.119\linewidth]{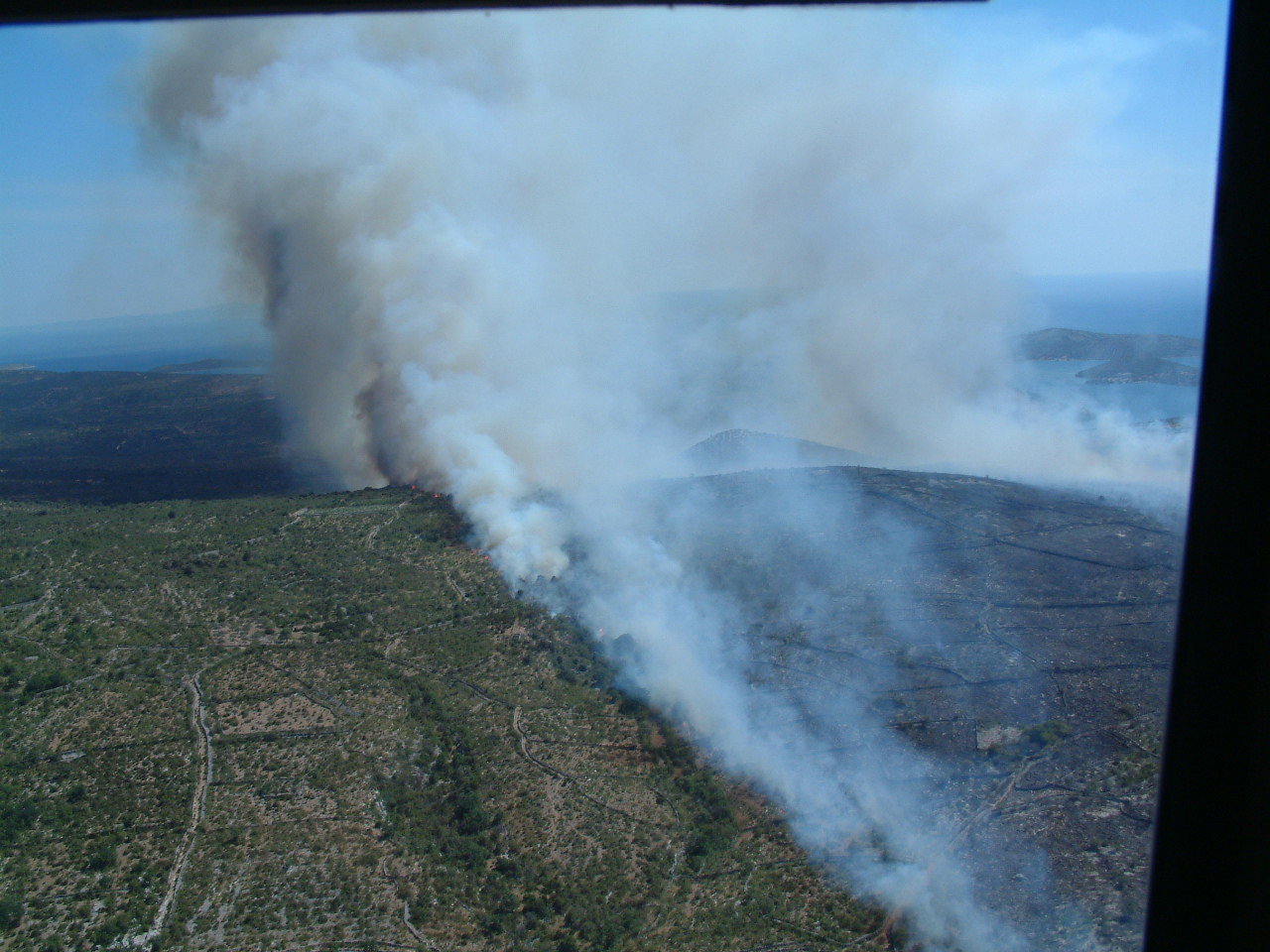}}&
   {\includegraphics[width=0.178\linewidth, height=0.119\linewidth]{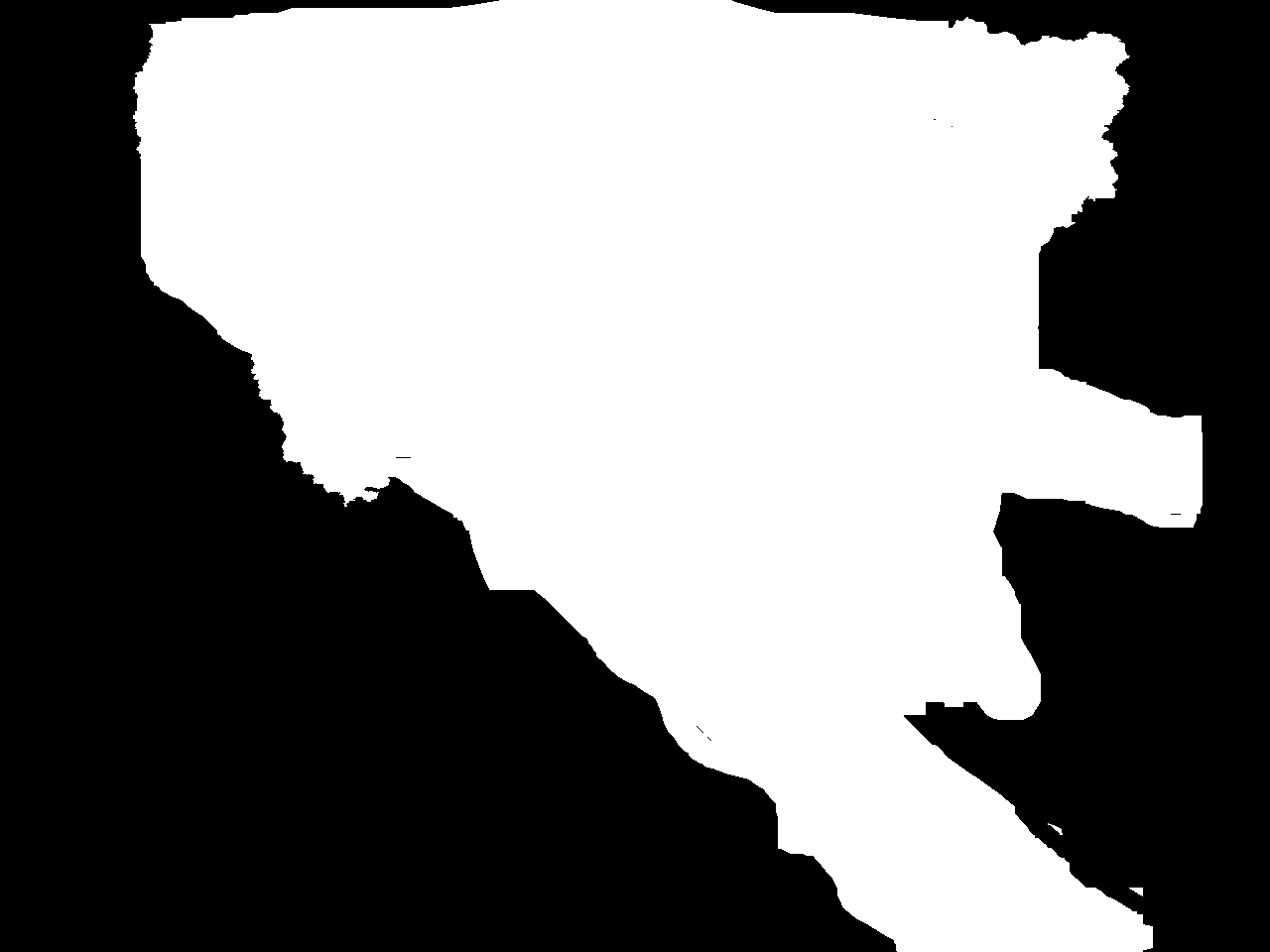}}&
   {\includegraphics[width=0.178\linewidth, height=0.119\linewidth]{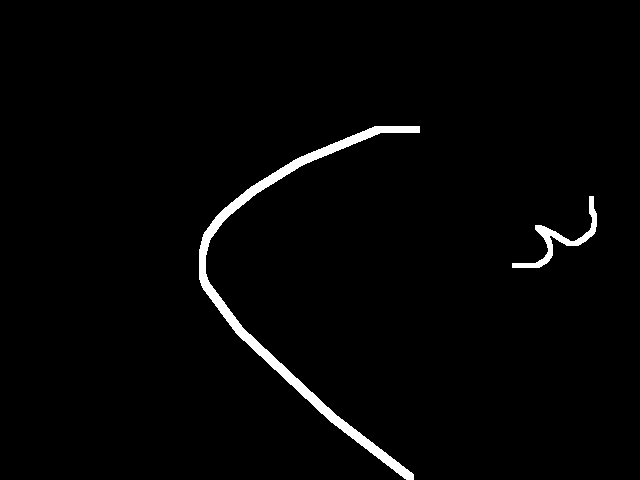}}
   \\
      {\includegraphics[width=0.178\linewidth, height=0.119\linewidth]{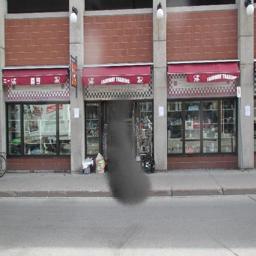}}&
   {\includegraphics[width=0.178\linewidth, height=0.119\linewidth]{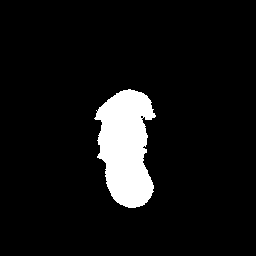}}&
    {\includegraphics[width=0.178\linewidth, height=0.119\linewidth]{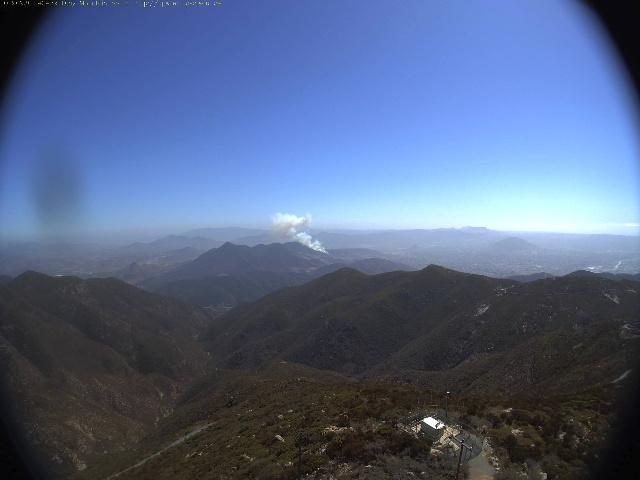}}&
   {\includegraphics[width=0.178\linewidth, height=0.119\linewidth]{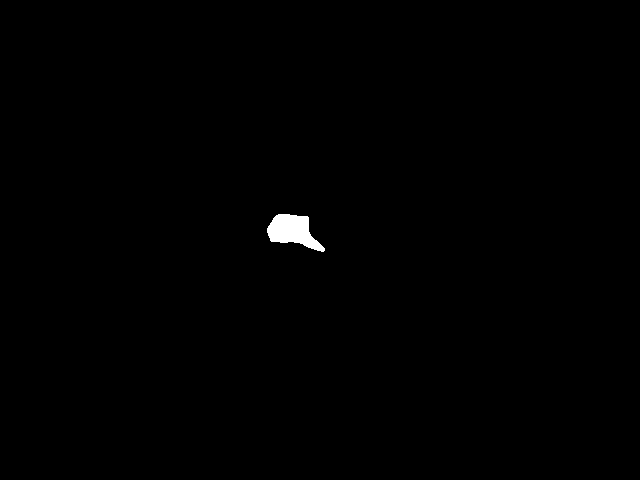}}&
   {\includegraphics[width=0.178\linewidth, height=0.119\linewidth]{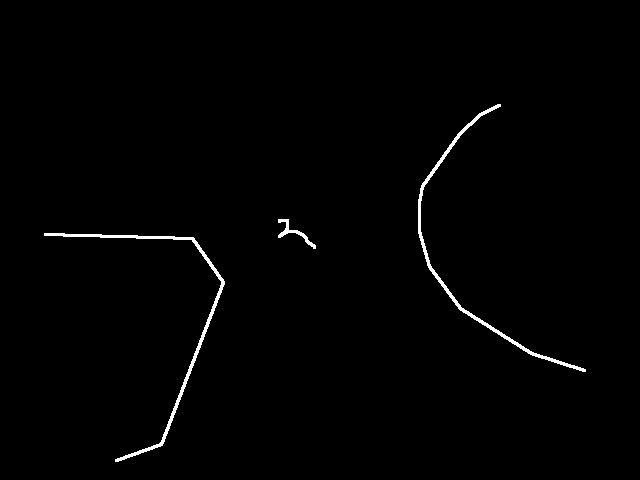}}
   \\
      {\includegraphics[width=0.178\linewidth, height=0.119\linewidth]{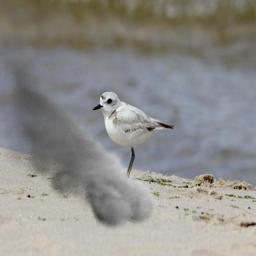}}&
   {\includegraphics[width=0.178\linewidth, height=0.119\linewidth]{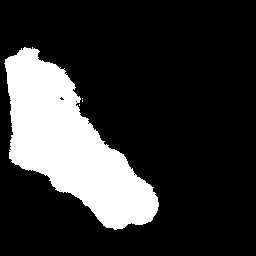}}&
    {\includegraphics[width=0.178\linewidth, height=0.119\linewidth]{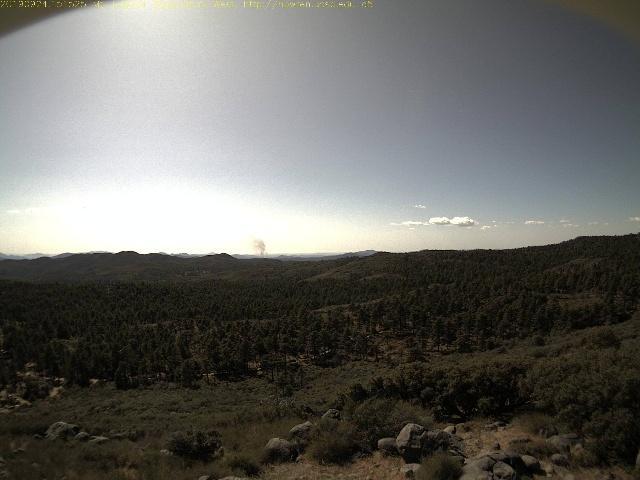}}&
   {\includegraphics[width=0.178\linewidth, height=0.119\linewidth]{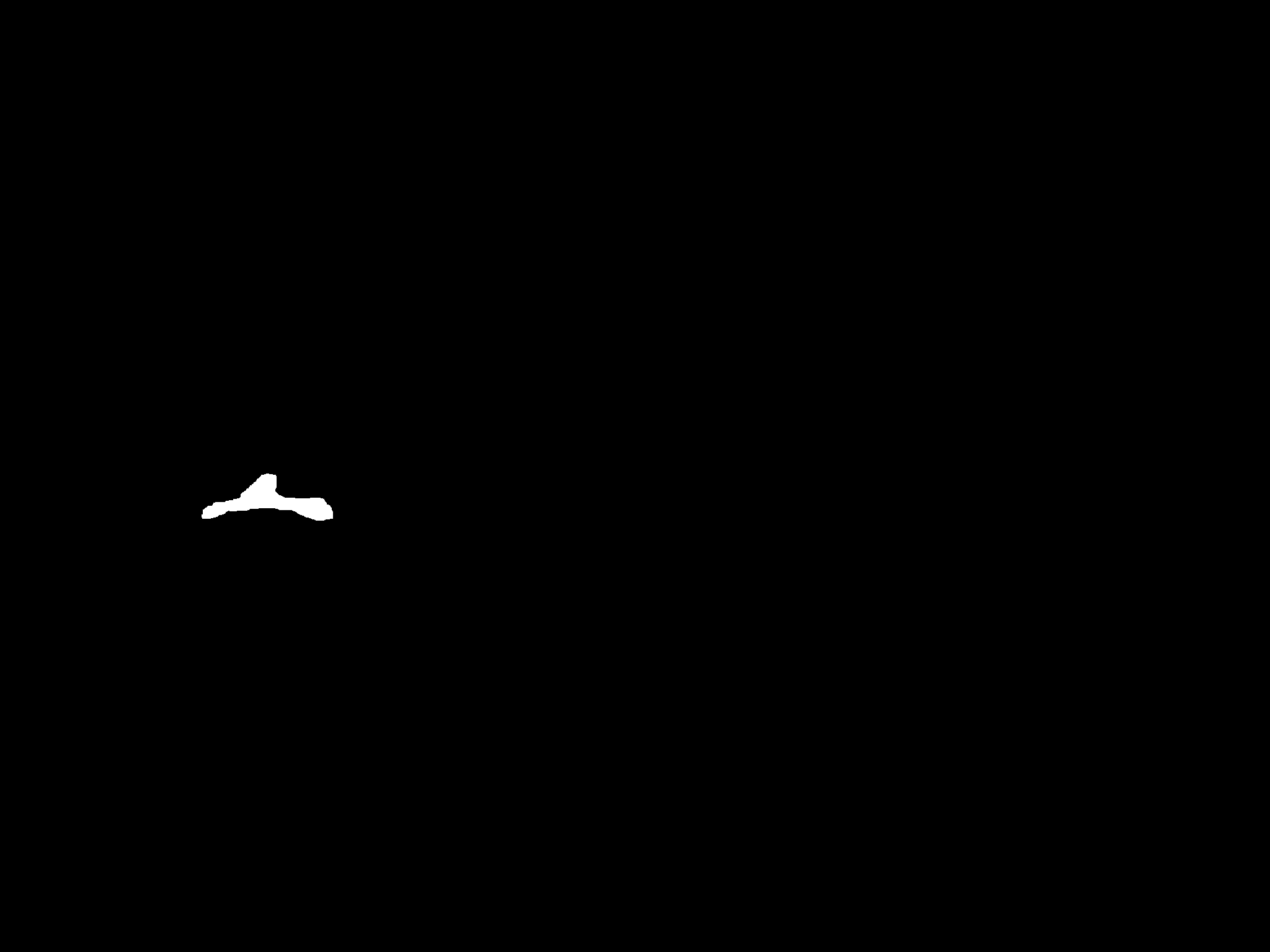}}&
   {\includegraphics[width=0.178\linewidth, height=0.119\linewidth]{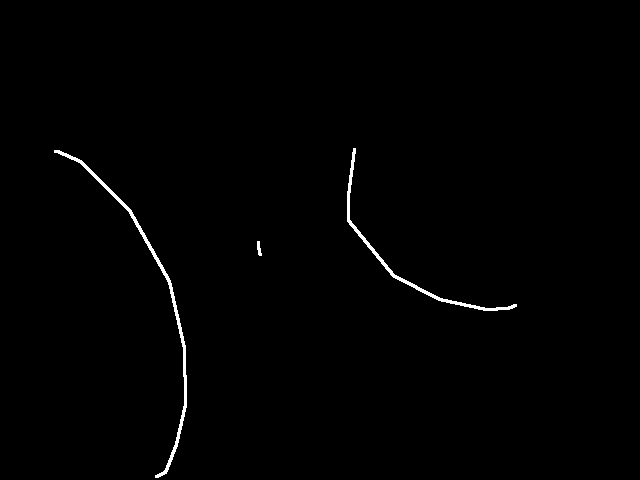}}
   \\

   \footnotesize{a} & \footnotesize{b} & \footnotesize{c} & \footnotesize{d} & \footnotesize{e} \\
   \end{tabular}
      \caption{Comparison between SYN70K and real images in our SMOKE5K dataset. From left to right: (a) the image in SYN70K. (b) the ground truth in SMOKE5K. (c) the image in SMOKE5K. (d) the ground truth in SMOKE5K. (e) the scribble annotation in SMOKE5K.}
      \vspace{-2mm}
      \label{fig1}
   \end{center}
%   \setlength{\abovecaptionskip}{0.cm}
% \setlength{\belowcaptionskip}{-0.cm}
%   \vspace{2mm}
\end{figure*}

In this section, we introduce more detail about our
% mainly describe the real images of our
SMOKE5K dataset.
% and its some special attributes. 
Different from the current largest synthetic dataset SYN70K, the real data in SMOKE5K is specifically for wildfire smoke captured from long-distance tower cameras, which is a cheap and effective option for monitoring wildfire. We mainly recognize six challenging object attributes of our dataset that are different from SYN70K and other conventional dense prediction tasks.
\begin{itemize}
    \item \textbf{Similar background}: The smoke object has a similar appearance with its background due to its translucent appearance, causing large ambiguity in both segmentation and labeling.
    \item \textbf{Similar foreground}: The outdoor images often contain similar confusing objects (false-positive) like haze, cloud, fog, and sun reflection, which makes it  difficult to identify smoke
    % . This makes 
    % smoke very difficult to distinguish 
    even by trained wildfire detection experts.
    \item  \textbf{Object occlusion}: The object can be partially occluded, resulting in disconnected parts.
    \item \textbf{Diverse shapes}: Smoke has diverse shapes due to its non-rigid shape, increasing the difficulty to perform precise segmentation.
    \item \textbf{Small objects}: The ratio between the smoke and the whole image can be even smaller than 1\% as we want to detect the smoke in the early phase.
    \item \textbf{Diverse Locations}: Most images in SYN70K (see Fig.\ref{fig1} (a)) dataset
    % and most images in some conventional binary segmentation tasks \ie~saliency object detection, camouflage object detection, usually 
    have strong center bias, where the smoke usually appears in the center of the image.
    % which means the location of most objects is in the center of the image. 
    Different from them, the smoke in our dataset can be in any position in the image.

\end{itemize}

Also, smoke images are difficult to precisely label due to the challenging attributes mentioned above, thus we also provide scribble annotations for weakly supervised smoke segmentation, shown in Fig.\ref{fig1} (e). Weakly supervised learning allows a model to learn from a weak supervision signal \ie~image-level \citep{weak_im1,weak_im2}, scribble-level \cite{scribble}, bounding box-level \citep{box1,box2} annotations. Scribble annotation is especially suitable for smoke segmentation as: 1) compared with pixel-wise annotation, scribble annotation is much cheaper and faster (It only takes 2$\sim$5 seconds for each image); 2) compared with image-level annotation, scribble annotation is effective for localizing small objects; 3) compared with bounding box-level annotation, scribble annotation is more flexible for diverse shape objects. 
\subsection{Dataset Visualisation}
The qualitative comparisons between SYN70K and our real images are visualized in Fig.\ref{fig1},  which further demonstrates the superior of our dataset.

\section{Qualitative Results}

\begin{figure*}[t]
\setlength{\abovecaptionskip}{0.cm}
   \begin{center}
     \renewcommand{\arraystretch}{0.2}
   \begin{tabular}{ c@{ } c@{ } c@{ } c@{ }  c@{ }  c@{ } c@{ } c@{ }  }

   {\includegraphics[width=0.11\linewidth, height=0.09\linewidth]{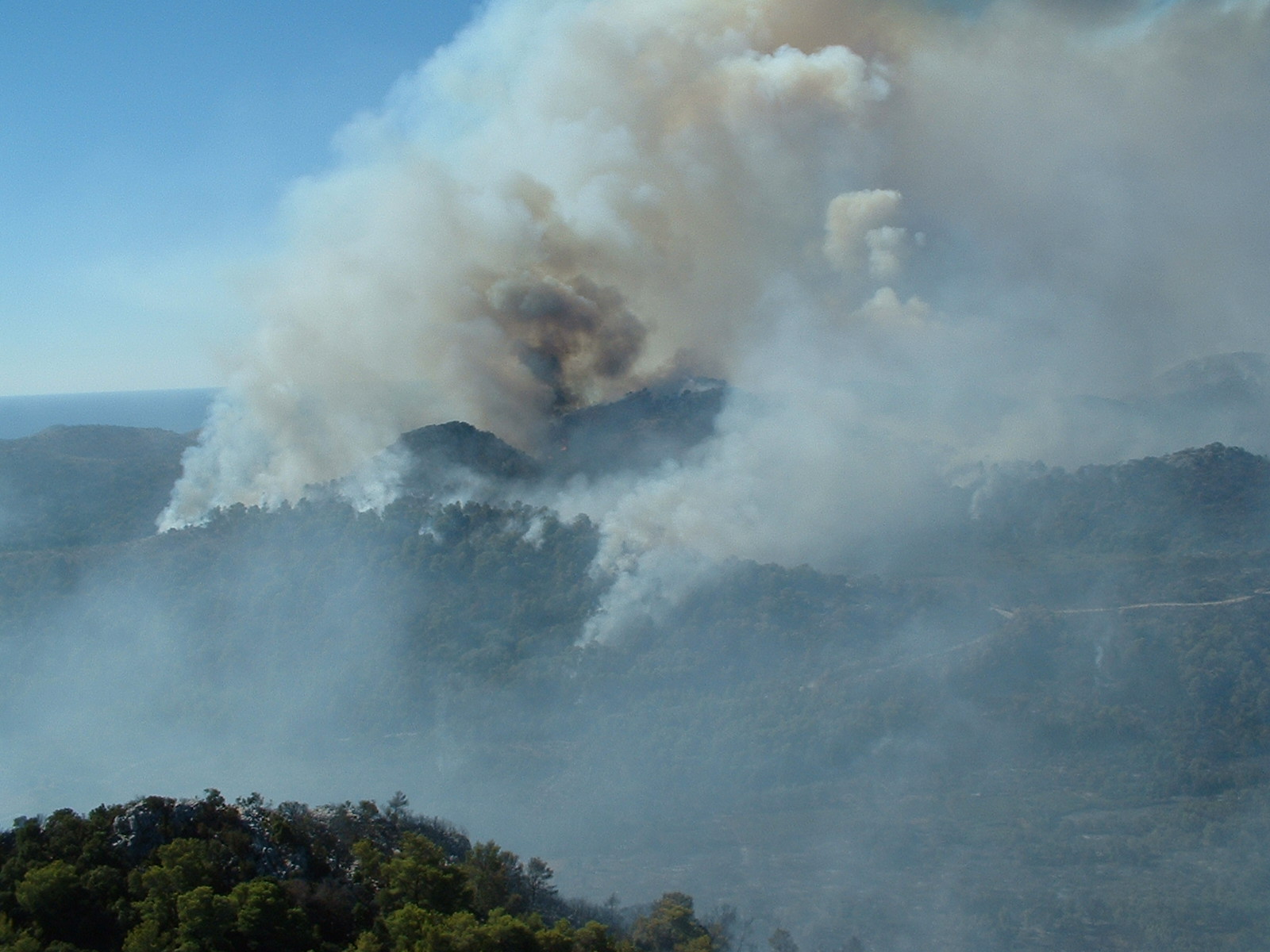}}&
   {\includegraphics[width=0.11\linewidth, height=0.09\linewidth]{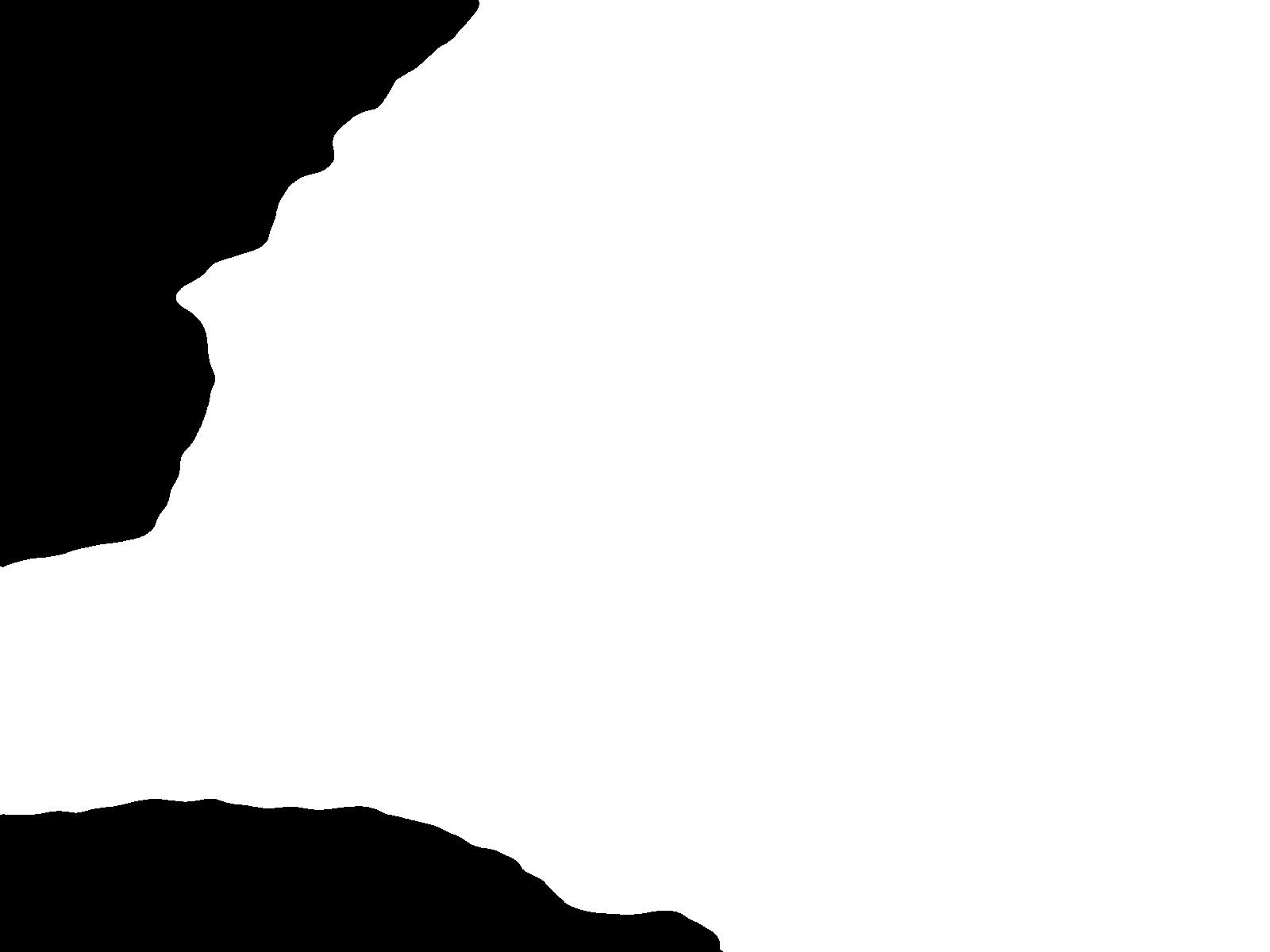}}&
   {\includegraphics[width=0.11\linewidth, height=0.09\linewidth]{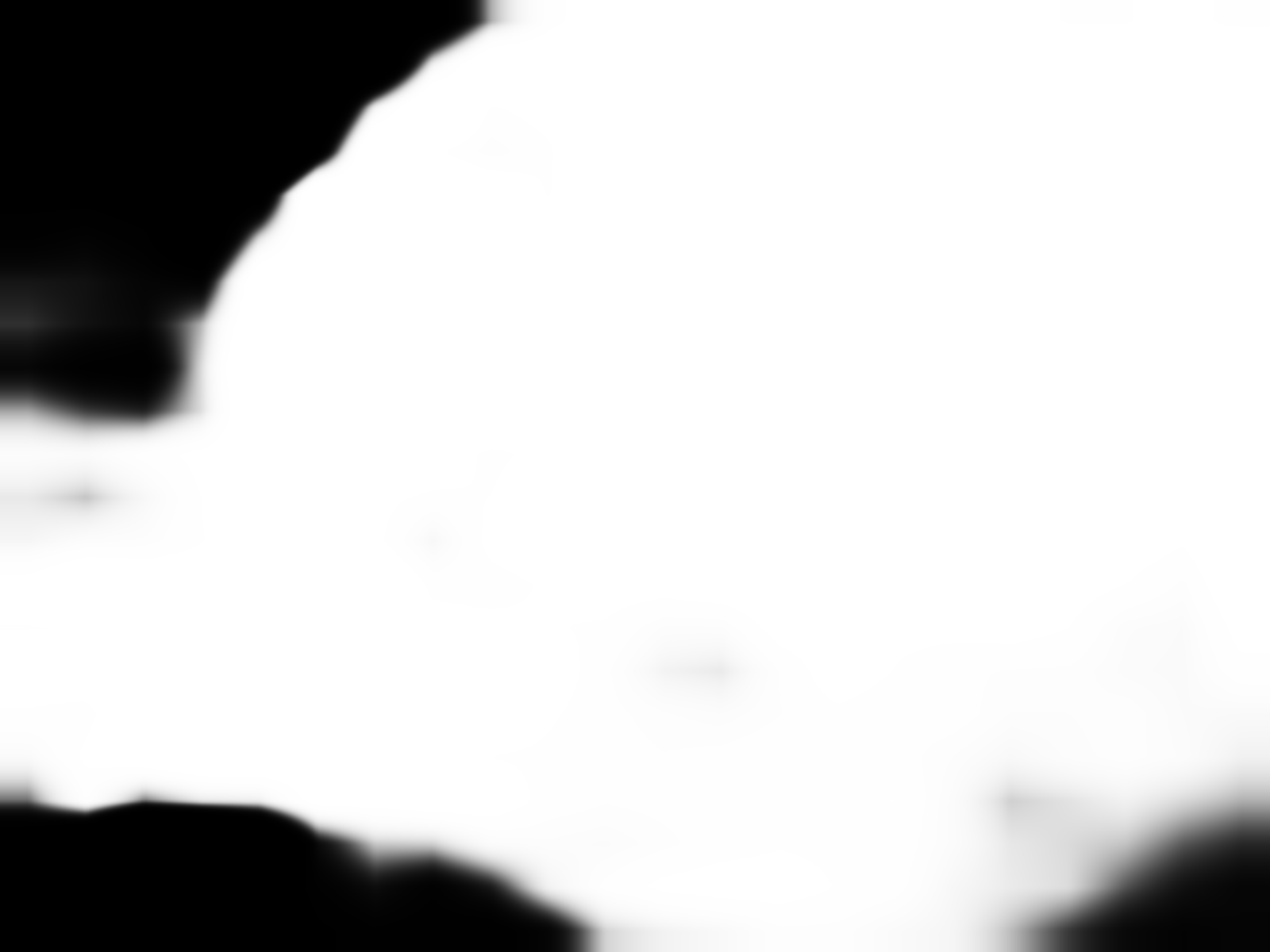}}&
   {\includegraphics[width=0.11\linewidth, height=0.09\linewidth]{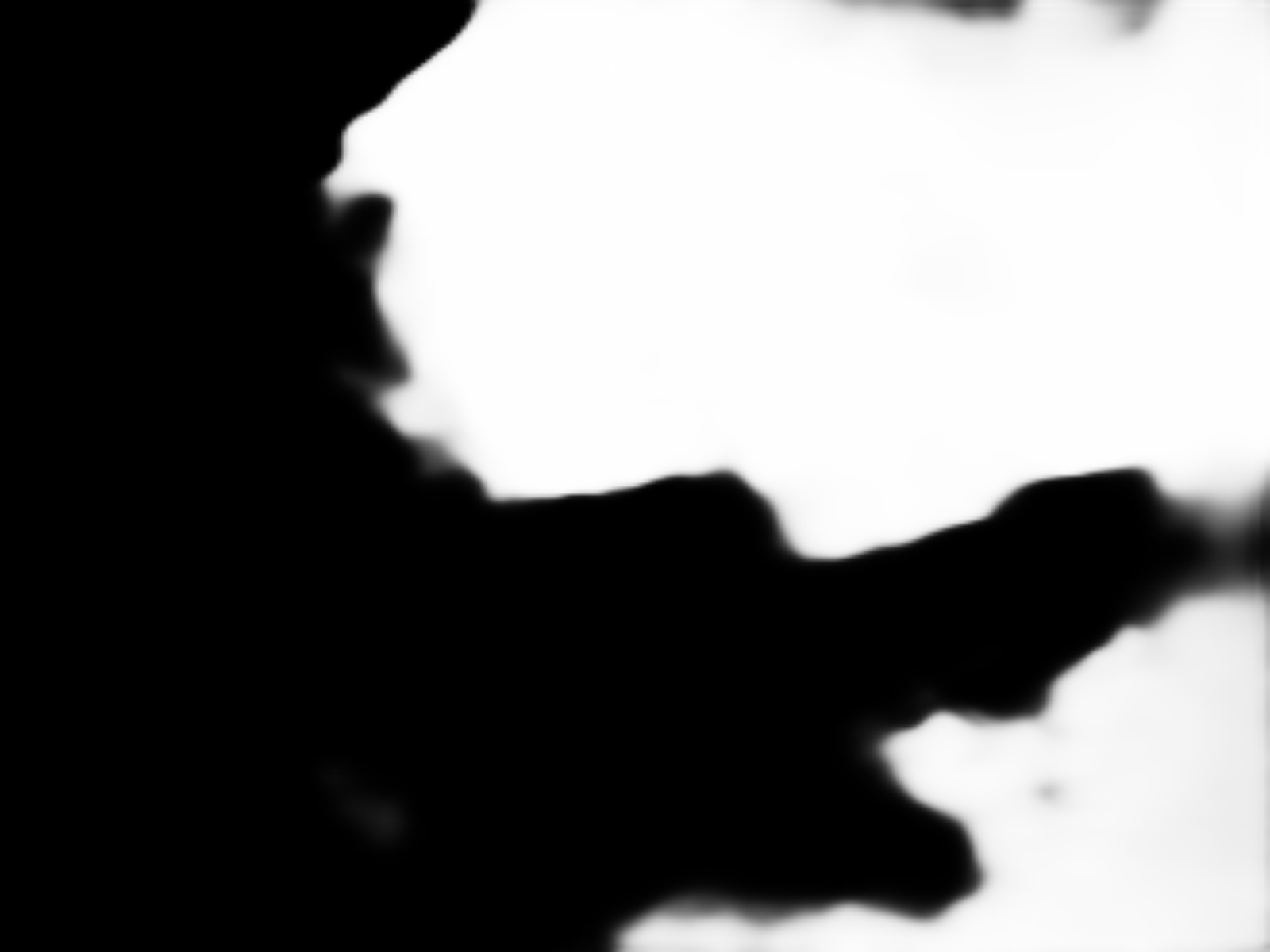}}&
   {\includegraphics[width=0.11\linewidth, height=0.09\linewidth]{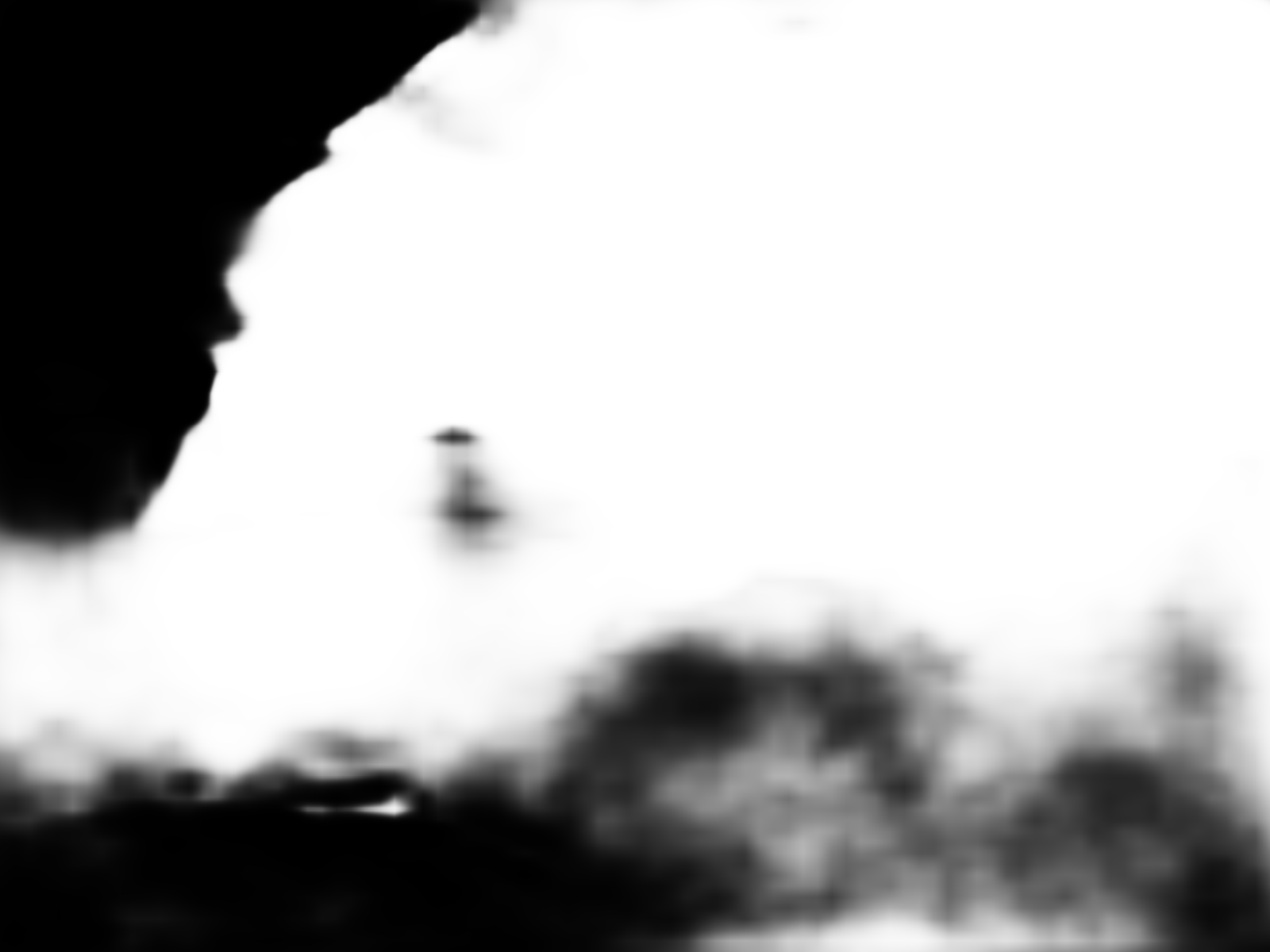}}&
      {\includegraphics[width=0.11\linewidth, height=0.09\linewidth]{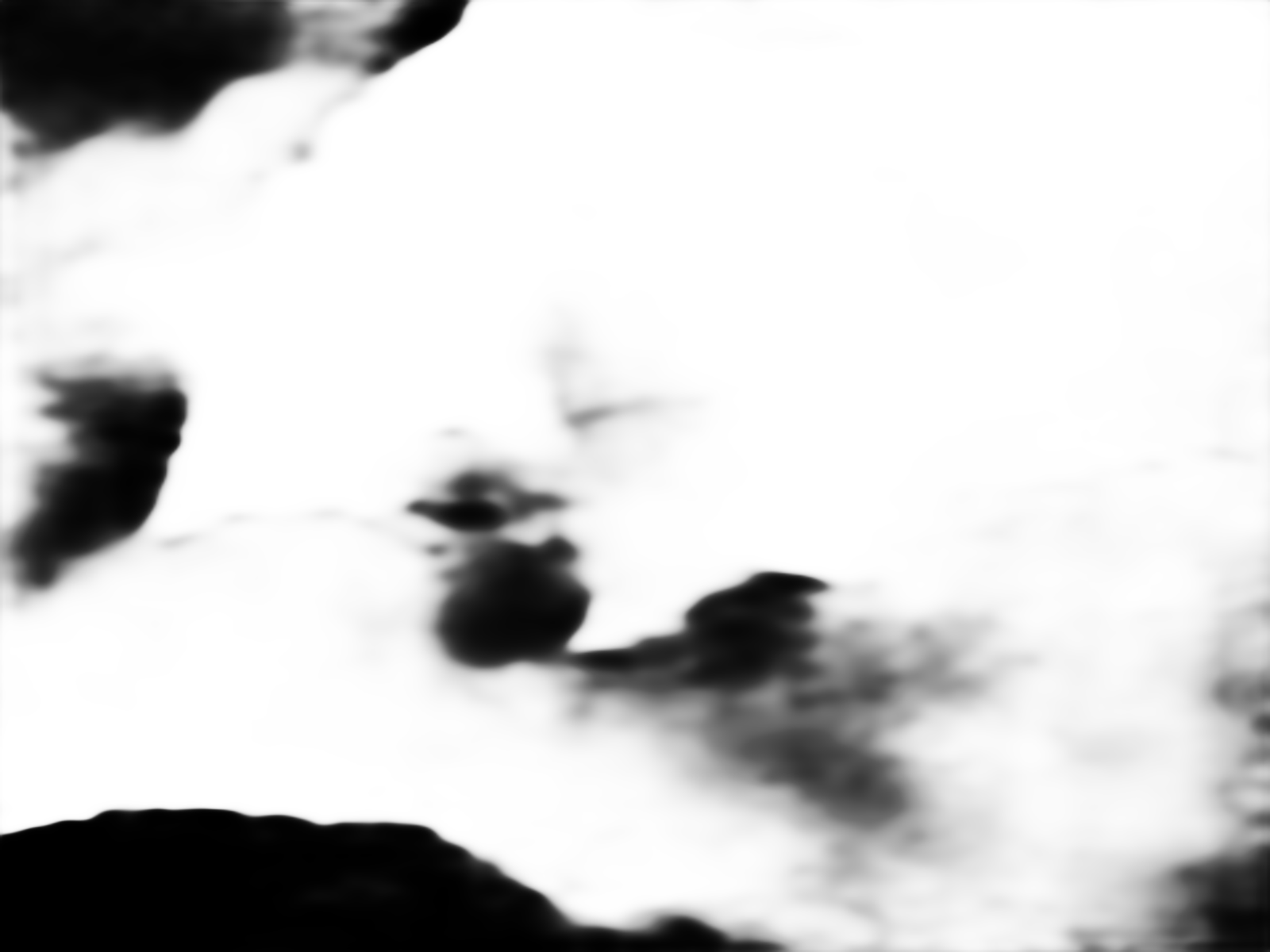}}&
   {\includegraphics[width=0.11\linewidth, height=0.09\linewidth]{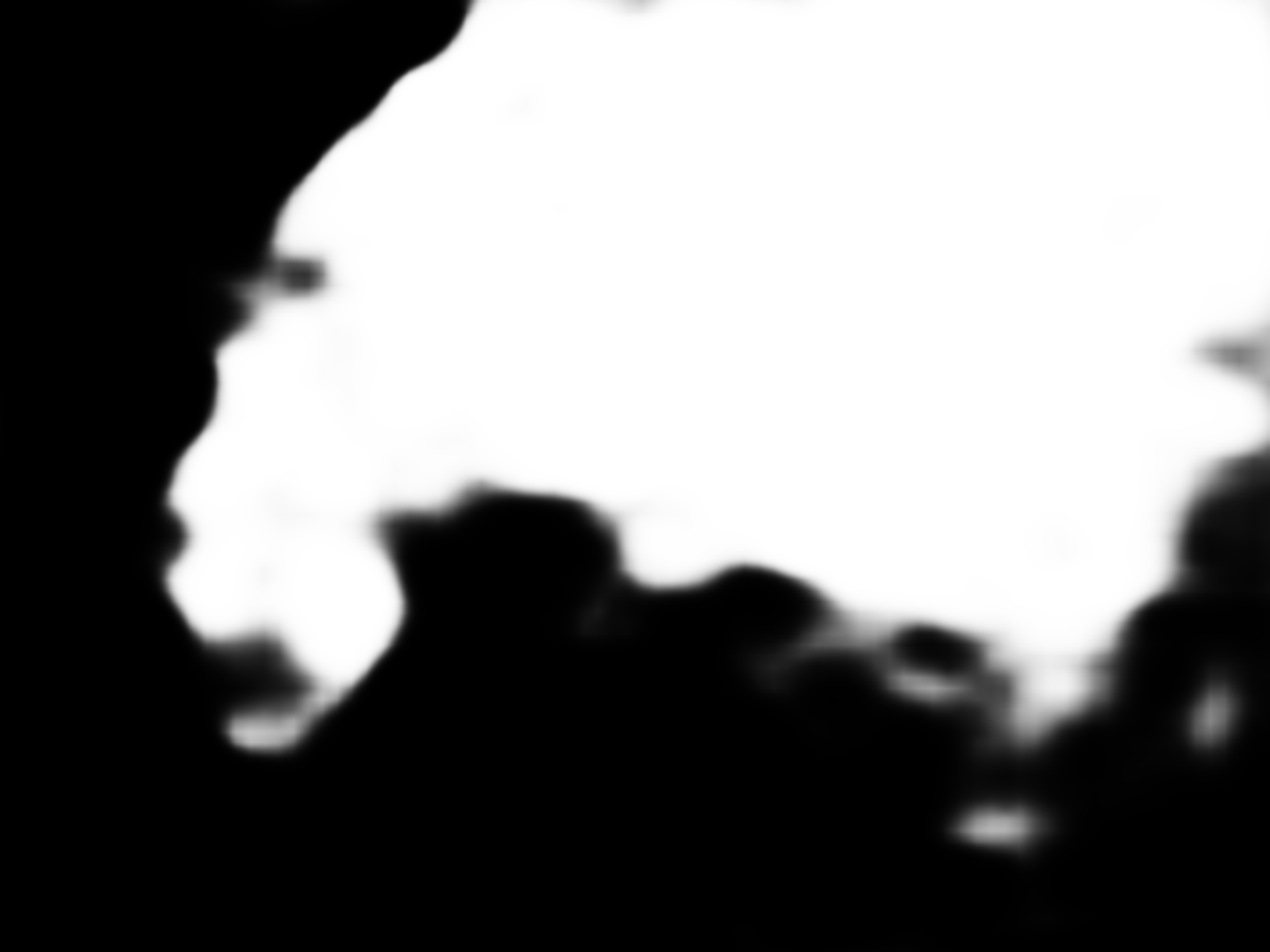}}&
   {\includegraphics[width=0.11\linewidth, height=0.09\linewidth]{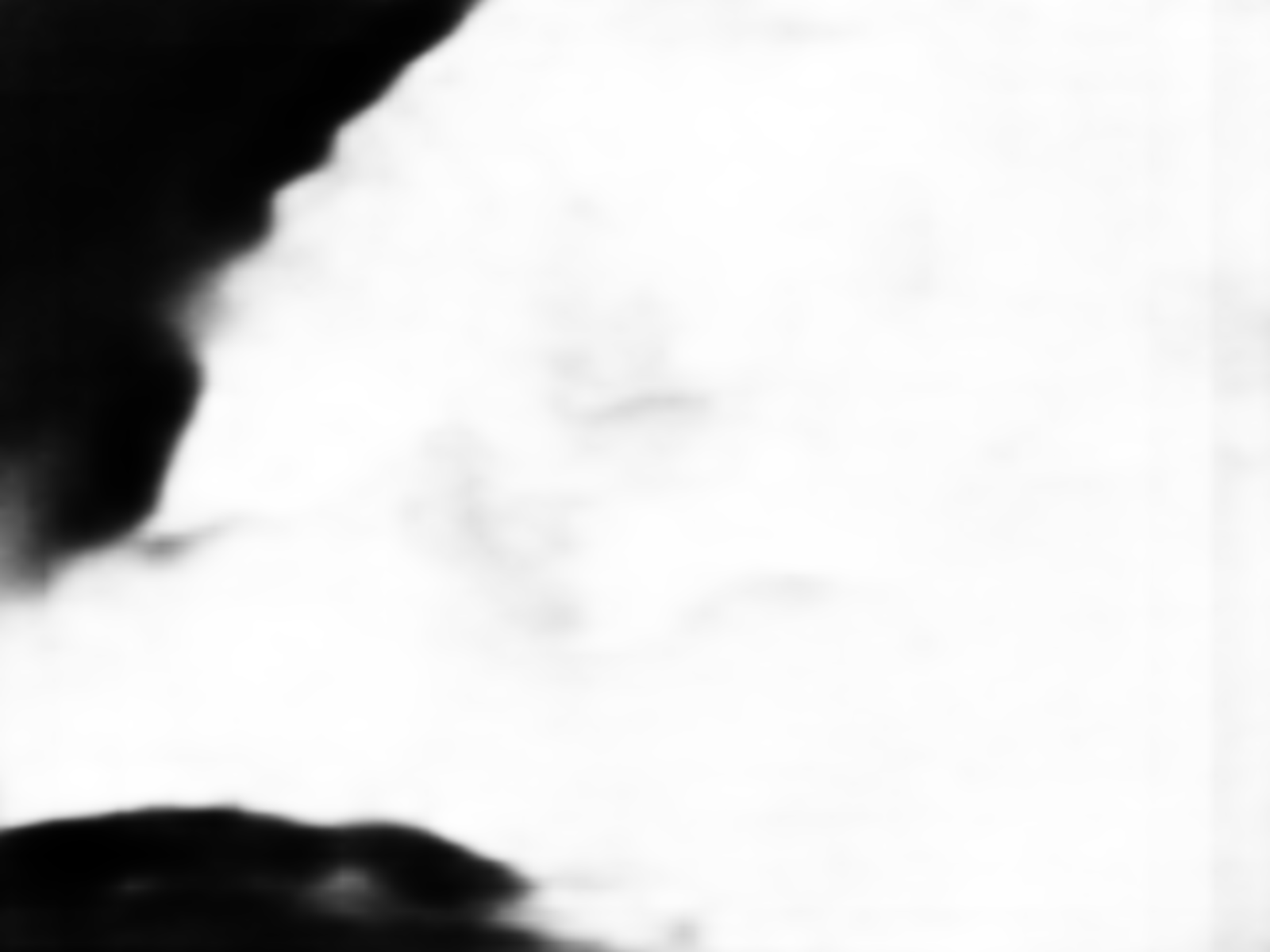}}
   
   \\
 
      {\includegraphics[width=0.11\linewidth, height=0.09\linewidth]{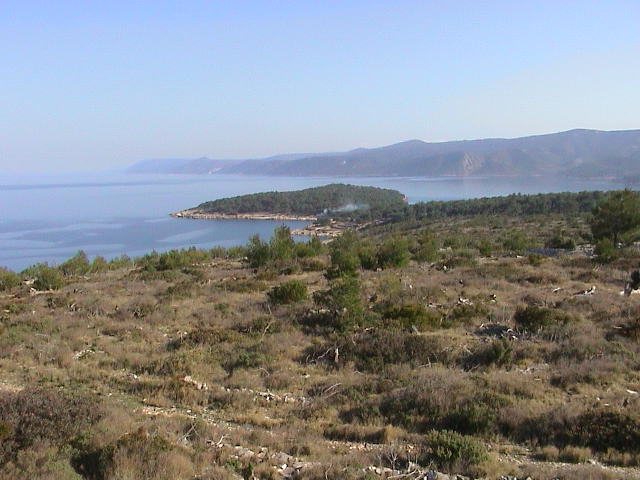}}&
   {\includegraphics[width=0.11\linewidth, height=0.09\linewidth]{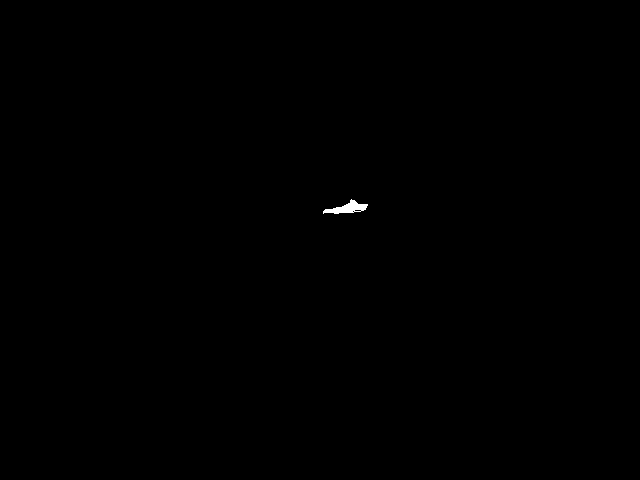}}&
   {\includegraphics[width=0.11\linewidth, height=0.09\linewidth]{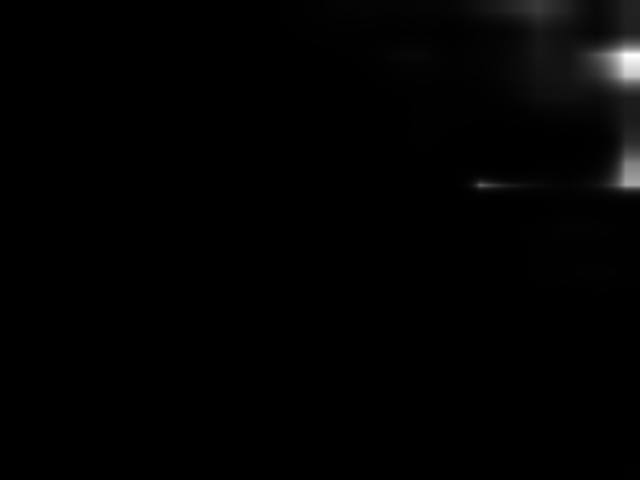}}&
   {\includegraphics[width=0.11\linewidth, height=0.09\linewidth]{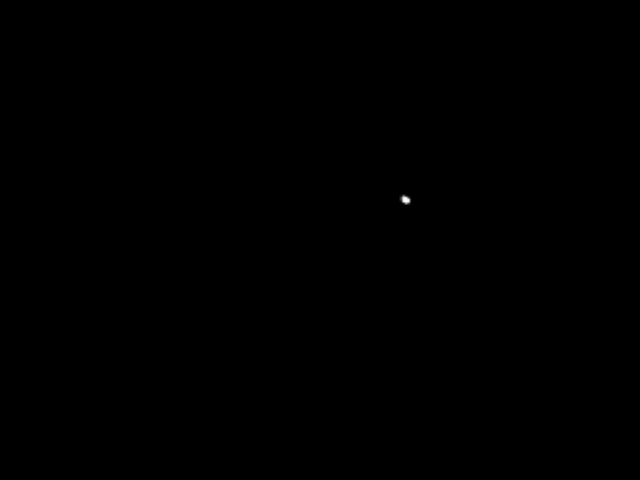}}&
   {\includegraphics[width=0.11\linewidth, height=0.09\linewidth]{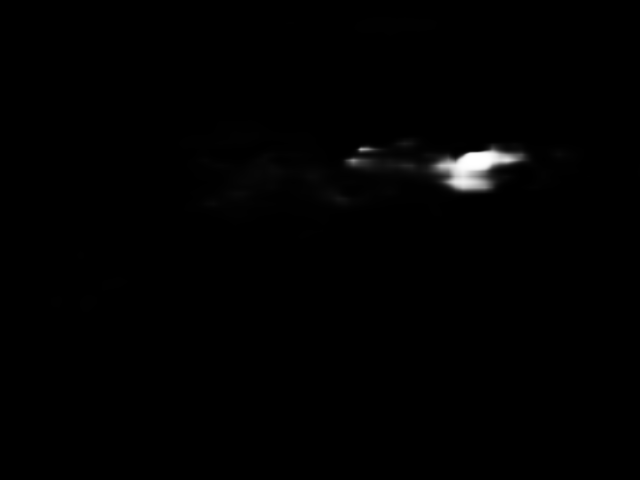}}&
      {\includegraphics[width=0.11\linewidth, height=0.09\linewidth]{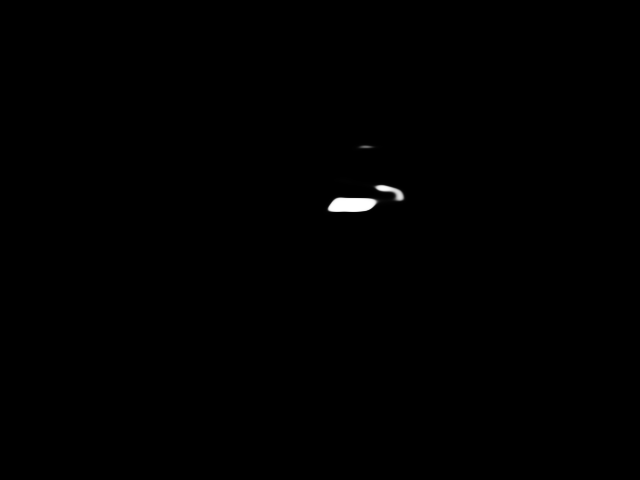}}&
   {\includegraphics[width=0.11\linewidth, height=0.09\linewidth]{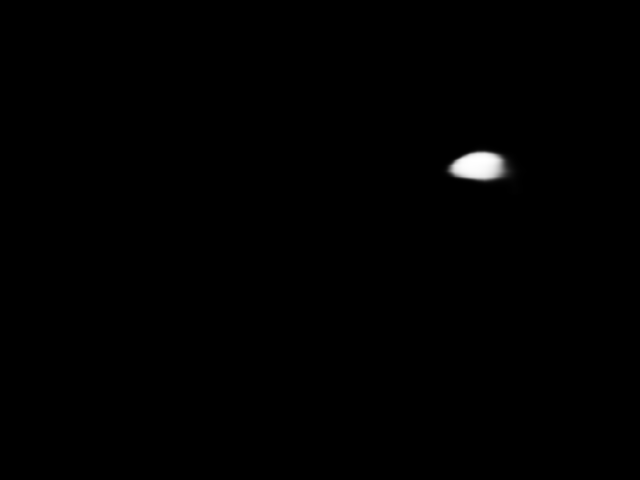}}&
   {\includegraphics[width=0.11\linewidth, height=0.09\linewidth]{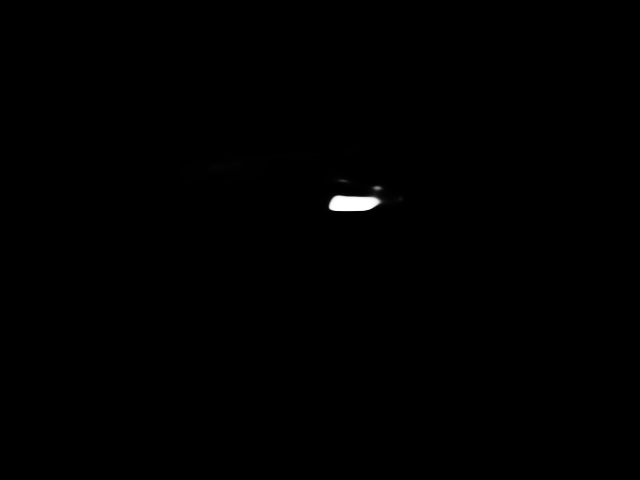}}
   
   \\
      {\includegraphics[width=0.11\linewidth, height=0.09\linewidth]{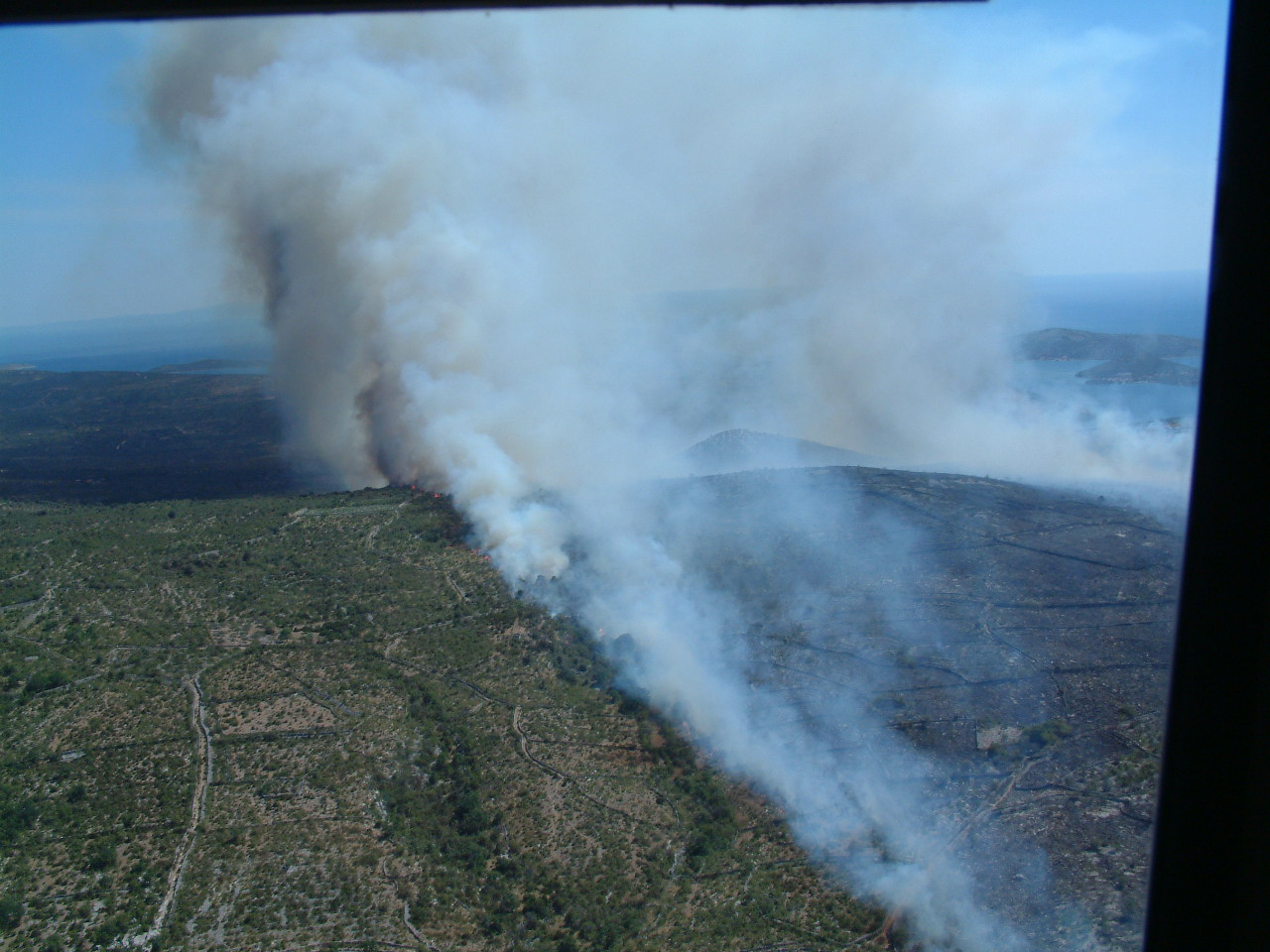}}&
   {\includegraphics[width=0.11\linewidth, height=0.09\linewidth]{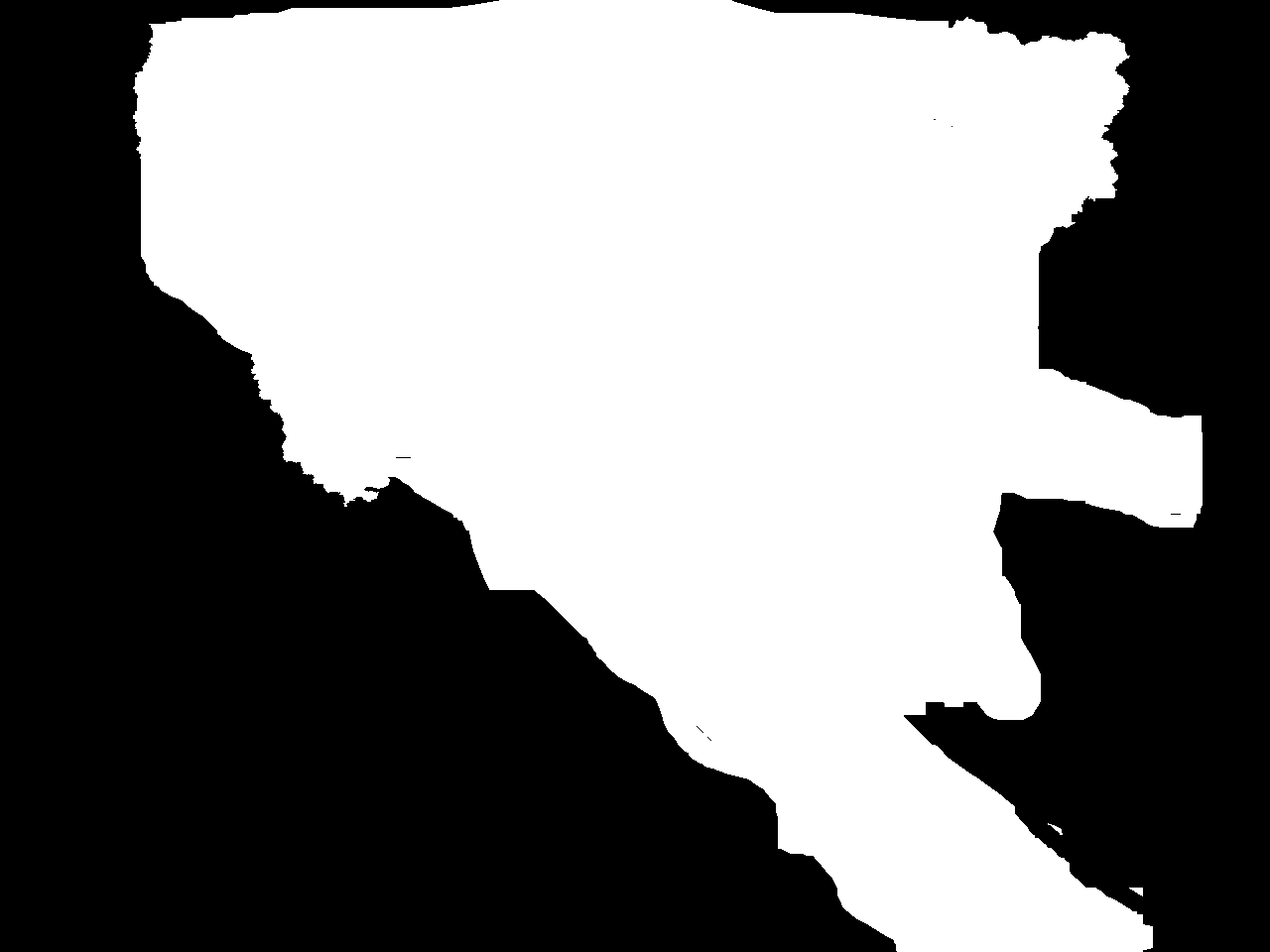}}&
   {\includegraphics[width=0.11\linewidth, height=0.09\linewidth]{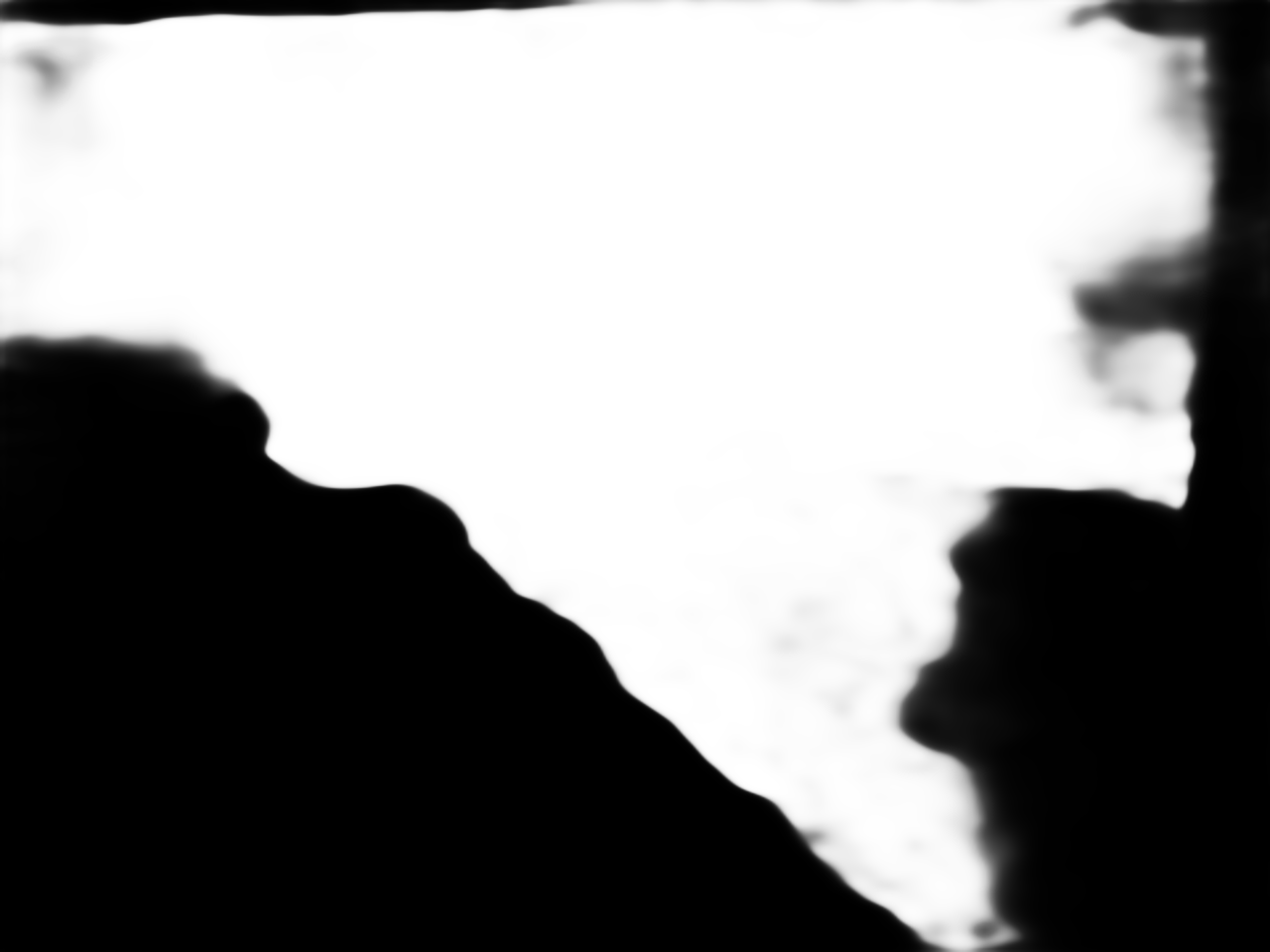}}&
   {\includegraphics[width=0.11\linewidth, height=0.09\linewidth]{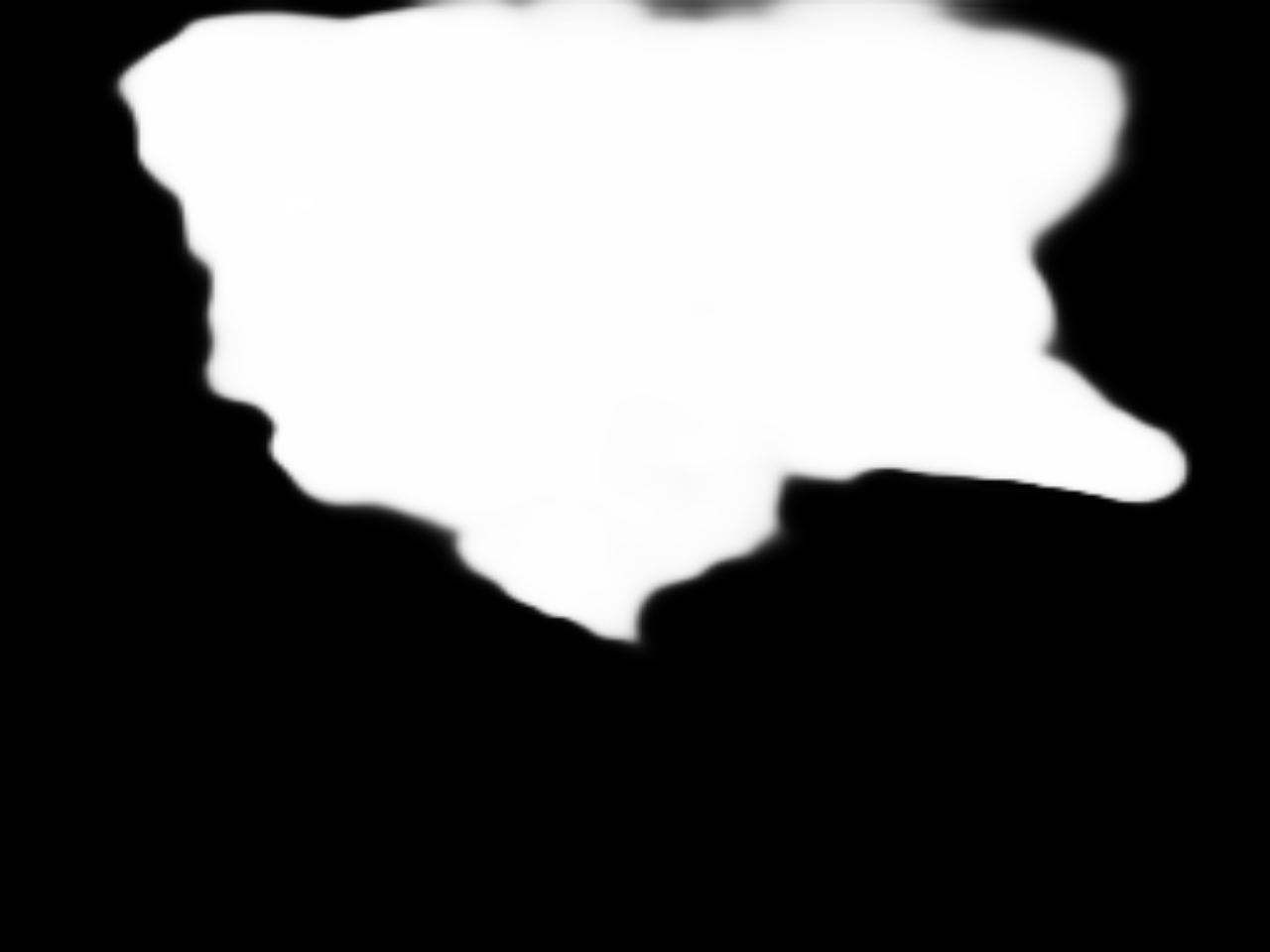}}&
   {\includegraphics[width=0.11\linewidth, height=0.09\linewidth]{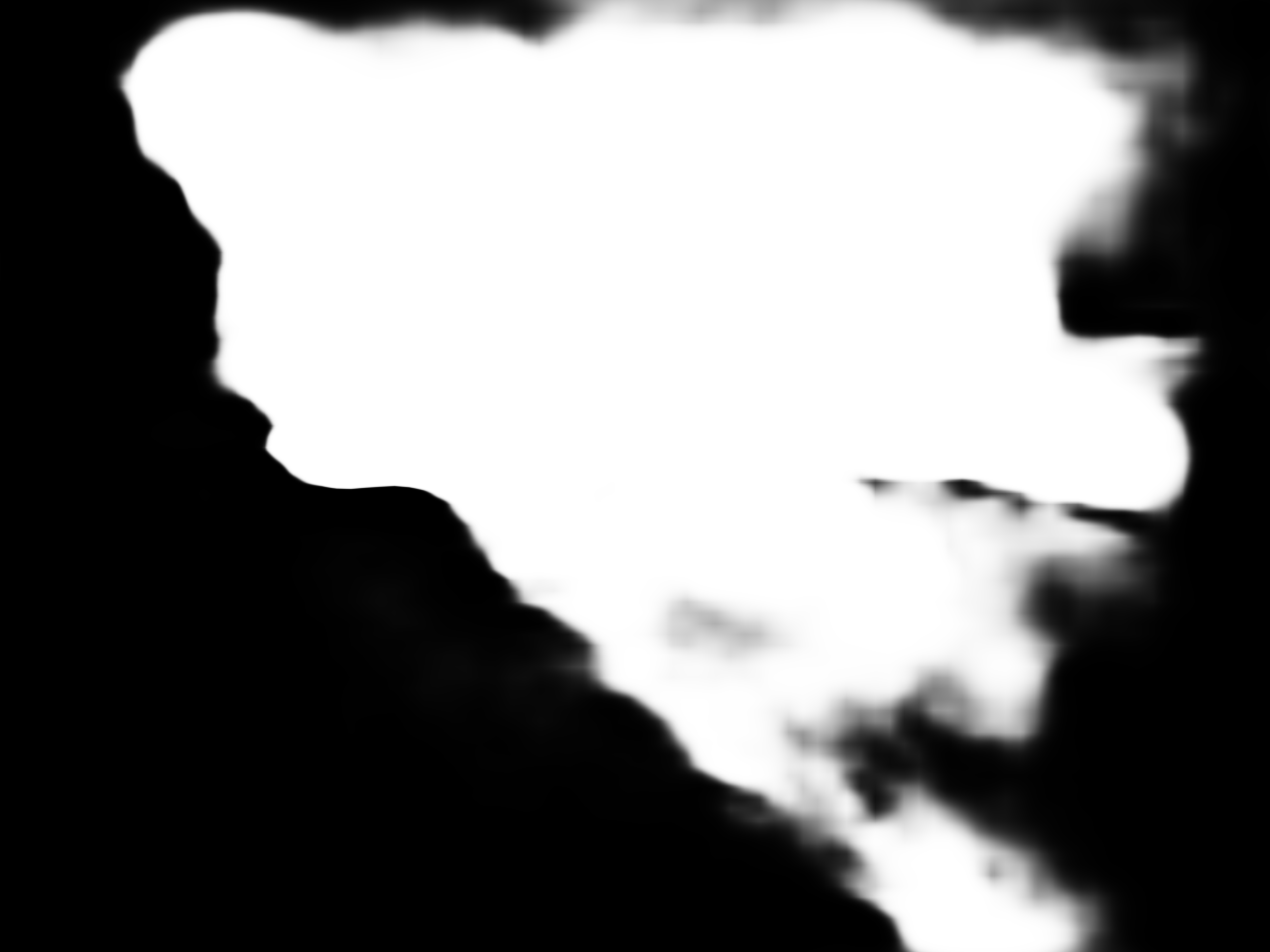}}&
      {\includegraphics[width=0.11\linewidth, height=0.09\linewidth]{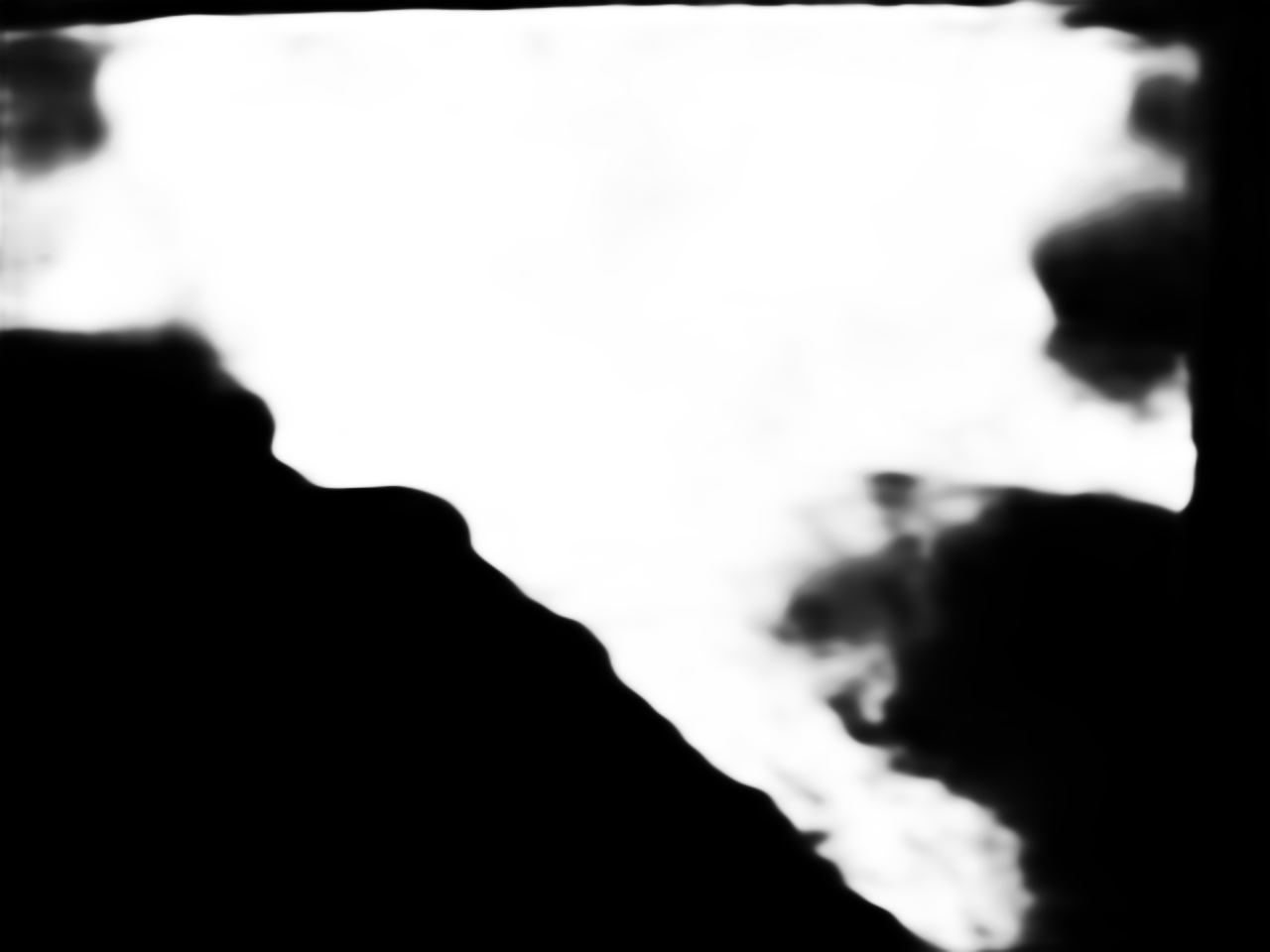}}&
   {\includegraphics[width=0.11\linewidth, height=0.09\linewidth]{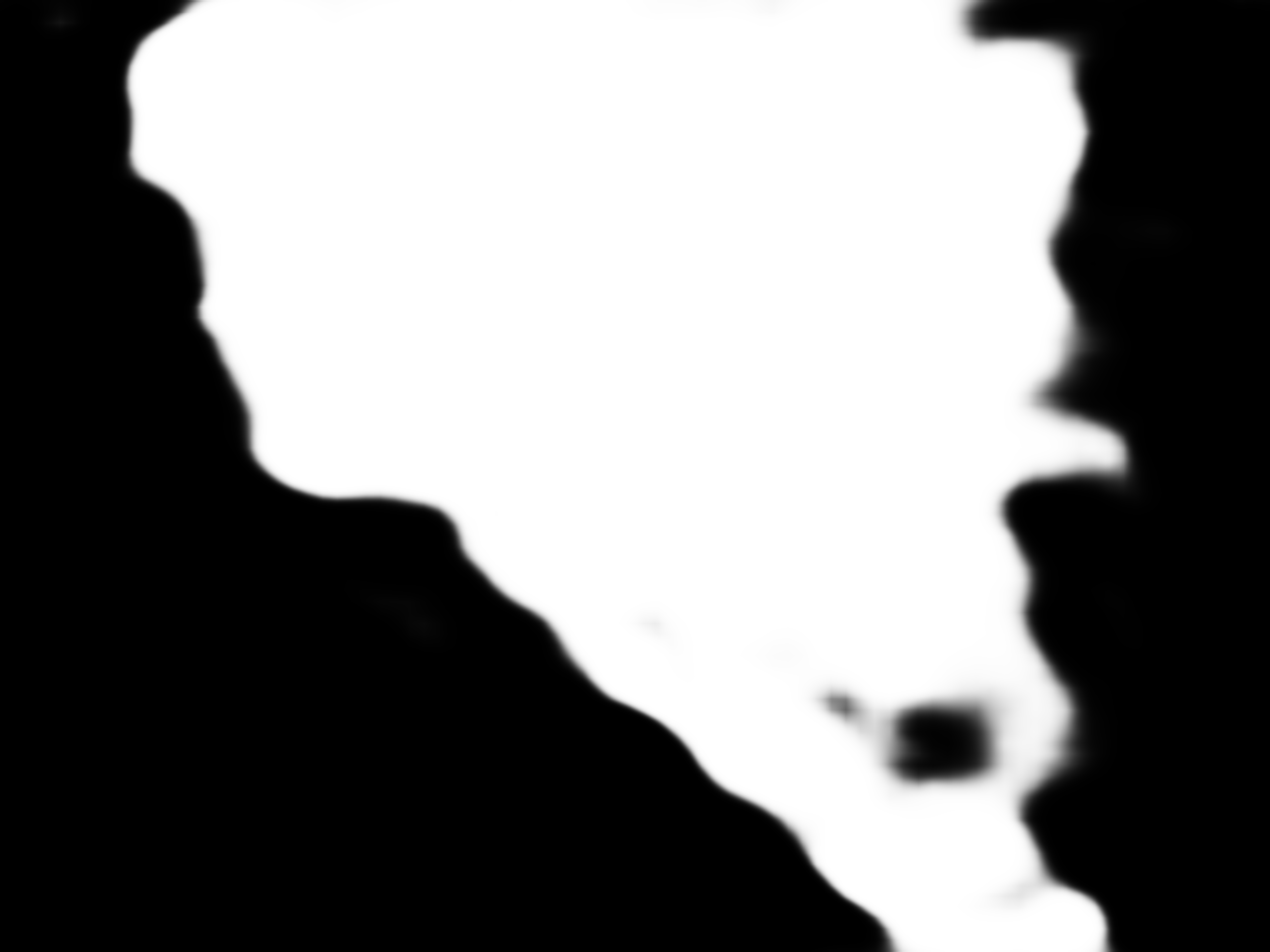}}&
   {\includegraphics[width=0.11\linewidth, height=0.09\linewidth]{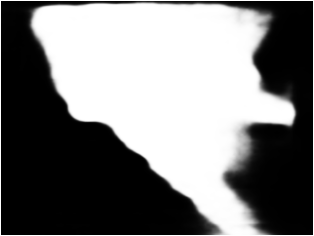}}
   
   \\
      {\includegraphics[width=0.11\linewidth, height=0.09\linewidth]{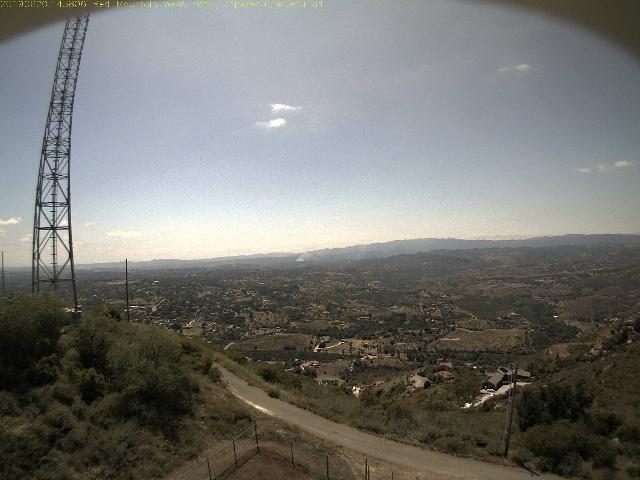}}&
   {\includegraphics[width=0.11\linewidth, height=0.09\linewidth]{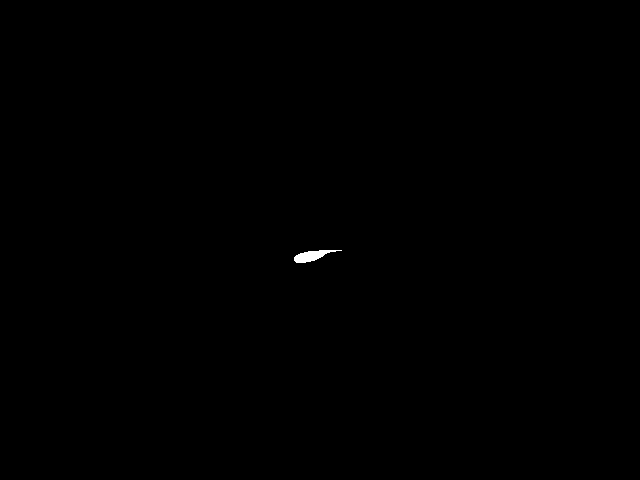}}&
   {\includegraphics[width=0.11\linewidth, height=0.09\linewidth]{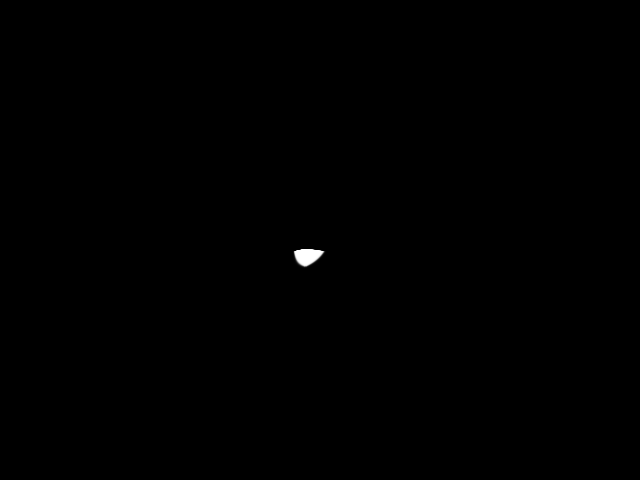}}&
   {\includegraphics[width=0.11\linewidth, height=0.09\linewidth]{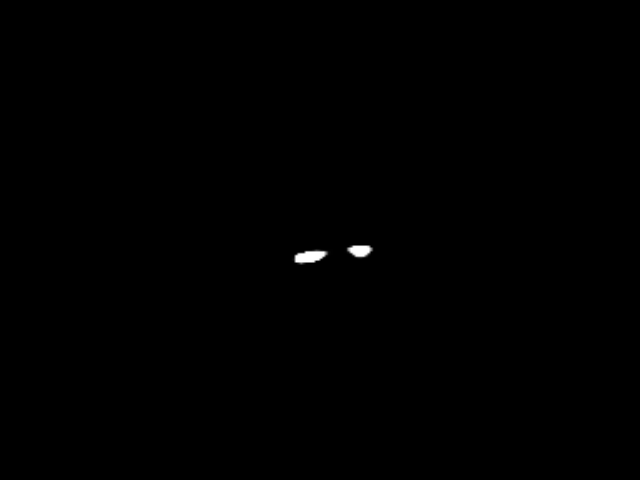}}&
   {\includegraphics[width=0.11\linewidth, height=0.09\linewidth]{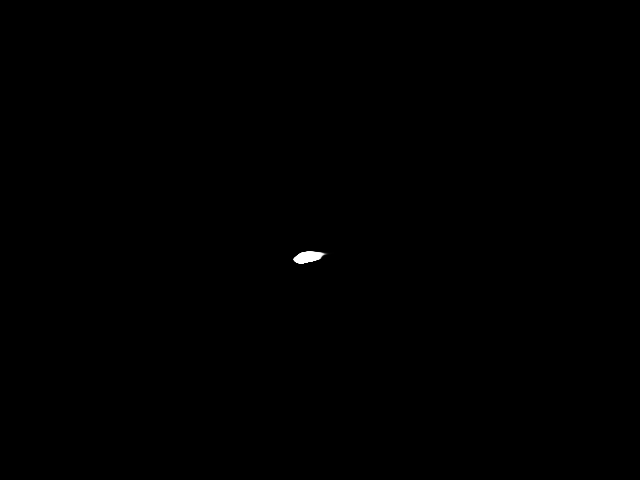}}&
      {\includegraphics[width=0.11\linewidth, height=0.09\linewidth]{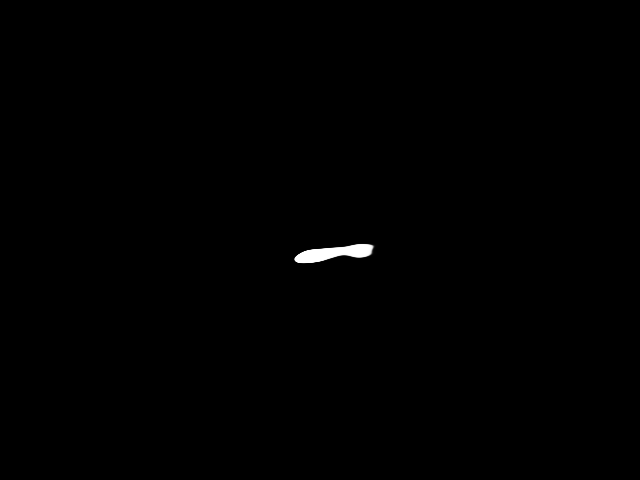}}&
   {\includegraphics[width=0.11\linewidth, height=0.09\linewidth]{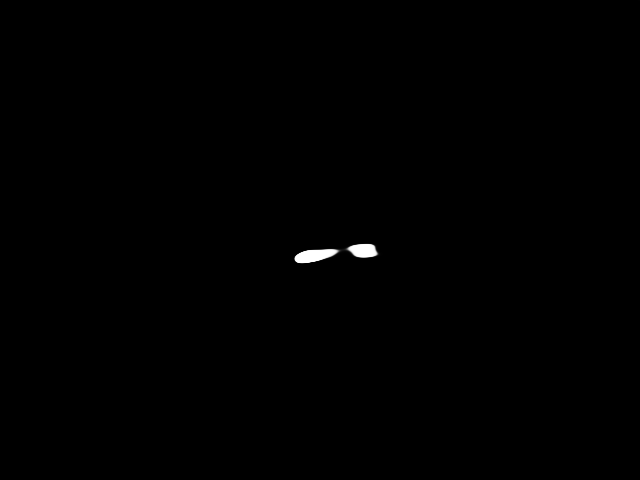}}&
   {\includegraphics[width=0.11\linewidth, height=0.09\linewidth]{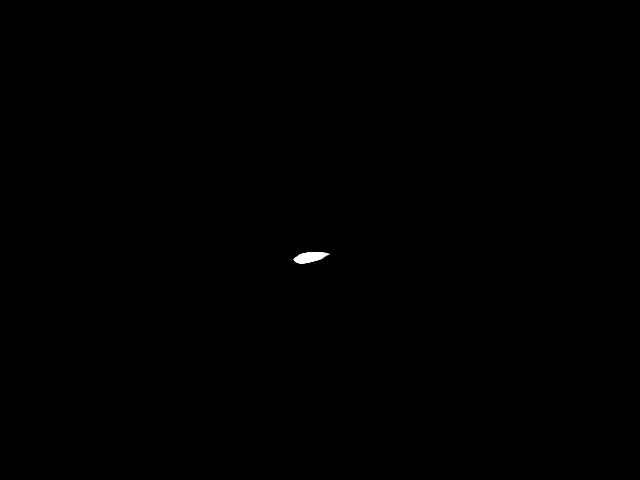}}
   
   \\
      {\includegraphics[width=0.11\linewidth, height=0.09\linewidth]{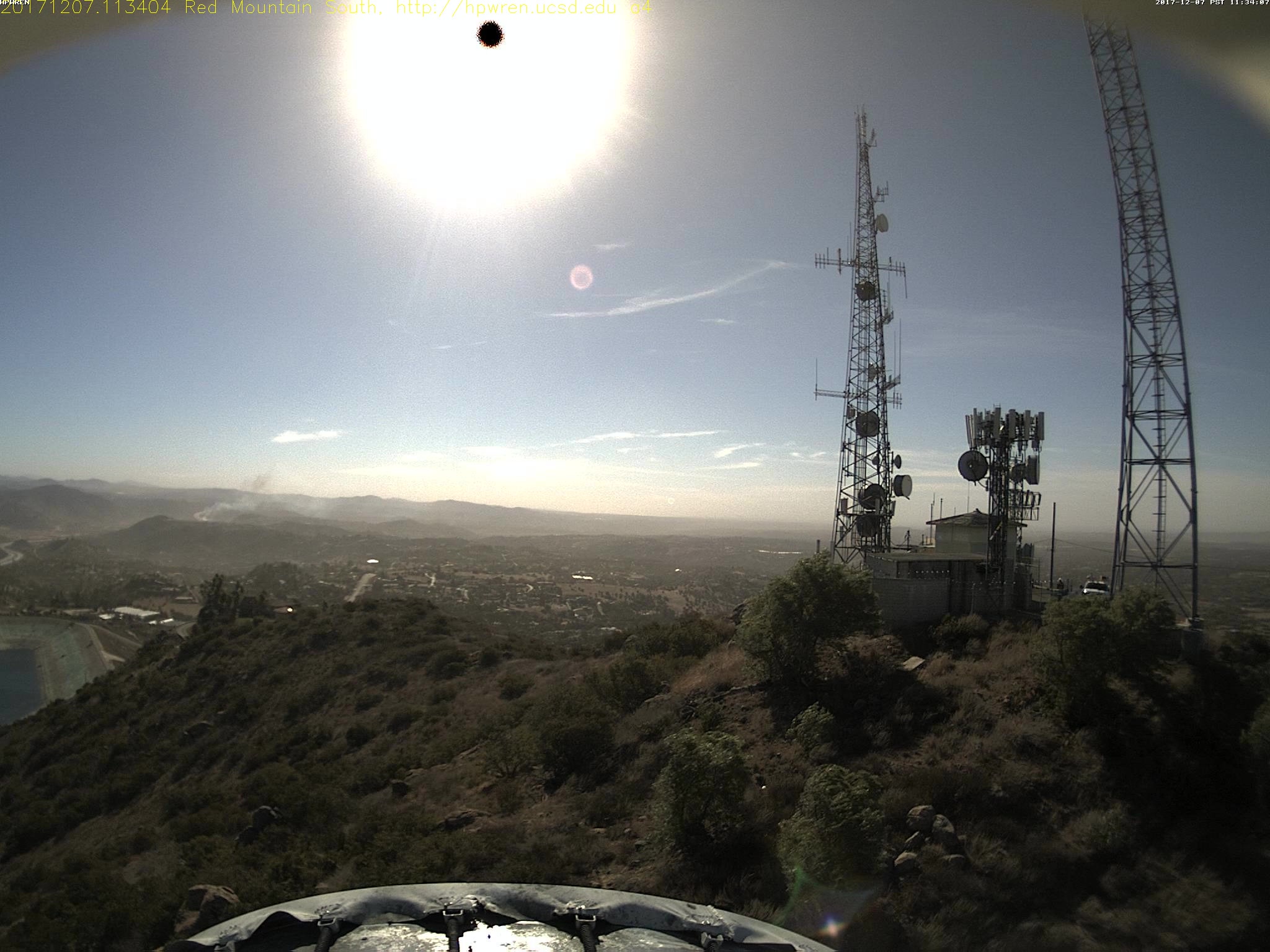}}&
   {\includegraphics[width=0.11\linewidth, height=0.09\linewidth]{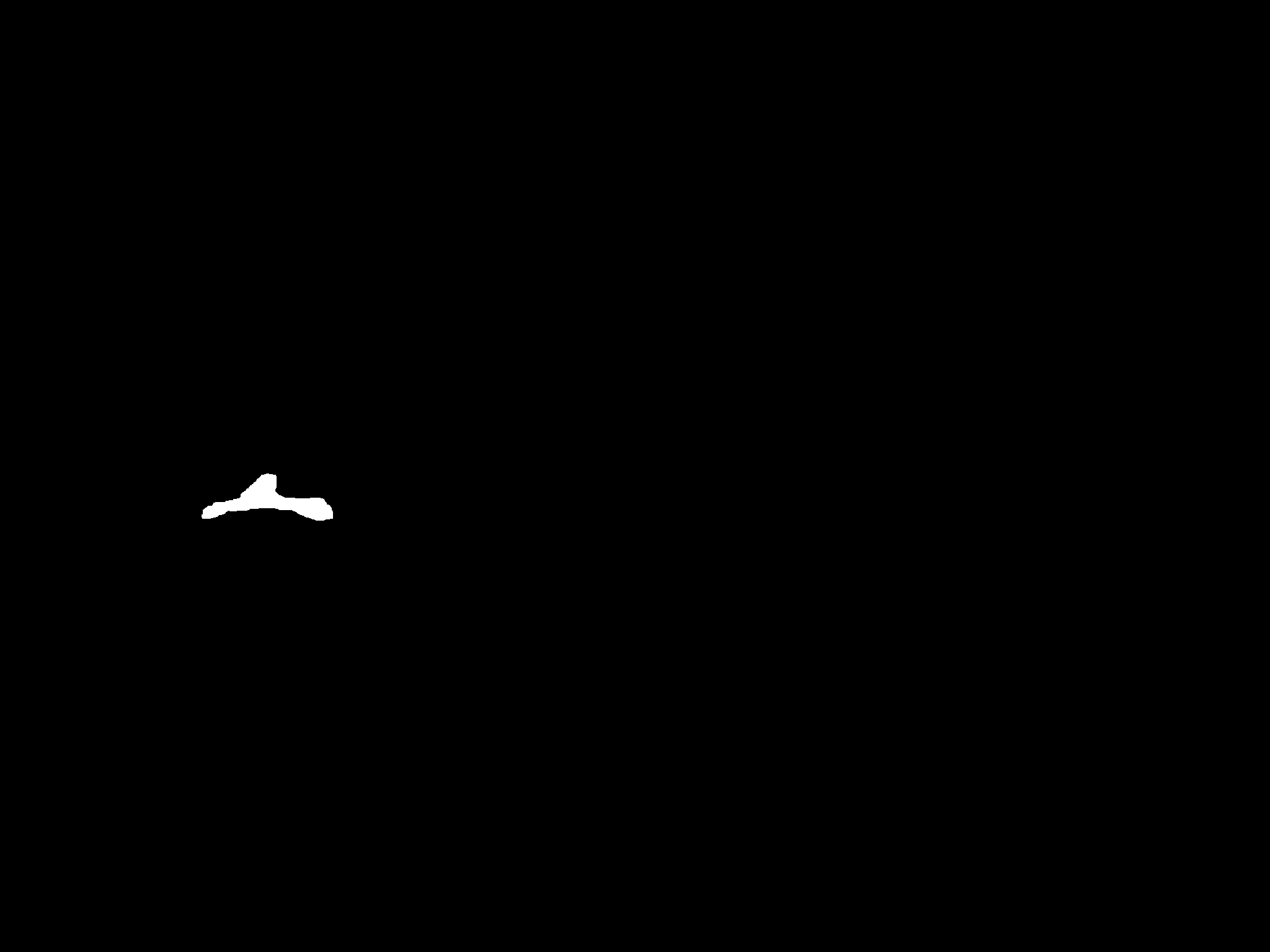}}&
   {\includegraphics[width=0.11\linewidth, height=0.09\linewidth]{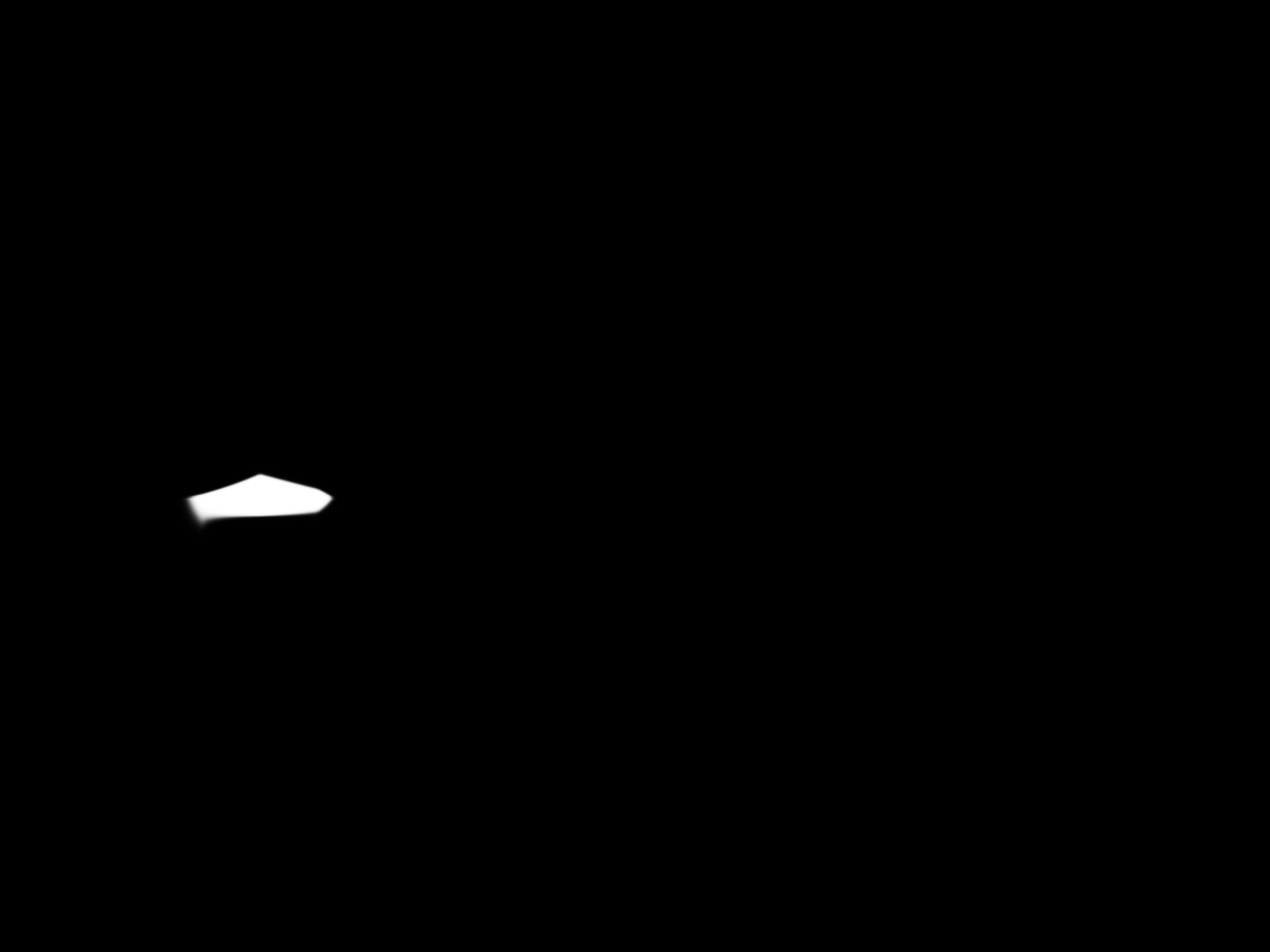}}&
   {\includegraphics[width=0.11\linewidth, height=0.09\linewidth]{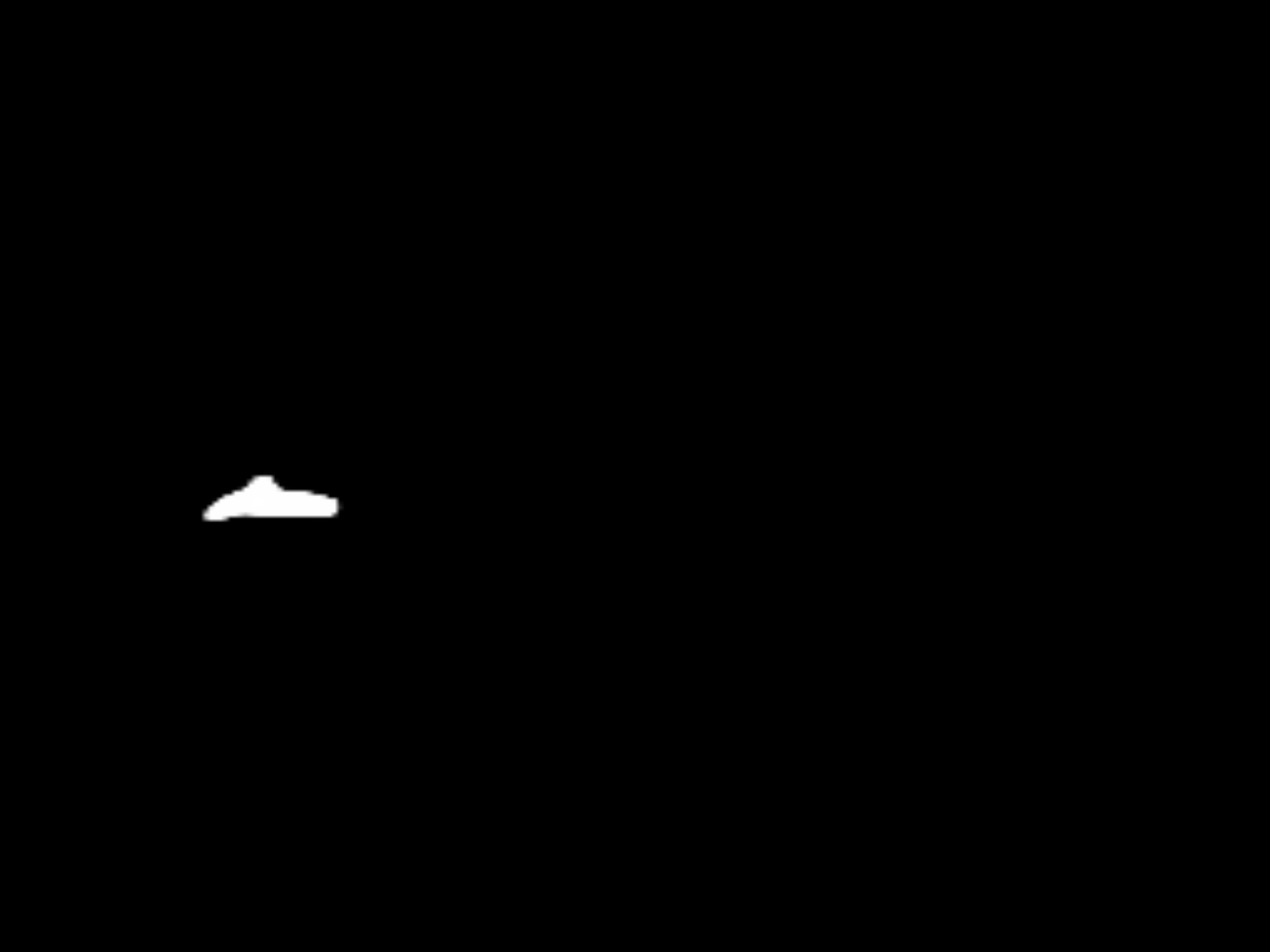}}&
   {\includegraphics[width=0.11\linewidth, height=0.09\linewidth]{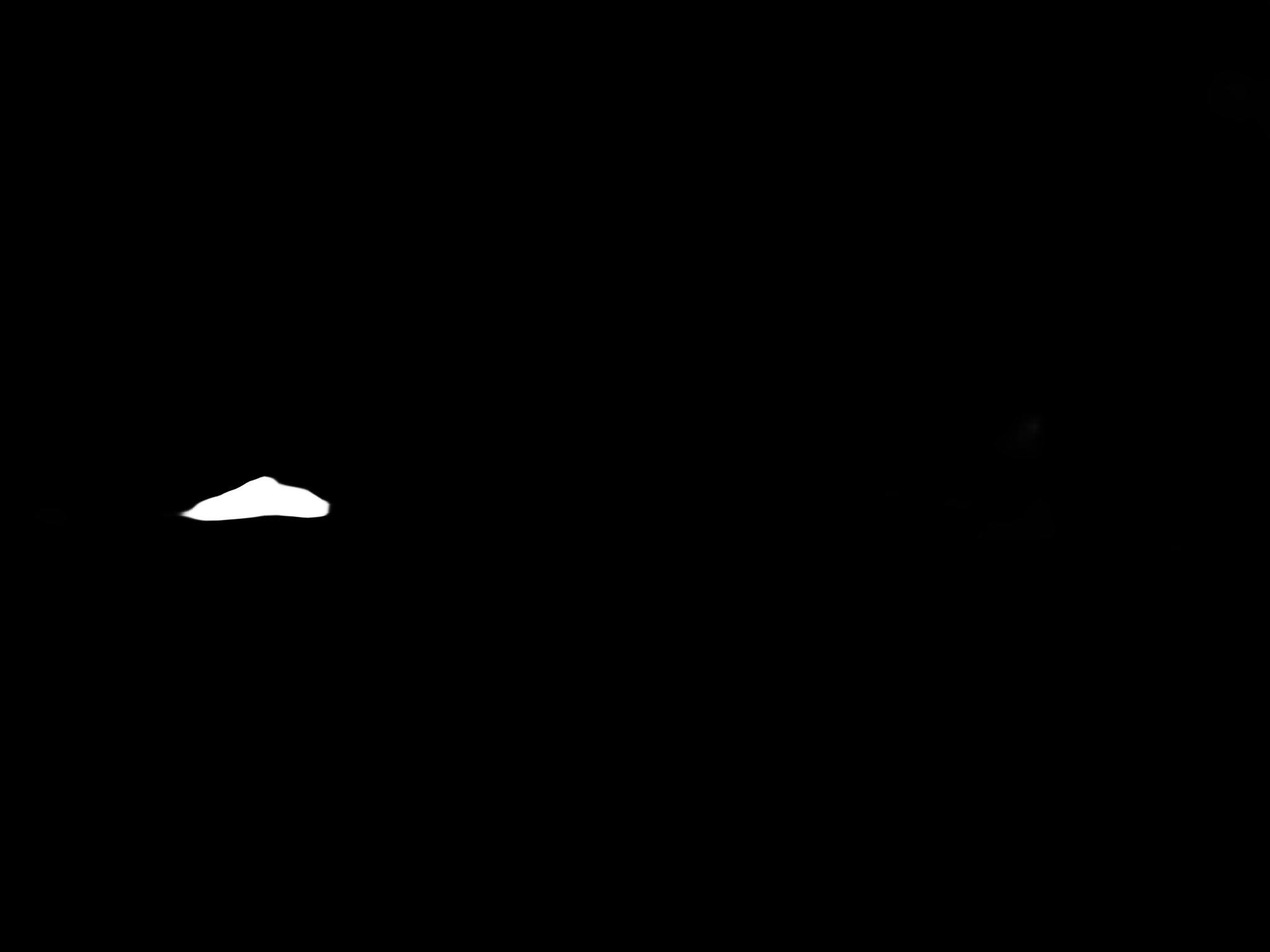}}&
      {\includegraphics[width=0.11\linewidth, height=0.09\linewidth]{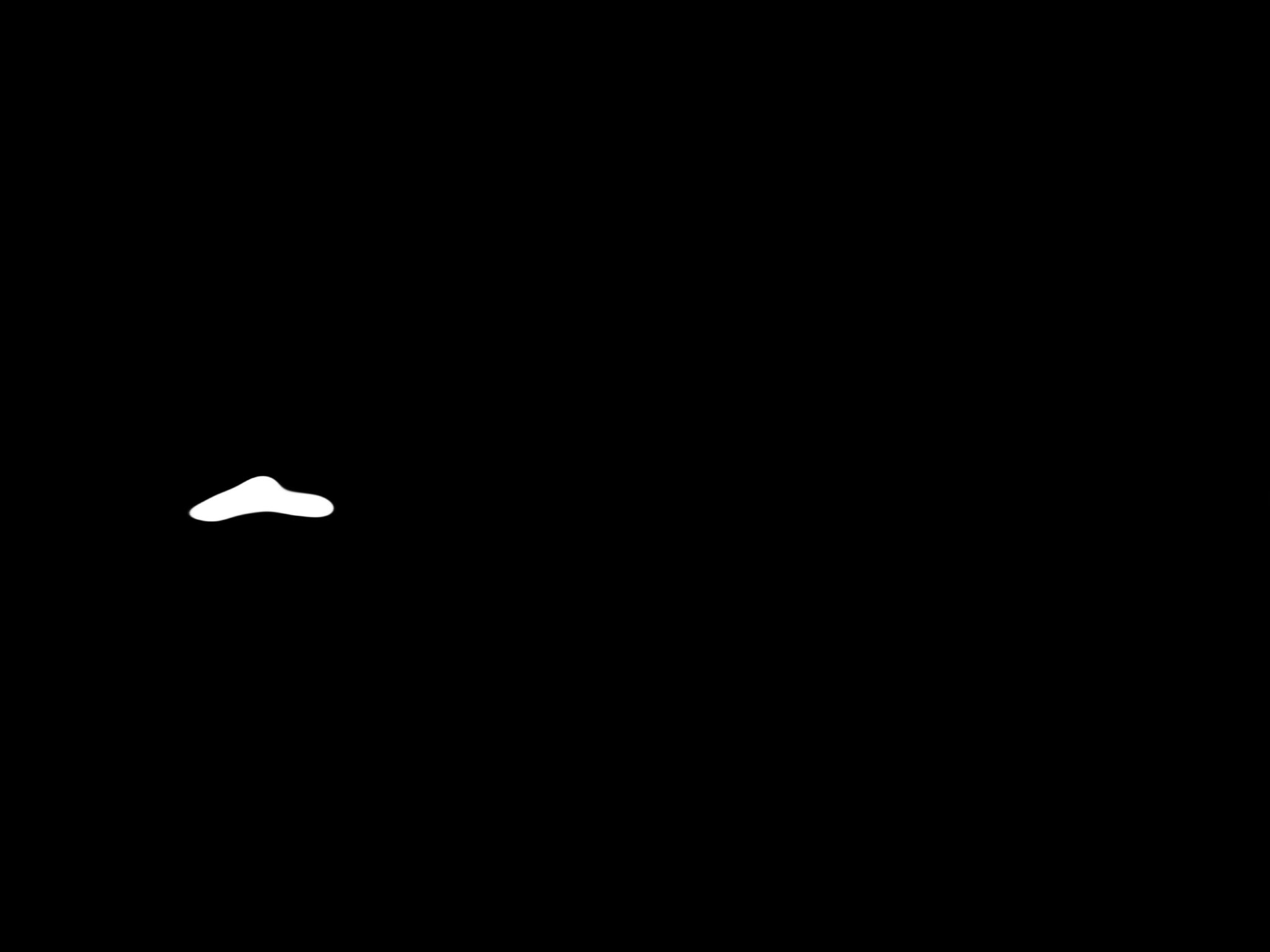}}&
   {\includegraphics[width=0.11\linewidth, height=0.09\linewidth]{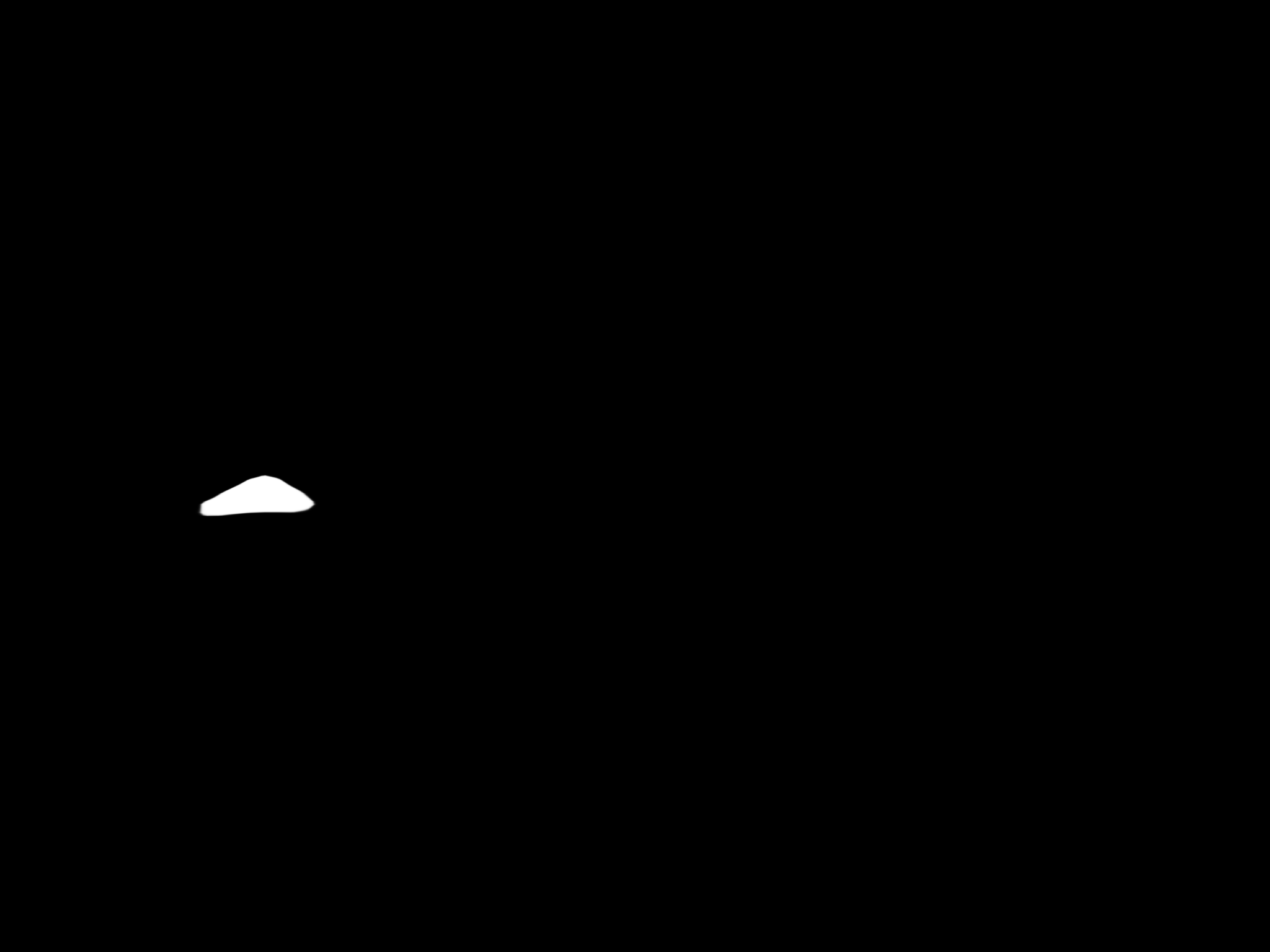}}&
   {\includegraphics[width=0.11\linewidth, height=0.09\linewidth]{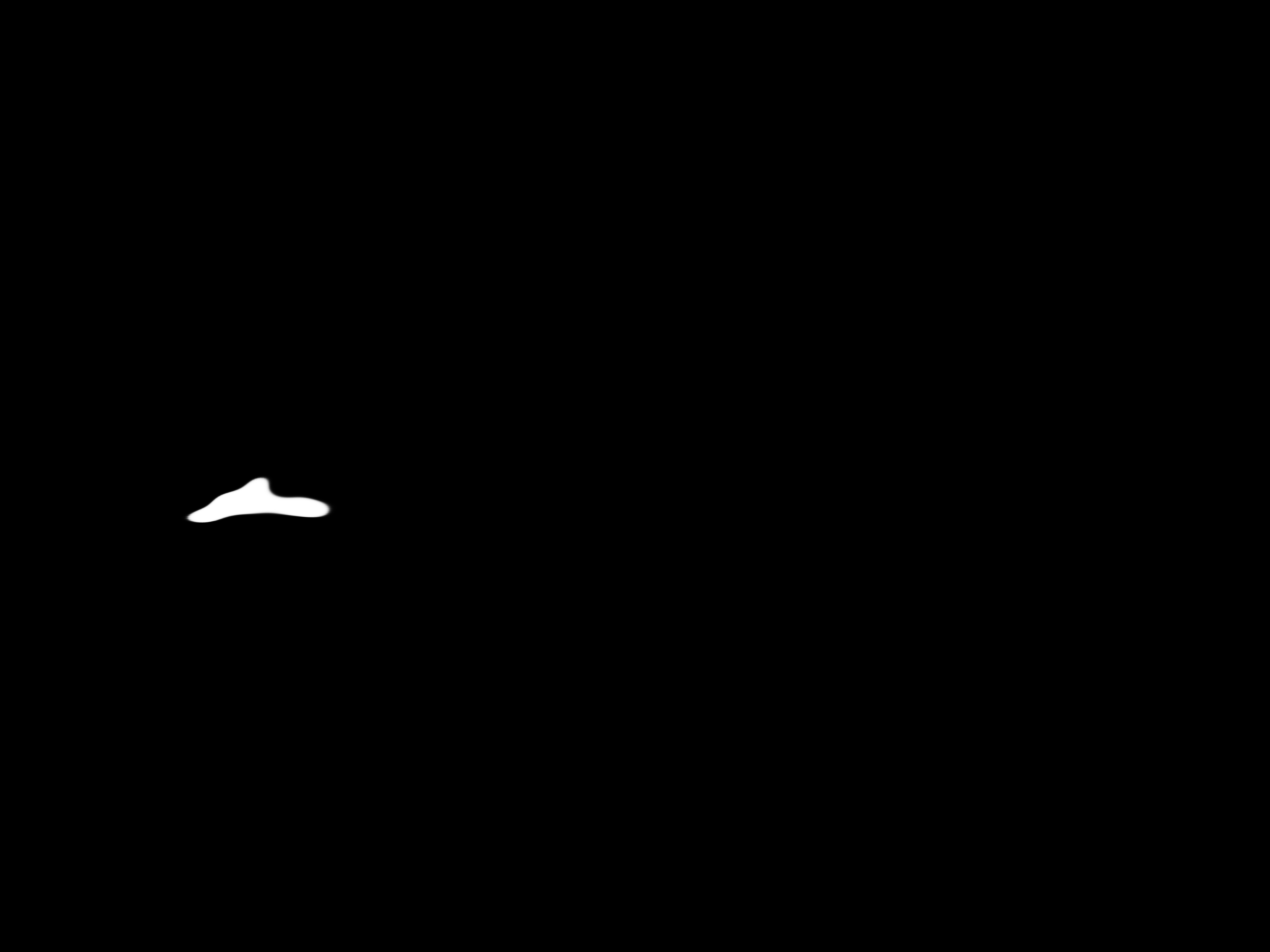}}
   
\\
     {\includegraphics[width=0.11\linewidth, height=0.09\linewidth]{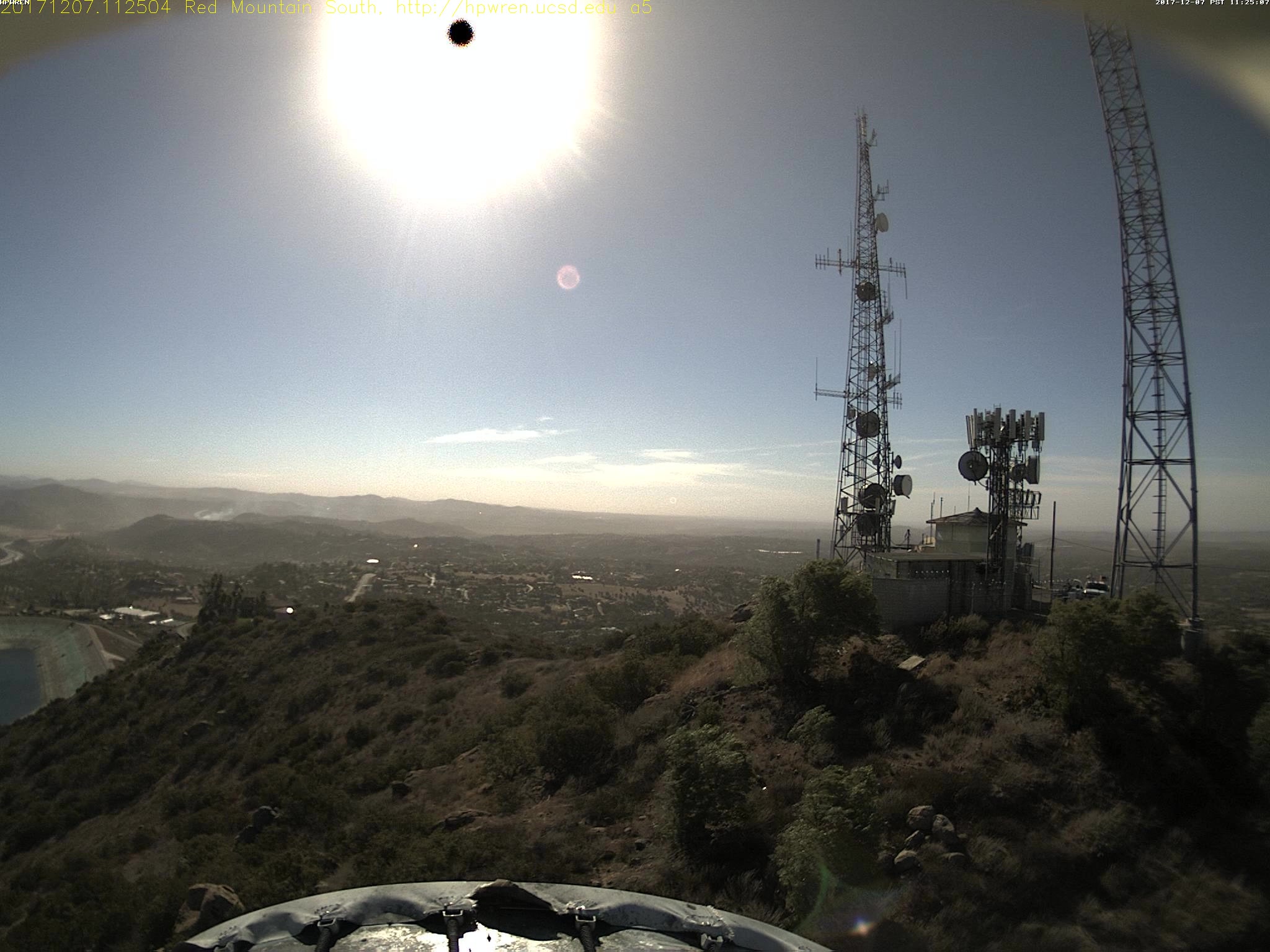}}&
   {\includegraphics[width=0.11\linewidth, height=0.09\linewidth]{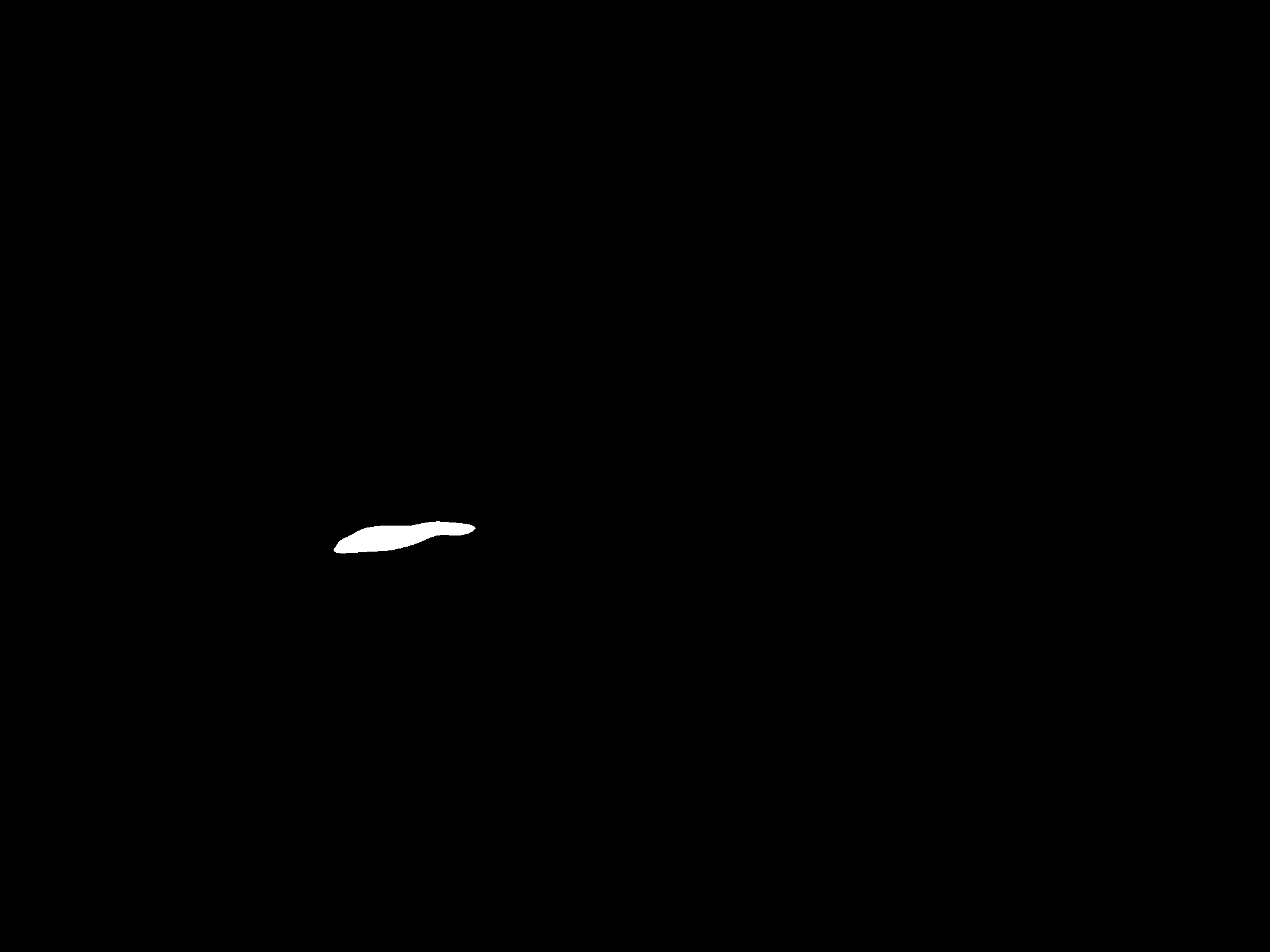}}&
   {\includegraphics[width=0.11\linewidth, height=0.09\linewidth]{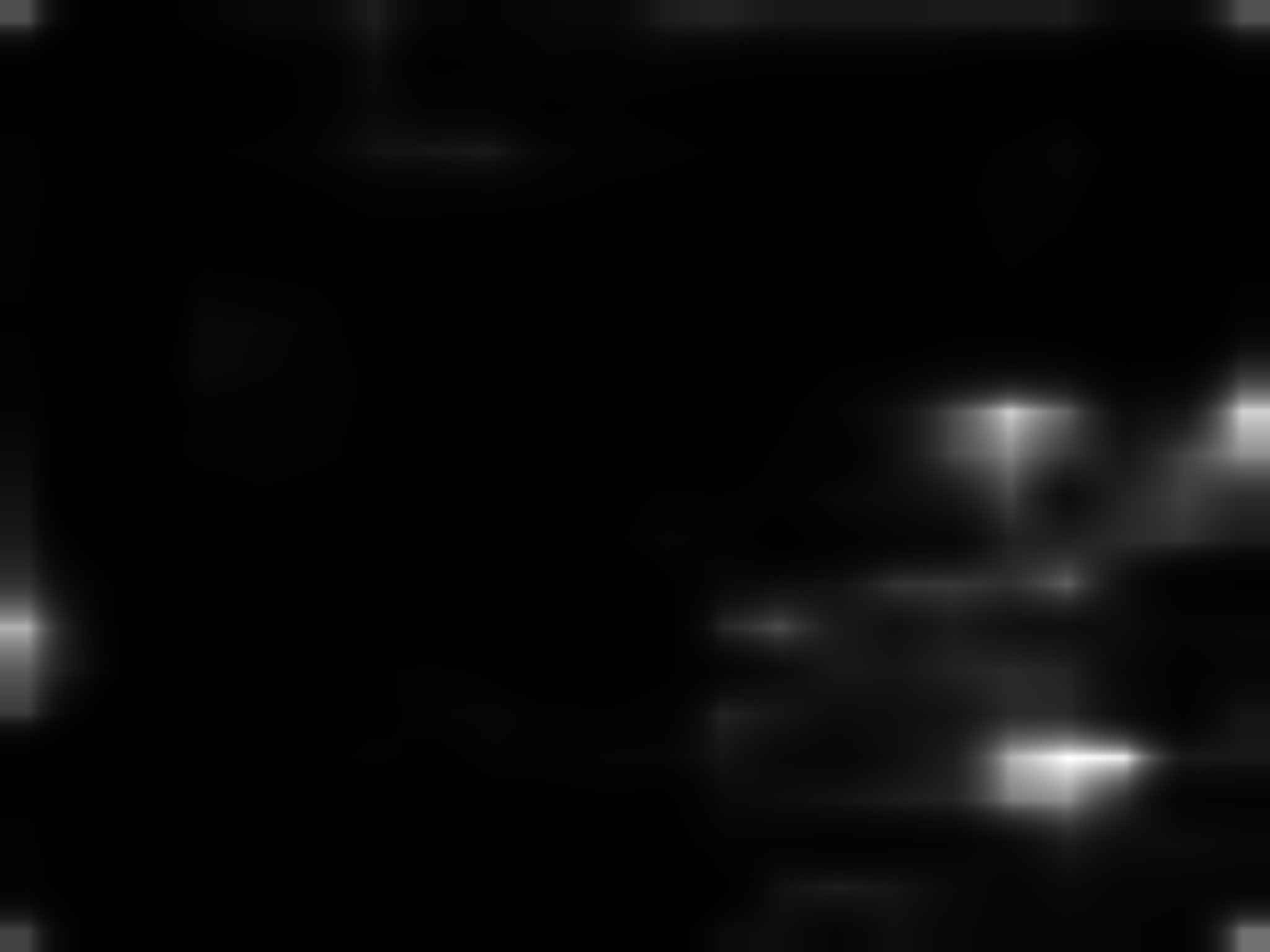}}&
   {\includegraphics[width=0.11\linewidth, height=0.09\linewidth]{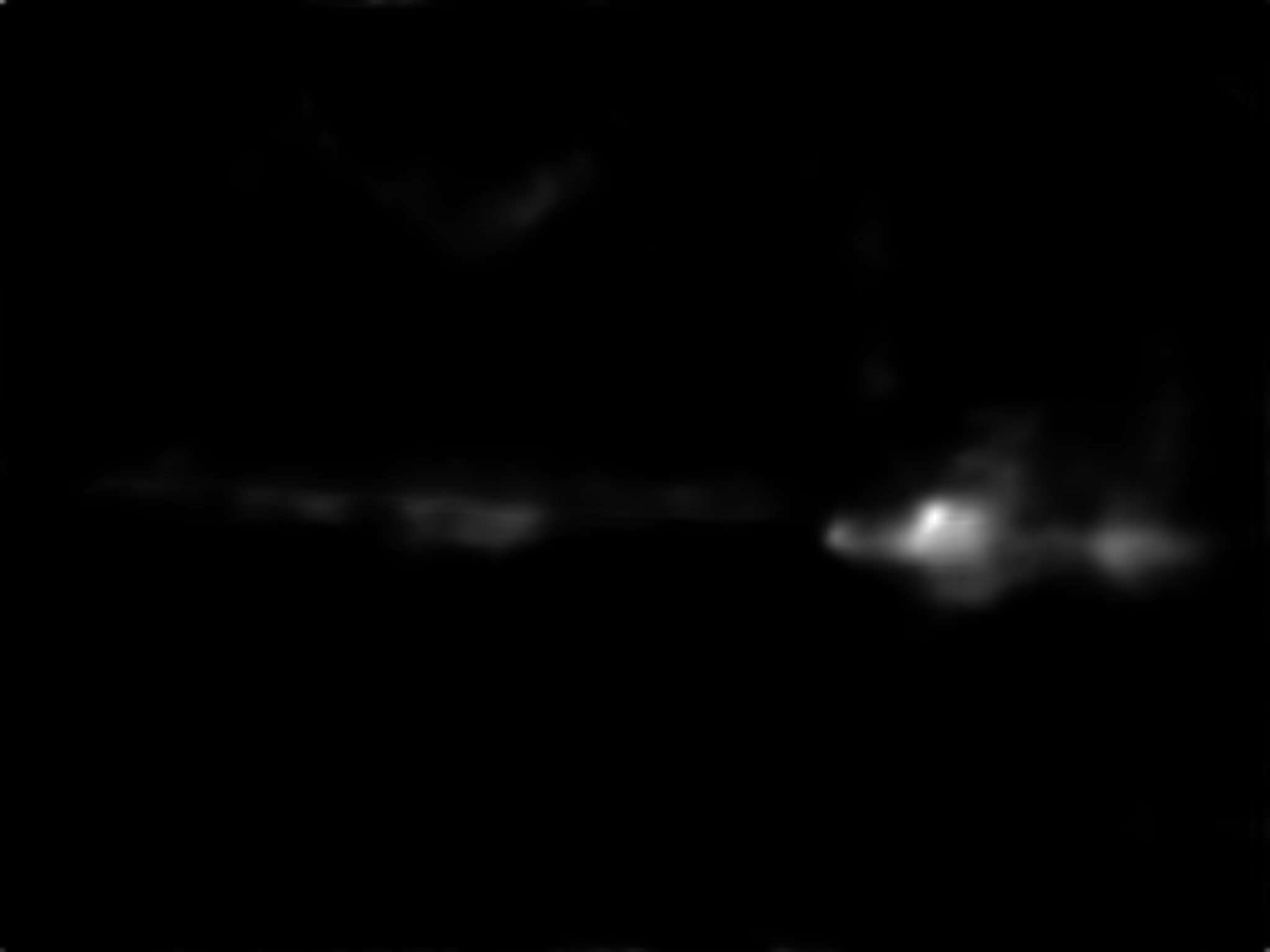}}&
   {\includegraphics[width=0.11\linewidth, height=0.09\linewidth]{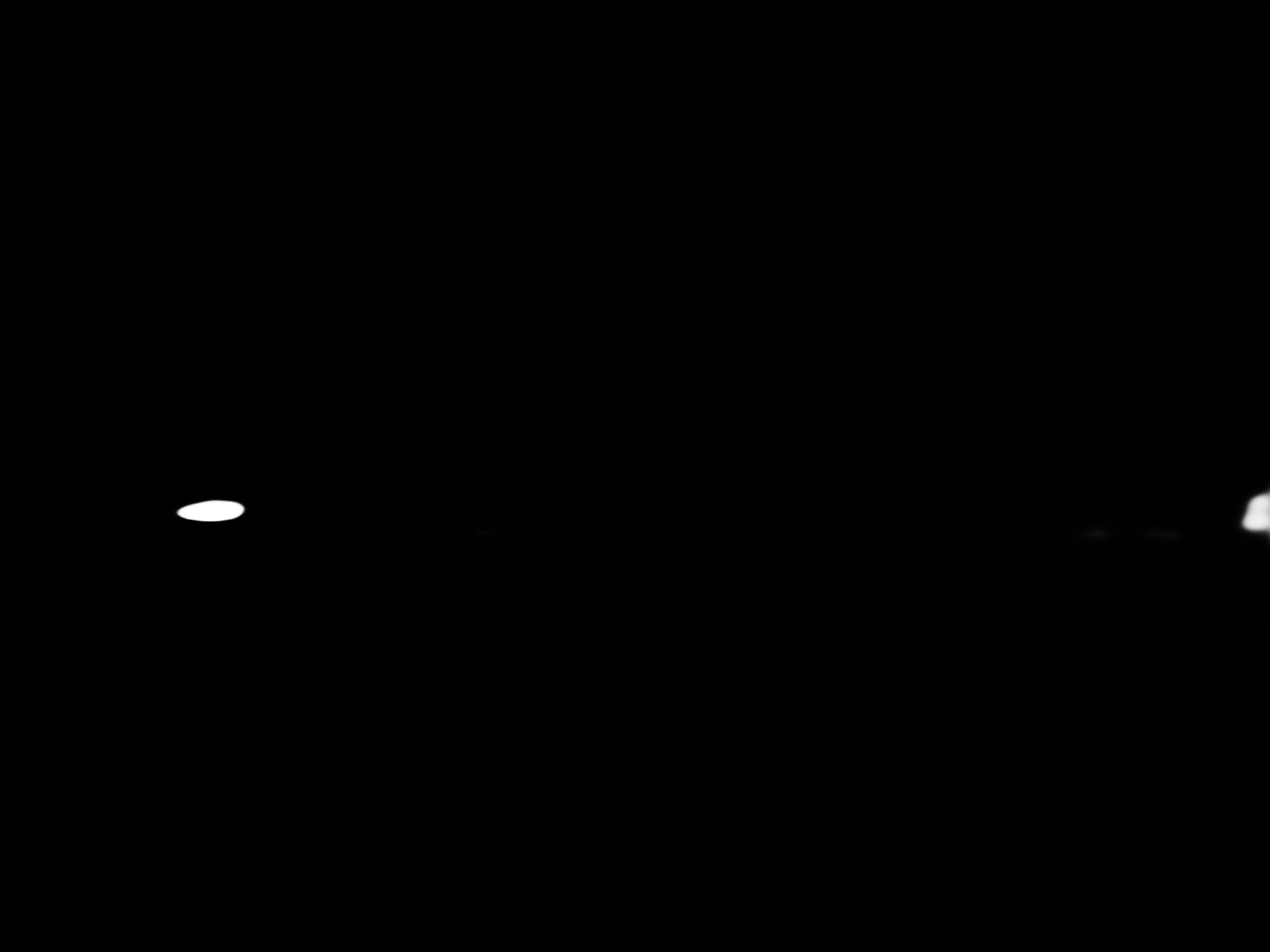}}&
      {\includegraphics[width=0.11\linewidth, height=0.09\linewidth]{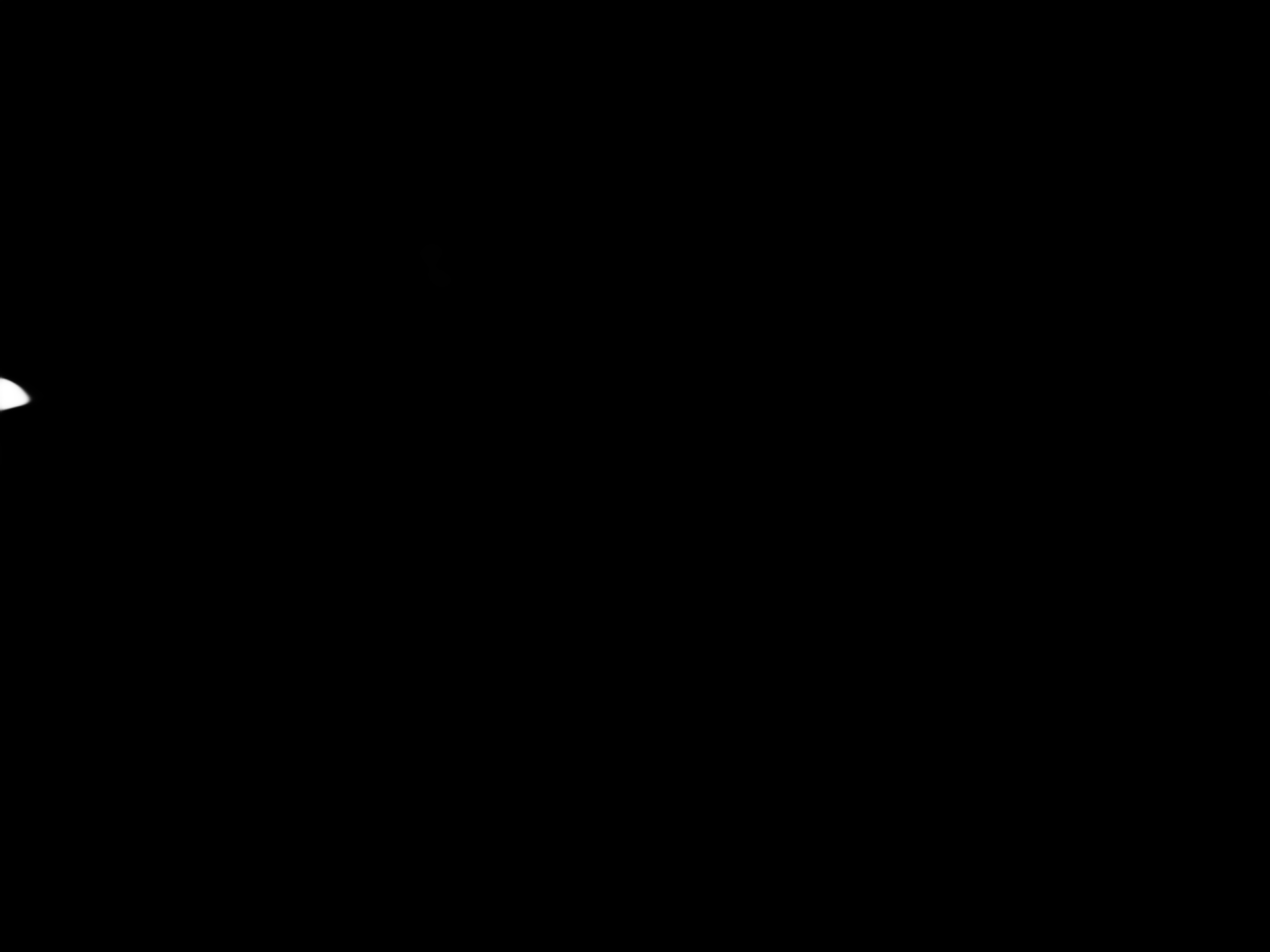}}&
   {\includegraphics[width=0.11\linewidth, height=0.09\linewidth]{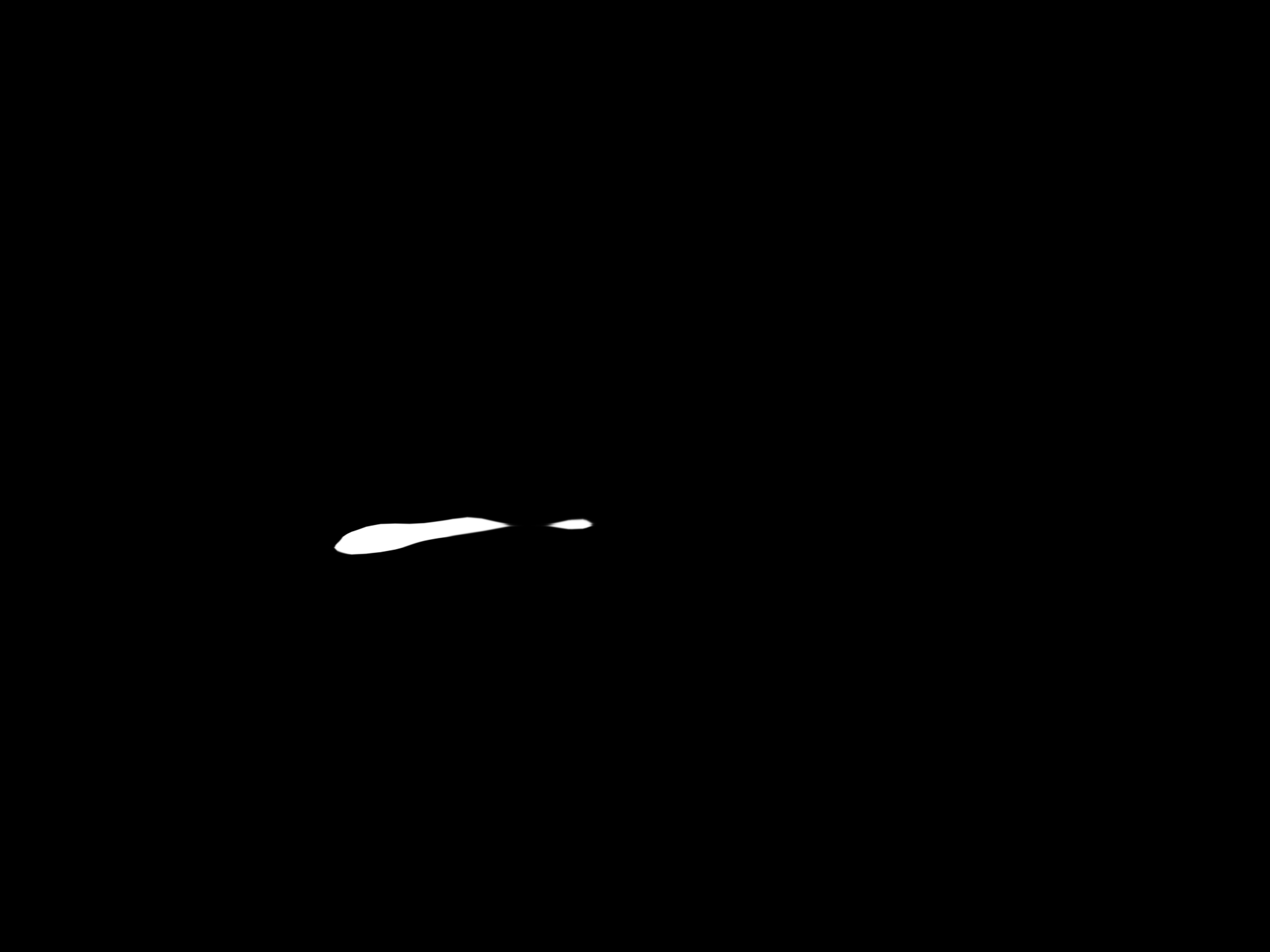}}&
   {\includegraphics[width=0.11\linewidth, height=0.09\linewidth]{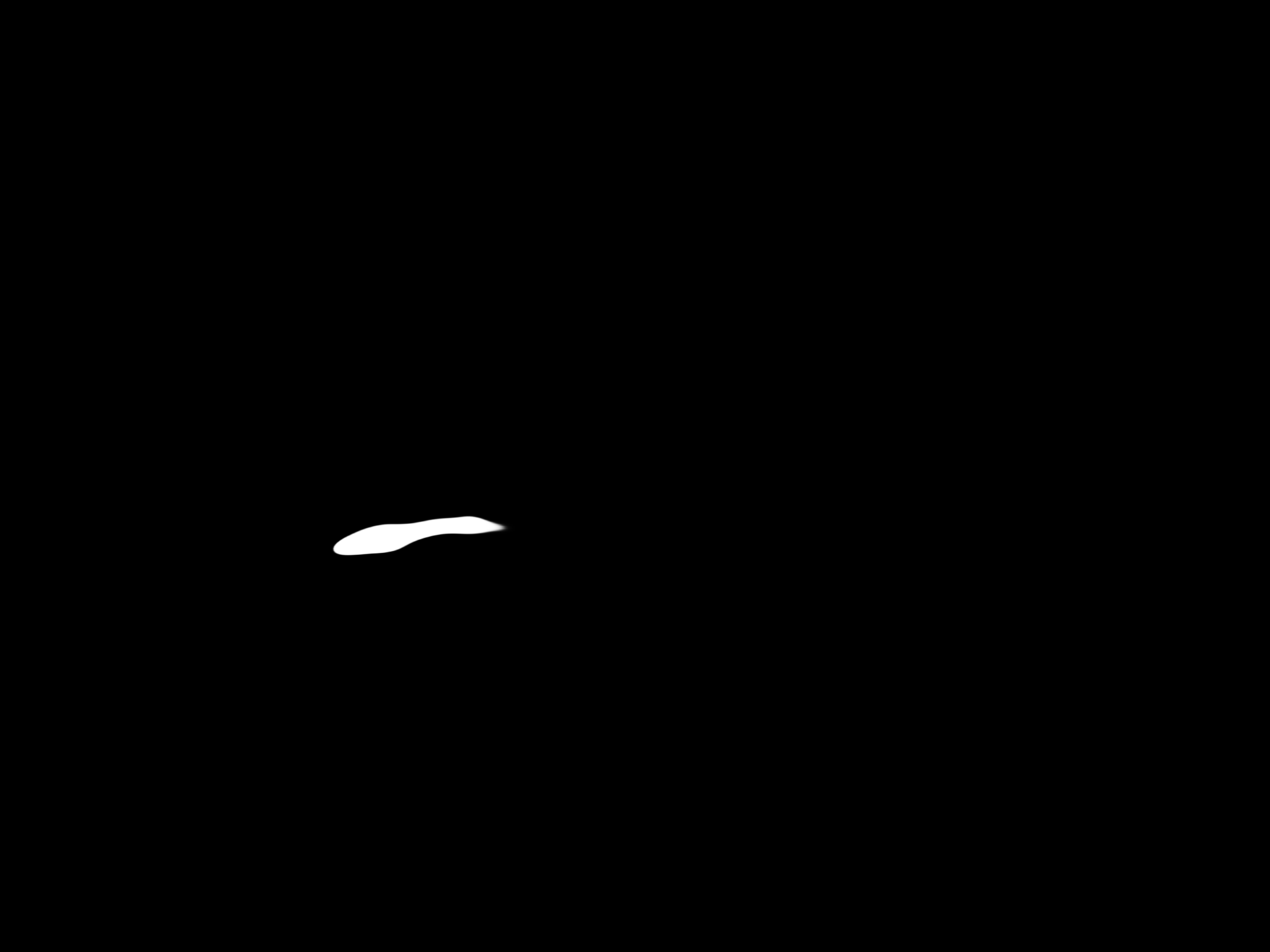}}
   
   \\
   \footnotesize{Image} & \footnotesize{GT} & \footnotesize{F3Net} & \footnotesize{BASNet} & \footnotesize{SCRN} & \footnotesize{ITSD} & \footnotesize{UCNet} & \footnotesize{Our} \\
   \end{tabular}
      \caption{Qualitative comparison of predictions by our model and other state-of-the-other models. The last two rows are smoke with different location in the same scenes.}
      \vspace{-2mm}
      \label{fig2}
   \end{center}
%   \setlength{\abovecaptionskip}{0.cm}
% \setlength{\belowcaptionskip}{-0.cm}
%   \vspace{2mm}
\end{figure*}

Comprehensive qualitative comparisons are visualized in Fig.~\ref{fig2}. It can be seen that our model can perform well in smoke with different scales, different shapes, and semi-transparent properties. As shown in  Fig.\ref{fig2},
% By seeing row 1st,4th, and 7th rows of Fig.\ref{fig2}, 
our model can capture the structure information well, which shows the effectiveness of our transmission loss.

\section{Dark Channel Prior Algorithm}

In this section, we introduce how we estimate the transmission map using the dark channel algorithm \cite{he} in more detail. It mainly includes two components.

Firstly, the dark channel of images is computed by \\
\begin{equation}
  I^{dark}(m)=\min \limits_{c \in {r,g,b}}(\min \limits_{n\in K(m)} (I^c(n))) 
  \label{d1}
\end{equation}\\
where $I^{c}$ is a color channel of an image $I$ and $K(m)$ is a local kernel of size $k\times k$ centered at $m$.

Secondly, we can estimate the transmission information of each pixel of the intensity image by:

\begin{equation}
    T(m)=1-\min \limits_{c}(\min\limits_{n \in K(m)}(\frac{I^c(n)}{A^c}))
\label{d2_supp}
\end{equation}
where $A$ is global atmospheric light, $I$ is the intensity image, $c$ is the color channel, $K(m)$ is a local patch of size $k\times k$ centered at pixel $m$ and then $\min \limits_{c}(\min\limits_{n
\in K(m)}(\frac{I^c(n)}{A^c})$ is the normalized haze map defiend by \cite{he}. We can see the estimated medium transmission is based on the global atmospheric light $A$. To get $A$, we follow \cite{he} and obtain it by picking the top $0.1\%$ brightest pixels in the dark channel $I^{dark}$ of the intensity image.
Finally, we use a guided filter \cite{g_filter} to refine transmission map $T$. The whole algorithm is described in Algorithm \ref{alg:Framwork}.

\begin{algorithm}[htb] 

\caption{ Dark channel prior based alogrithm.} 
\label{alg:Framwork} 
\hspace*{\algorithmicindent} \textbf{Input}: RGB image $I$.

\hspace*{\algorithmicindent} \textbf{Output}: A transmission map $\hat{T}.$
\begin{algorithmic}[1]

\STATE Calculate the dark channel image based on Equation \ref{d1};  

\STATE Estimate the global atmosphere light $A$ by picking the top 0.1\% brightest pixels in the dark channel, and the pixels with the highest intensity in the input image $I$ is selected as the atmospheric light $A$.;  

\STATE  Calculate the transmission map $T$ in Equation \ref{d2_supp}; 

\STATE Get the refined transmission map $\hat{T}$ by a guided filter;

\textbf{return} $\hat{T}$; 

\end{algorithmic} 
\end{algorithm}

\section{Discussions about Model detail}

\textbf{Hyperparameter Analysis:}
 (1) The dimension of latent space is important for the model's performance. We tried different numbers in the range: [2-32],
% tuned the hyper-parameter from 2 to 32 
and find relatively stable performance in the range: [8-16]. We finally set the latent space dimension as 8, yielding the best performance. (2) For the weights of different losses, empirically, we tried the transmission loss weight $\lambda_{1}$ in the range: [0.1-0.6], and
% 0.1, 0.2, 0.3, and 0.6, we 
observed the best performance when $\lambda_{1}=0.3$ and the worst when $\lambda_{1}=0.6$. For the uncertainty calibration loss weight $\lambda_{2}$, we tried the range: [0.005-0.1], and
% a range of 0.005 to 0.1 and  we 
observed the best performance when $\lambda_{2}=0.01$.\\
\textbf{Inference Time:} Processing speed is also an important factor for real-time smoke segmentation. During the inference stage, we only need to keep a portion of the model, which is the Bayesian latent variable model, to produce smoke detection results with no sampling process. Compared with conventional segmentation models, our model only additionally introduces a ﬁve layers CNN model. For input images of size 480 x 480, our model can process 6 images/sec, which is comparable with conventional segmentation models.

\end{document}